\documentclass[10pt,journal,compsoc]{IEEEtran}
\usepackage{etex}

\ifCLASSOPTIONcompsoc
  \usepackage[nocompress]{cite}
\else
  \usepackage{cite}
\fi
\ifCLASSINFOpdf
\else
\fi
\usepackage[export]{adjustbox}

\usepackage[export]{adjustbox}

\usepackage{epsfig}
\usepackage{graphicx}
\usepackage{amsmath}
\usepackage{amssymb}
\usepackage{makecell}
\usepackage{pbox}
\usepackage{xcolor}
\usepackage[utf8]{inputenc} % allow utf-8 input
\usepackage[T1]{fontenc}    % use 8-bit T1 fonts
\usepackage{url}
\usepackage{booktabs}       % professional-quality tables
\usepackage{amsfonts}       % blackboard math symbols
\usepackage{nicefrac}       % compact symbols for 1/2, etc.
\usepackage{microtype}      % microtypography
\usepackage{multirow,graphicx,paralist}
\usepackage{float}
\usepackage{wrapfig,lipsum}
\usepackage{dsfont}
\usepackage{mathtools}
\usepackage{algorithm,algpseudocode}
\usepackage{multicol}
\usepackage{ragged2e}

\usepackage{xcolor}
\DeclarePairedDelimiter\floor{\lfloor}{\rfloor}

\usepackage{array}
\newcommand{\rowfonttype}{}% Current row font
\newcommand{\rowfont}[1]{% Set current row font
	\gdef\rowfonttype{#1}#1%
}

\makeatletter
\newcommand*{\rom}[1]{\expandafter\@slowromancap\romannumeral #1@}

\usepackage[colorinlistoftodos]{todonotes}
\usepackage[letterpaper=true,colorlinks=true, linkbordercolor =red, urlcolor=black, bookmarks=false]{hyperref}

\usepackage{url}	
\colorlet{violet}{black}

\begin{document}
\title{Relative Saliency and Ranking: \\ Models, Metrics, Data and Benchmarks}

\author{Mahmoud Kalash\textsuperscript{\textbf{*}},
        Md Amirul Islam\textsuperscript{\textbf{*}},
        and~ Neil D. B. Bruce% <-this % stops a space
\IEEEcompsocitemizethanks{\IEEEcompsocthanksitem $^*$ Both authors contributed equally to this work
	
	\IEEEcompsocthanksitem M. Kalash was with the Department
of Computer Science, University of Manitoba, Canada.\protect 	\hspace{0.1cm} E-mail: kalashm@cs.umanitoba.ca
\IEEEcompsocthanksitem M. A. Islam (Corresponding author) and N. Bruce are with the Department of Computer Science, Ryerson University and Vector Institute, Toronto, Canada.% <-this % stops an unwanted space
\hspace{0.1cm} E-mail: \{amirul, bruce\}@scs.ryerson.ca}
}

\markboth{IEEE TRANSACTIONS ON PATTERN ANALYSIS AND MACHINE INTELLIGENCE}
{Shell \MakeLowercase{\textit{et al.}}: Bare Demo of IEEEtran.cls for Computer Society Journals}



\IEEEtitleabstractindextext{%
	\justify
\begin{abstract}
%\todo[inline]{Please go through the Abstract to incorporate new changes in the paper if needed}
Salient object detection is a problem that has been considered in detail and \textcolor{black}{many solutions have been proposed}. In this paper, we argue that work to date has addressed a problem that is relatively ill-posed. Specifically, there is not universal agreement about what constitutes a salient object when multiple observers are queried. This implies that some objects are more likely to be judged salient than others, and implies a relative rank exists on salient objects. Initially, we present a novel deep learning solution based on a hierarchical representation of relative saliency and stage-wise refinement. Further to this, we present data, analysis and baseline benchmark results towards addressing the problem of salient object ranking. Methods for deriving suitable ranked salient object instances are presented, along with metrics suitable to measuring algorithm performance. In addition, we show how a derived dataset can be successively refined to provide cleaned results that correlate well with pristine ground truth in its characteristics and value for training and testing models. Finally, we provide a comparison among prevailing algorithms that address salient object ranking or detection to establish initial baselines providing a basis for comparison with future efforts addressing this problem. \textcolor{black}{The source code and data are publicly available via our project page:} \textrm{\href{https://ryersonvisionlab.github.io/cocosalrank.html}{ryersonvisionlab.github.io/cocosalrank}}
\end{abstract}

\begin{IEEEkeywords}
%\todo[inline]{Check the index terms as we added two new terms as suggested}
Saliency, Saliency Ranking, \textcolor{violet}{Salient Instance}, Salient Object Detection, Relative Rank, Dataset, \textcolor{violet}{Benchmark}
\end{IEEEkeywords}}

\maketitle

\IEEEdisplaynontitleabstractindextext

% \fi
\IEEEpeerreviewmaketitle

\section{Introduction}
%\todo[inline]{Please go through the whole introduction to make sure it is consistent}
\textcolor{black}{The problem of salient object detection~\cite{cheng2014global,itti1998model, liu2010learning, perazzi2012saliency, goferman2011context}} ~\cite{ yan2013hierarchical,shi2013pisa,yang2013saliency,li2014secrets} has been well studied, and much progress has been made. The objective in this problem domain is to select an object or objects in an image that are important, striking, stand-out or draw attention. The majority of work in salient object detection considers either a single salient object~\cite{amulet_2017, ucf_2017, he2017delving, hou2016deeply, wang2016saliency, wang2017stagewise, hu2017deep,li2016deep, lee2016deep, liu2016dhsnet, zhao2015saliency,li2015visual} or multiple salient objects~\cite{jia2015adaptive, najibi2017towards, zhang2016unconstrained, Wang_2018_CVPR}, but does not consider that what is salient may vary from one person to another, and certain objects may be met with more universal agreement concerning their importance. In this work, we bring to light a consideration that has long been neglected in this domain; individual observers may have differences in opinion about what is salient, and moreover, the definition of a salient object is relatively equivocal. This implies that while one or more objects may be salient, there may be more agreement for certain object than others. With respect to a problem definition, salient object detection can be extended to a problem of ranking salient objects. \textcolor{violet}{As such, \textit{saliency ranking} requires the detection of salient instances in a given image, and assignment of a rank to each salient instance based on its degree of saliency.}
%\todo[inline]{Definition of Saliency Ranking as suggested. Please feel free to modify it.}

There is a paucity of data that includes salient objects that are hand-segmented by multiple observers. It is important to note that any labels provided by a small number of observers (including one) does not allow for discerning the relative importance of objects. Implicit assignment of relative salience based on gaze data~\cite{xia2017and} also presents difficulties, given a different cognitive process than a calculated decision that involves manual labeling~\cite{koehler2014saliency}. Moreover, gaze data is relatively challenging to interpret given factors such as centre bias, visuomotor constraints, and other latent factors~\cite{bruce2015computational, accik2014real}. To overcome some of these shortcomings, we have re-purposed the PASCAL-S dataset~\cite{li2014secrets} via further processing \textcolor{violet}{(described in Sec.~\ref{sec:stack})} to provide a set of data denoted as PASCAL-SR that overcomes some of the limitations of traditional efforts.
\begin{figure}
	\centering
	\setlength\tabcolsep{1.0pt}
	\def\arraystretch{0.3}
	\resizebox{0.4\textwidth}{!}{
		\begin{tabular}{*{3}{c }}
			\includegraphics[width=0.24\textwidth]{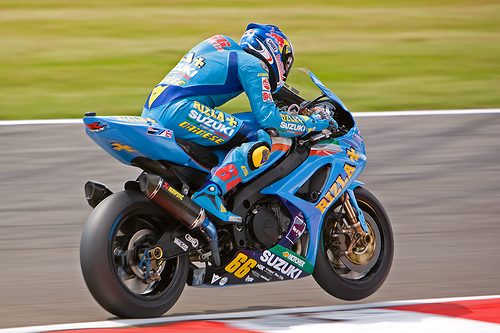}&
			\includegraphics[width=0.24\textwidth]{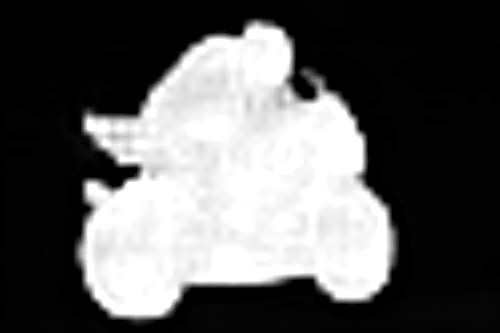} &
			\includegraphics[width=0.24\textwidth]{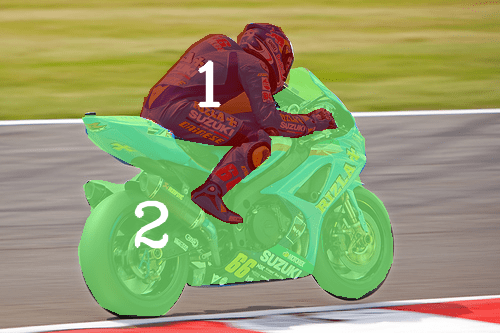}
			\\
			
			\includegraphics[width=0.24\textwidth]{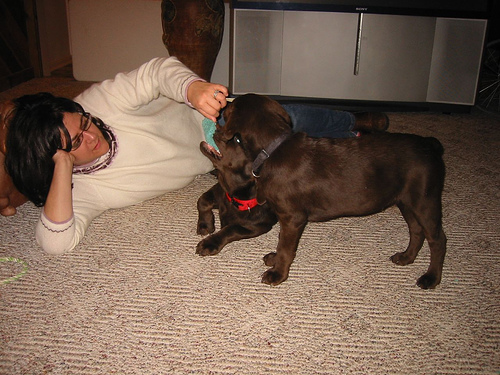}&
			\includegraphics[width=0.24\textwidth]{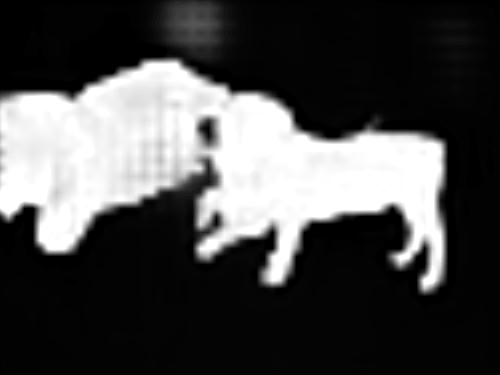} &
			\includegraphics[width=0.24\textwidth]{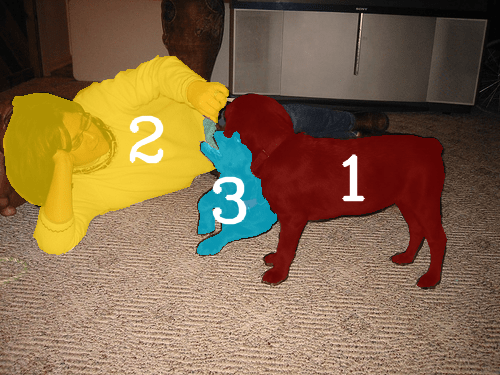}
			\\			
		\end{tabular}}
		\caption{We present a solution in the form of a deep neural network to detect salient objects and consider their relative ranking based on salience of these objects. Left to right: input image, detected salient regions, and rank order of salient objects. Assigned numbers and colors indicate the rank order of different salient object instances.}
		\label{fig:intro1}
		\vspace{-0.4cm}
	\end{figure}

Therefore, in this paper we consider the problem of salient object detection more broadly. This includes detection of all salient regions, and accounting for inter-observer variability by assigning confidence to different salient regions. We augment the PASCAL-S dataset via further processing to provide ground truth in a form that accounts for relative salience. Success is measured against other algorithms based on the rank order of salient objects relative to ground truth orderings in addition to traditional metrics. %Recent efforts also consider the problem of \emph{salient object subitizing}.
It is our contention that this determination should be possible by a model that provides detection of salient objects (see Fig.~\ref{fig:intro1}).
\begin{figure}
	\centering
	\includegraphics[width=0.49\textwidth]{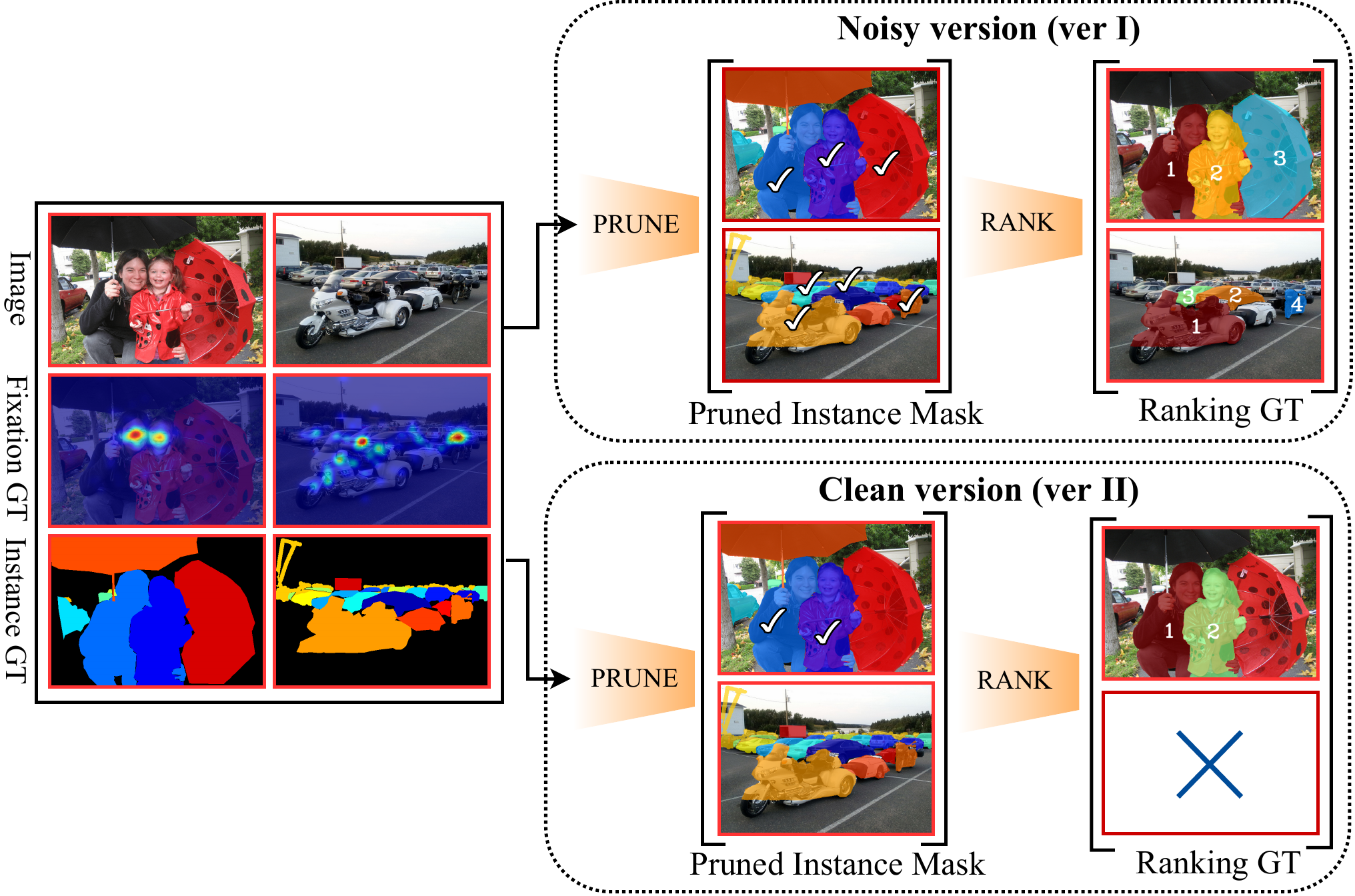}
	\vspace{-0.7cm}
	\caption{An illustration of the COCO-SalRank dataset. Our dataset provides salient object instances and their relative rank order (relative salience). Due to the large number of instances present in some images, an \textit{instance pruning} process assigns a rank only to instances that receive a sufficiently high degree of attention. We provide two different versions of our proposed ranking dataset in the form of a noisy and cleaned version. Assigned numbers and colors indicate the rank order of salient object instances with chroma corresponding to a numeric scale.}
	\label{fig:dataset}
	\vspace{-0.4cm}
\end{figure}

\textcolor{violet}{While the PASCAL-S dataset provides selections from multiple observers, this takes a different form than the data derived from a re-purposed version of this dataset included in the current study. For PASCAL-S, results are reported based on varying the threshold required among observers for an object to be considered salient. This implies a solution to multiple binary salient object detection problems and hints at the ill-posed nature of the problem. What this does not consider is an explicit ranking or ordering of salient objects and instead considers instances of the traditional problem subject to a varying threshold for making a binary judgment of what is salient. In contrast, PASCAL-SR dataset is non-binary and includes a specific relative ranking. This requires a non-binary decision and a more balanced control over precision and recall in assigning saliency. Moreover, it's characteristics provide a guide for the development of the more extensive COCO-SalRank dataset presented.}

Although the PASCAL-SR dataset may be made suitable for addressing the problem of ranking salient objects; in order to train deep learning models a significant number of examples are needed; this is also true of evaluation of algorithms that seek to assign a relative rank to salient objects. Therefore, we extend our work in this paper to provide a dataset for saliency ranking, and associated benchmarks, based on images from the MS-COCO dataset~\cite{lin2014microsoft}. This is accomplished by combining existing measurements diagnostic of human attention~\cite{jiang2015salicon} with existing object annotations. This process is more challenging and nuanced than one might initially expect; MS-COCO labels only cover specific object categories, and vary significantly in the precision with which objects are segmented.%, and include many examples that are over-segmented or under-segmented.

With that said, we propose a set of methods that prune an initial set of 10k labeled images based on a careful choice of formal criteria for inclusion/rejection. Images and/or labels that remain are assigned a relative ranking based on a simulated gaze tracking process~\cite{jiang2015salicon} (see Fig.~\ref{fig:dataset}). While there are significant differences between manual choice of salient objects and simulated gaze data, we demonstrate how rank values based on the latter can be treated to produce rankings that approximate the former. This process is validated in comparing rank-order assignments based on manual selection on an alternative dataset using the same criteria. Moreover, we demonstrate that training of models on the dataset we provide produces more capable models than those trained on PASCAL-SR, even in the case that training uses no images from PASCAL-SR. It has been noted that in training deep learning models, there is considerable robustness to even large amounts label noise~\cite{rolnick2017deep}. With that said, our experimentation shows only a very small deviation in algorithmic assignment of rank when compared against click-based ground truth, and also presents a general end-to-end approach for generating saliency ranking data.% suitable for crowdsourcing.

The contribution of this paper extends from the proposed model presented in our prior work~\cite{cvpr18_rank} that 1. Generalizes the problem of salient object detection to salient object ranking which includes inter-observer variability and considers relative rank of salient objects. 2. Presents a new model that predicts salient objects according to the traditional form of this problem, \textcolor{violet}{salient instance detection} and relative ranking.
We extend our prior work in the following respects:
\begin{itemize}
	\item We introduce a novel set of methods that make use of existing gaze or related data paired with object annotations to generate a large scale benchmark dataset for saliency ranking. We also propose metrics suitable to measuring success in a relative object saliency landscape.
	
	\item The discussion and details provided for ground truth generation stands on its own, but also highlights important nuances of the problem and thus provides a roadmap for other similar efforts for generating suitable data for training and testing. The more extensive analysis of metrics also sheds further light on their suitability for this problem and examines this space in more detail than~\cite{cvpr18_rank}.
	
	\item  The process for generating the dataset is validated in comparing rank-wise assignments made to an alternate dataset using the criteria we propose for algorithmic production of COCO-SalRank. Validation comes from rank order agreement to labels in PASCAL-SR which are produced by a process of manual selection by human participants. This implies a larger dataset in COCO-SalRank suitable for both training and evaluation. Moreover, we demonstrate that training of models on the dataset we provide produces more capable models than those trained on PASCAL-SR when tested on either of the datasets.
	
	\item We provide new state-of-the-art baseline scores for the saliency ranking problem on PASCAL-SR and the proposed dataset, while also providing a corpus of data and code to the community that provides significant value for training and evaluation for a relatively nascent problem domain.
\end{itemize}
\section{Background}\label{sec:background}
\subsection{Salient Object Detection:}
Convolutional Neural Networks (CNNs) have raised the bar in performance for many problems in computer vision including salient object detection. CNN based models are able to extract more representative and complex features than hand crafted features used in less contemporary work \cite{li2013saliency,yan2013hierarchical,kim2014salient} which has promoted widespread adoption.

Some CNN based methods exploit superpixel and object region proposals to achieve accurate salient object detection \cite{hu2017deep,li2016deep,lee2016deep,li2016deepsaliency,zhao2015saliency,li2015visual}. Such methods follow a multi-branch architecture where a CNN is used to extract semantic information across different levels of abstraction to generate an initial saliency prediction. Subsequently, new branches are added to obtain superpixels or object region proposals, which are used to improve precision of the predictions.

As an alternative to superpixels and object region proposals, other methods~\cite{luo2017non,hou2016deeply,amulet_2017} predict saliency per-pixel by aggregating multi-level features. Luo et al. \cite{luo2017non} integrate local and global features through a CNN that is structured as a multi-resolution grid. Hou et al.~\cite{hou2016deeply} implement stage-wise short connections between shallow and deeper feature maps for more precise detection and inferred the final saliency map considering only middle layer features. Zhang et al.~\cite{amulet_2017} combine multi-level features as cues to generate and recursively fine-tune multi-resolution saliency maps which are refined by boundary preserving refinement blocks and then fused to produce final predictions.

Other methods~\cite{liu2016dhsnet,wang2016saliency,ucf_2017,islam2017salient} use an end-to-end encoder-decoder architecture that produces an initial coarse saliency map and then refines it stage-by-stage to provide better localization of salient objects. Liu and Han \cite{liu2016dhsnet} propose a network that combines local contextual information step-by-step with a coarse saliency map. Wang et al. \cite{wang2016saliency} propose a recurrent fully convolutional network for saliency detection that includes priors to correct initial saliency detection errors. Zhang et al.~\cite{ucf_2017} incorporate a reformulated dropout after specific convolutional layers to quantify uncertainty in the convolutional features, and a new upsampling method to reduce artifacts of deconvolution which results in a better boundary for salient object detection.

In contrast to the above described approaches, we achieve spatial precision through stage-wise refinement by applying novel mechanisms to control information flow through the network while also importantly including a \emph{stacking} strategy that implicitly carries the information necessary to determine relative saliency.
%\vspace{-0.1cm}
\subsection{Universal Saliency Detection Benchmarks}
\textcolor{violet}{There has long been growing interest in cognitive science disciplines to understand where people look and direct their gaze while interacting with complex indoor or outdoor scenes~\cite{bruce2015computational}. This can be examined by direct measurement of gaze, or by manual selection of important regions, with the latter presenting a situation less prone to low-level biases including centre-bias and those derived from oculomotor constraints~\cite{bruce2015computational}}. Predictive models that address both of these processes have been defined as visual saliency detection. However, in the selection task (as with free-viewing gaze), there is no universal agreement on what constitutes a salient object when opinions are elicited from multiple observers. Until very recently~\cite{cvpr18_rank,Islam2018SemanticsMS}, the literature has failed to acknowledge this nuance of salient object detection. A large number of proposals have been made on methods for predicting salient targets; these studies focus on a binary prediction of a \emph{universal saliency} label that considers \textcolor{violet}{salient objects}~\cite{amulet_2017,ucf_2017,he2017delving,hou2016deeply,wang2016saliency,wang2017stagewise,hu2017deep,li2016deep,lee2016deep,liu2016dhsnet,zhao2015saliency,li2015visual}. \textcolor{violet}{Other studies~\cite{jia2015adaptive,najibi2017towards,zhang2016unconstrained} detect salient objects with multiple bounding boxes but do not consider the variability that may exist across humans in deciding what is salient}. In deference to the apparently deeper problem definition that may be attached to salient object detection, our work~\cite{cvpr18_rank} extended the traditional problem to consider salient object ranking. However, given that this implies a distinct problem domain, there are limitations on what data is currently available for training and/or testing models and the extent of benchmarking that has been carried out.

Several saliency detection~\cite{yan2013hierarchical,shi2013pisa,yang2013saliency,li2014secrets,jiang2015salicon} or eye tracking~\cite{jiang2015salicon} datasets have been \textcolor{violet}{created} and shared with the community to promote saliency research. The majority of these datasets share common features in providing ground-truth annotation as a binary mask (background vs salient object) assigned to each image that is based on one observer, or subject to a threshold when a few opinions are present. Currently, only the Pascal-S dataset is widely available for addressing the saliency ranking problem (with suitable post-processing as in PASCAL-SR) as it provides ground-truth corresponding to multiple observer's agreement across 12 observers. In order to progress further in addressing this problem, there is a dire need for larger-scale datasets to provide a stronger set of data for training and evaluating models. Pascal-S provides ground-truth that implicitly captures relative salience, but only for $\approx$ 1000 examples. The current emphasis on, and success of deep learning architectures generally requires large-scale datasets that implies an even greater need for alternative datasets and suitable benchmark results and metrics. Moreover, the array of current saliency detection datasets is heavily entrenched in the traditional problem definition and associated experimental analysis. We therefore seek to allow research efforts on this problem to rapidly progress, in ensuring availability of large-scale data that serves to improve saliency ranking performance among models, and to allow for immediate adoption of solutions that explore new research directions in assessing \emph{relative saliency} as well as complementing existing datasets and the historical legacy of work in this area.

Obtaining ground-truth by manual labeling is crucial for computer vision applications. Amazon Mechanical Turk (AMT) has been used extensively for labeling large-scale datasets in distributing the labeling task among many human annotators. The crowdsourcing approach has been directed to different tasks that derive desired output labels from human annotation (e.g. assigning a category label, segmenting objects, providing bounding boxes). In this paper, we propose a novel approach to provide a benchmark dataset which involves a hybrid of algorithmic coalescence of multiple disparate labels from common datasets (MSCOCO~\cite{lin2014microsoft}), with optional refinement by a human as a secondary stage. This approach is validated by comparing rank-order assignments with the smaller extant alternative dataset that has exact labels. In this light, this produces not only a dataset, but also algorithmic means for converting other suitable datasets in a similar fashion provided instance level segmentation and a guiding signal such as gaze is present.

\begin{figure*}[t]
	\centering
	\includegraphics[width=0.99\textwidth]{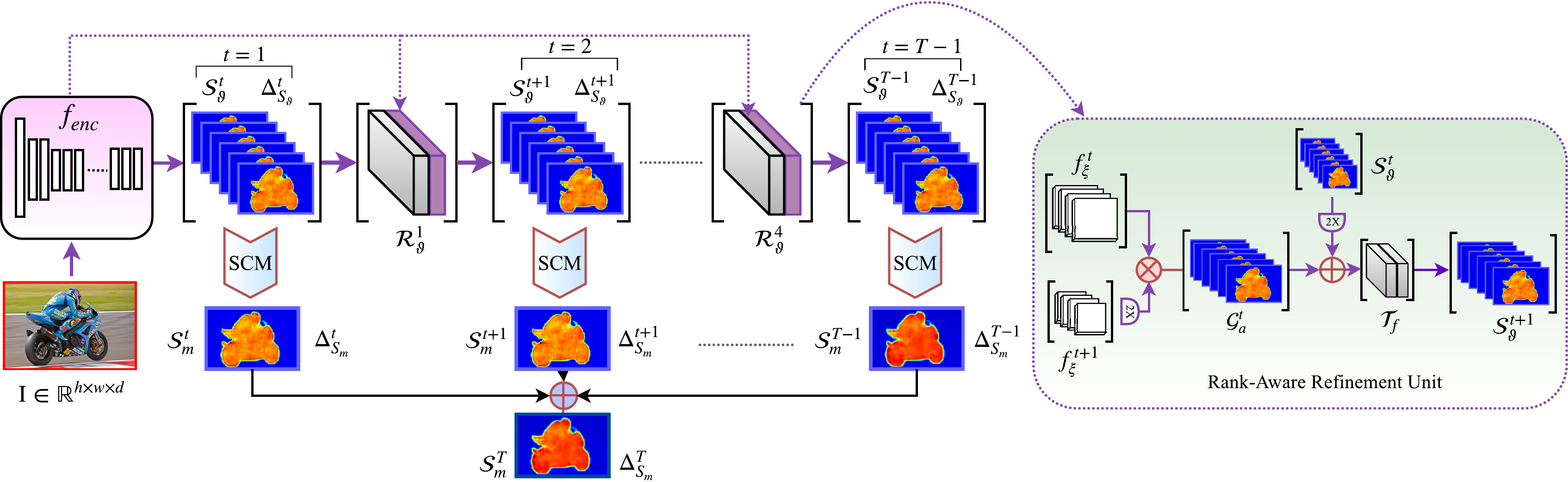}
	\vspace{-0.3cm}
	\caption{Illustration of our proposed network architecture. In the encoder network \textcolor{violet}{($f_{enc}$)}, the input image is processed with a feed-forward encoder to generate a coarse nested relative salience stack ($\mathcal{S}_\vartheta^t$). We append a Stacked Convolutional Module (SCM) on top of $\mathcal{S}_\vartheta^t$ to obtain a coarse saliency map $\mathcal{S}_m^t$. Then, a stage-wise refinement network, comprised of rank-aware refinement units \textcolor{violet}{($\mathcal{R}_\vartheta^1, \mathcal{R}_\vartheta^2,..., \mathcal{R}_\vartheta^4 $)} (dotted box in the figure), successively refines each preceding NRSS ($\mathcal{S}_\vartheta^t$) and produces a refined NRSS ($\mathcal{S}_\vartheta^{t+1}$). A fusion layer combines predictions from all stages to generate the final saliency map ($\mathcal{S}_m^T$). We provide supervision ($\Delta_{S_{\vartheta}}^t$, $\Delta_{S_m}^t$) at the outputs ($\mathcal{S}_\vartheta^t$, $\mathcal{S}_m^t$) of each refinement stage. The architecture based on iterative refinement of a stacked representation is capable of effectively detecting multiple salient objects and their rank.}
	\label{fig:network}
	\vspace{-0.2cm}
\end{figure*}
\textcolor{violet}{The groundwork for some of what is presented in this paper appeared previously~\cite{cvpr18_rank}, in which we introduced the saliency ranking problem along with an effective deep learning solution (RSDNet~\cite{cvpr18_rank})}. This presents a deeper problem than traditional salient object detection. In this work, we specifically emphasize promoting rapid progress in this domain by providing new state-of-the-art baseline scores, a new benchmark dataset and highlight guidelines for data creation and nuances critical to producing a dataset of value to addressing the problem. This is also accompanied by in depth analysis of datasets, characteristics of saliency ranking, and considerations going forward. This presents another significant contribution that will provide stronger capabilities for models and guidance on model success advancing progress in the field related to this problem. 
\section{Proposed Network Architecture}\label{sec:approach}
\textcolor{black}{We propose a new end-to-end framework for solving the problem of detecting multiple salient instances and ranking the instances according to their degree of saliency}. Our proposed salient object detection network is inspired by the success of convolution-deconvolution pipelines~\cite{noh2015learning,liu2016dhsnet,islam2017gated} that include a feed-forward network for initial coarse-level prediction. Then, we provide a stage-wise refinement mechanism over which predictions of finer structures are gradually restored. Fig.~\ref{fig:network} shows the overall architecture of our proposed network. The encoder stage serves as a feature extractor that transforms the input image to a rich feature representation, while the refinement stages attempt to recover lost contextual information to yield accurate predictions and ranking. We begin by describing how the initial coarse saliency map is generated in section~\ref{feed-forward-network}. This is followed by a detailed description of the stage-wise refinement network, and multi-stage saliency map fusion in sections~\ref{stage-wise-refinement} and section~\ref{fusion} respectively.

\subsection{Feed-forward Network for Coarse Prediction}
\label{feed-forward-network}
Recent feed-forward deep learning models applied to high-level vision tasks (e.g. image classification~\cite{He2015,simonyan2014very}, object detection~\cite{ren2015faster}) employ a cascade comprised of repeated convolution stages followed by spatial pooling. Down-sampling by pooling allows the model to achieve a highly detailed semantic feature representation with relatively poor spatial resolution at the deepest stage of encoding, also marked by spatial coverage of filters that is much larger in extent. The loss of spatial resolution is not problematic for recognition problems; however, pixel-wise labeling tasks (e.g. semantic segmentation, salient object detection) require pixel-precise information to produce accurate predictions. Thus, we choose Resnet-101~\cite{He2015} as our encoder network (fundamental building block) due to its superior performance in classification and segmentation tasks. Following prior works on pixel-wise labeling~\cite{chen2016deeplab,islam2017gated}, we use the dilated ResNet-101~\cite{chen2016deeplab} to balance the semantic context and fine details, resulting in an output feature map reduced by a factor of 8. More specifically, given an input image $I\in\mathbb{R}^{h\times w\times d}$, our encoder network produces a feature map of size  $\floor*{\frac{h}{8}, \frac{w}{8}} $. To augment the backbone of the encoder network with a top-down refinement network, we first attach one extra convolution layer with $3\times 3$ kernel and \textcolor{violet}{$N$ channels ($N$ denotes total number of individual observers involved in the labeling process)} to obtain a \textit{Nested Relative Salience Stack} (NRSS). Then, we append a \textit{Stacked Convolutional Module} (SCM) to compute the coarse level saliency score for each pixel. \textcolor{violet}{The SCM consists of three convolutional layers for generating the desired saliency map. The initial convolutional layer has 6 channels with a $3 \times 3$ kernel, followed by two convolutional layers having 3 channels with $3 \times 3$ kernel and one channel with $1 \times 1$ kernel respectively. Each of the channels in the SCM learns a soft weight for each spatial location of the nested relative salience stack in order to label pixels based on confidence that they belong to a salient object. The described operations can be expressed as:}  %Moreover, we utilize atrous pyramid pooling~\cite{chen2016deeplab} to gather more global contextual information.
\begin{gather}
\mathcal{S}_\vartheta^t = \mathcal{C}_{3\times3}(f_{enc}(I;\mathcal{W});\Theta), \hspace{0.2cm} \mathcal{S}_m^t = \partial(\mathcal{S}_\vartheta^t)
\end{gather}
where \textit{I} is the input image and ($\mathcal{W}, \Theta$) denote the parameters of the convolution $\mathcal{C}$. $\mathcal{S}_\vartheta^t$ is the coarse level NRSS for stage \textit{t} that encapsulates different degrees of saliency for each pixel (akin to a prediction of the proportion of observers that might agree an object is salient), $\mathcal{S}_m^t$ refers to the coarse level saliency map, and \textcolor{violet}{$\partial$ refers to SCM. $f_{enc}(.)$ denotes} the output feature map generated by the encoder network. \textcolor{violet}{Note that our encoder network might be replaced with any alternative baseline network} and we have considered a few such choices in our experiments section.
\subsection{Stage-wise Refinement Network}\label{stage-wise-refinement}
Most existing works~\cite{liu2016dhsnet, wang2017stagewise, amulet_2017, hou2016deeply} that have shown success for salient object detection typically share a common structure of stage-wise decoding to recover per-pixel categorization. Although the deepest stage of an encoder has the richest possible feature representation, relying only on convolution and unpooling at the decoding stages to recover lost information may degrade the quality of predictions~\cite{islam2017gated}. So, the spatial resolution that is lost at the deepest layer may be gradually recovered from earlier representations. This intuition appears in proposed refinement based models that include skip connections~\cite{long2015fully,islam2017gated,amulet_2017, hou2016deeply} between encoder and decoder layers. However, how to effectively combine local and global contextual information remains an area deserving further analysis. Inspired by the success of refinement based approaches~\cite{long2015fully, islam2017label,islam2017gated,Islam2018arxiv}, we propose a multi-stage fusion based refinement network to recover lost contextual information in the decoding stage by combining an initial coarse representation with finer features represented at earlier layers. The refinement network is comprised of successive stages of rank-aware refinement units that attempt to recover missing spatial details in each stage of refinement and also preserve the relative rank order of salient objects. Each stage refinement unit takes the preceding NRSS with earlier finer scale representations as inputs and carries out a sequence of operations to generate a refined NRSS that contributes to obtain a refined saliency map. Note that refining the hierarchical NRSS implies that the refinement unit is leveraging the degree of agreement at different levels of SCMs to iteratively improve confidence in relative rank and overall saliency. As a final stage, refined saliency maps generated by the SCMs are fused to obtain the overall saliency map.

\subsubsection{Rank-Aware Refinement Unit}
Previous saliency detection networks~\cite{wang2017stagewise,liu2016dhsnet} proposed refinement across different levels by directly integrating representations from earlier features. Following~\cite{islam2017gated}, we integrate gate units in our rank-aware refinement unit that control the information passed forward to filter out the ambiguity relating to figure-ground and salient objects. The initial NRSS ($\mathcal{S}_\vartheta^t$) generated by the feed-forward encoder provides input for the first refinement unit \textcolor{violet}{($\mathcal{R}_\vartheta^1$)}. Note that one can interpret $\mathcal{S}_\vartheta^t$ as the predicted saliency map in the \textcolor{violet}{refinement} process, but our model forces the channel dimension to be the same as the number of participants involved in labeling salient objects. The refinement unit takes the gated feature map $\mathcal{G}_a^t$ generated by the gate unit~\cite{islam2017gated} as a second input. As suggested by~\cite{islam2017gated}, we obtain $\mathcal{G}_a^t$ by combining two consecutive feature maps \textcolor{violet}{($f_\xi^t$ and $f_\xi^{t+1}$ )} from the encoder network (see dotted box in Fig.~\ref{fig:network}). We first upsample the preceding $\mathcal{S}_\vartheta^t$ to double its size. \textcolor{violet}{A transformation function $\mathcal{T}_f$ comprised of a sequence of operations (convolution followed by batch normalization and ReLU) is applied} on upsampled $\mathcal{S}_\vartheta^t$ and $\mathcal{G}_a^t$ to obtain the refined NRSS ($\mathcal{S}_\vartheta^{t+1}$). We then append the \textit{SCM} module on top of $\mathcal{S}_\vartheta^{t+1}$ to generate the refined saliency map $\mathcal{S}_m^{t+1}$. Finally, the predicted $\mathcal{S}_\vartheta^{t+1}$ is fed to the next stage rank-aware refinement unit. Note that, we only forward the NRSS to the next stage, allowing the network to learn contrast between different levels of confidence for salient objects. Unlike other approaches, we apply supervision for both the refined NRSS and the refined saliency map. The procedure for obtaining the refined NRSS and the refined saliency map for all stages is identical. The described operations may be summarized as follows:
\begin{gather}
  \mathcal{S}_\vartheta^{t+1} =  w^b \ast \mathcal{T}_f(\mathcal{G}_a^t, u(\mathcal{S}_\vartheta^t)), \hspace{0.1cm} S_m^{t+1} = w_s^b \ast  \partial(\mathcal{S}_\vartheta^{t+1})
\end{gather}
where $u$ represents the upsample operation; $w^b$ and $w_s^b$ denotes the parameter for the transformation function $\mathcal{T}_f$  and SCM ($\partial$ in the equation) respectively. Note that \textit{t} refers to particular stage of the refinement process.
\subsection{Multi-Stage Saliency Map Fusion}\label{fusion}
Predicted saliency maps at different stages of the refinement units are capable of finding the location of salient regions with increasingly sharper boundaries. Since all the rank-aware refinement units are stacked together on top of each other, the network allows each stage to learn specific features that are of value in the refinement process. These phenomena motivate us to combine different level SCM predictions, since the internal connection between them is not explicitly present in the network structure. To facilitate interaction, we add a fusion layer at the end of network that concatenates the predicted saliency maps of different stages, resulting in a fused feature map $\mathcal{S}_m^{\hat{f}} $. Then, we apply a $1 \times 1$ convolution layer $\Upsilon$ to produce the final predicted saliency map $\mathcal{S}_m^{T}$ of our network. Note that our network has T predictions \textcolor{violet}{(in our case T=6)}, including one fused prediction and T-1 stage-wise predictions. We can write the operations as follows:
\begin{gather}
	\mathcal{S}_m^{\hat{f}} =  \eth(\mathcal{S}_m^{t}, \mathcal{S}_m^{t+1}, .... , \mathcal{S}_m^{T-1}), \hspace{0.1cm} \mathcal{S}_m^{T} = w^f \ast \Upsilon(\mathcal{S}_m^{\hat{f}})	
\end{gather}
 where $\eth$ denotes the cross channel concatenation; $w^f$ is the resultant parameter for obtaining the final prediction.
 %\vspace{-0.3cm}
 \subsection{Stacked Representation of Ground-truth} \label{sec:stack}
 The ground-truth for salient object detection or segmentation contains a set of numbers defining the degree of saliency for each pixel. The traditional way of generating binary masks is by thresholding which implies that there is no notion of relative salience. Since we aim to explicitly model observer agreement, using traditional binary ground-truth masks is unsuitable. To address this problem, we propose to generate a set of stacked ground-truth maps for PASCAL-S dataset that corresponds to different levels of saliency (defined by inter-observer agreement) denoted as PASCAL-SR. Given a ground-truth saliency map $\mathcal{G}_m$, we obtain a stack $\mathcal{G}_{\vartheta}$ of \textit{N} ground-truth maps ($\mathcal{G}_i, \mathcal{G}_{i+1}, ....., \mathcal{G}_N$) where each map $\mathcal{G}_i$ includes a binary indication that at least $i$ observers judged an object to be salient (represented at a per-pixel level). \textit{N} is the number of different participants involved in labeling the salient objects. The stacked ground-truth saliency maps $\mathcal{G}_{\vartheta}$ provides better separation for multiple salient objects (see Eq. \eqref{eq:stack_gt} for illustration) and also naturally acts as the relative rank order that allows the network to learn to focus on degree of salience. It is important to note the nested nature of the stacked ground truth wherein $\mathcal{G}_{i+1} \subseteq \mathcal{G}_{i}$. This is important conceptually as a representation wherein $\mathcal{G}_i = 1 \iff$ exactly $i$ observers agree, results in zeroed layers in the ground truth stack, and large changes to ground truth based on small differences in degree of agreement.
 \begin{equation}\label{eq:stack_gt}
 \mathcal{G}_{\vartheta} =
 \begin{bmatrix} \mathcal{G}_i \\ \includegraphics[width=0.05\textwidth]{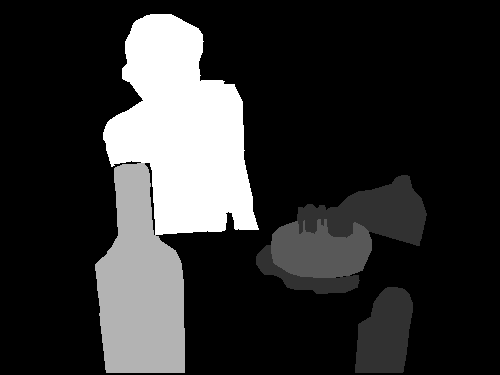} \end{bmatrix}
 \begin{bmatrix} \mathcal{G}_{i+1} \\ \includegraphics[width=0.05\textwidth]{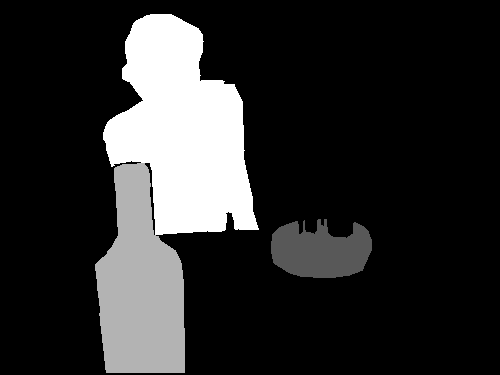} \end{bmatrix}
 \begin{bmatrix} \mathcal{G}_{i+2} \\ \includegraphics[width=0.05\textwidth]{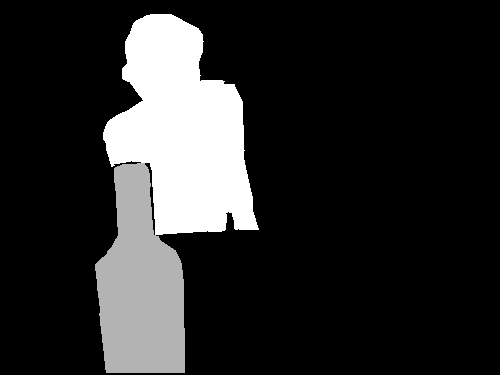} \end{bmatrix}
 \begin{bmatrix} \\... \\\\ \end{bmatrix}
 \begin{bmatrix} \mathcal{G}_{N} \\ \includegraphics[width=0.05\textwidth]{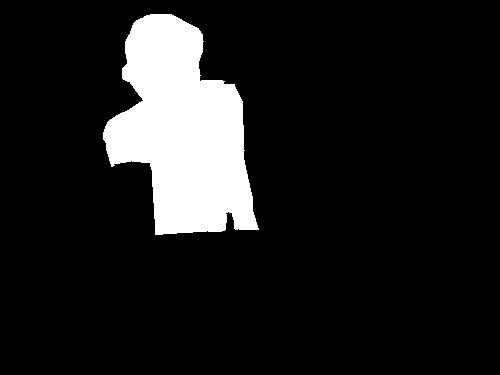} \end{bmatrix}
% \vspace{-0.5cm}
 \end{equation}
 %%%%%%%%%%%%%
 %\vspace{-0.2cm}
\subsection{Training the Network}
Our proposed network produces a sequence of nested relative salience stacks (NRSS) and saliency maps at each stage of refinement; however, we are principally interested in the final fused saliency map. Each stage of the network is encouraged to repeatedly produce NRSS and a saliency map with increasingly finer details by leveraging preceding NRSS representations. We apply an auxiliary loss at the output of each refinement stage along with an overall master loss at the end of the network. Both of the losses help the optimization process. In more specific terms, let $I\in\mathbb{R}^{h\times w\times 3}$ be a training image with ground-truth saliency map $ \mathcal{G}_m\in\mathbb{R}^{h\times w}$. As described in Sec.~\ref{sec:stack}, we generate a stack of ground-truth saliency maps $\mathcal{G}_{\vartheta}\in\mathbb{R}^{h\times w\times N}$. To apply supervision on the NRSS ($S_\vartheta^t$) and saliency map $S_{m}^t$, we first down-sample $\mathcal{G}_\vartheta$ and $\mathcal{G}_m$ to the size of $S_\vartheta^t$ generated at each stage resulting in $\mathcal{G}_\vartheta^t$ and $\mathcal{G}_m^t$. Then, at each refinement stage we define pixel-wise euclidean loss $\Delta_{S_\vartheta}^t$ and $\Delta_{S_m}^t$ to measure the difference between ($S_\vartheta^t, \mathcal{G}_\vartheta^t$) and ($S_{m}^t, \mathcal{G}_{m}^t$) respectively. We can summarize these operations as:
{
\begin{gather}
\Delta_{S_{\vartheta}}^t(W) = \frac{1}{2dN} \sum_{i=1}^{d}  \sum_{z=1}^{N} (x_i(z)-y_i(z))^2 \nonumber\\
\Delta_{S_m}^t(W) = \frac{1}{2d} \sum_{i=1}^{d} (x_i-y_i)^2 \nonumber \\
L_{aux}^t (W)  = \Delta_{S_{\vartheta}}^t + \Delta_{S_m}^t
\end{gather}
}
\noindent where $x \in {\rm I\!R}^d$ and  $y \in {\rm I\!R}^d$ ($d$ denotes the spatial resolution) are the vectorized ground-truth and predicted saliency map. $x_i$ and $y_i$ refer to a particular pixel of $S_\vartheta^t$ and $G_\vartheta^t$ respectively. \textit{W} denotes the parameters of whole network and $N$ refers to total number of ground-truth slices ($\text{N=12}$ in PASCAL-SR case and $\text{N=5}$ in COCO-SalRank dataset described in Sec.~\ref{sec:cocosalrank}). The final loss function of the network combining master and auxiliary losses can be written as:

{{\footnotesize
\begin{gather}
L_{final}(W)= L_{mas}(W) + \sum_{t=1}^{T-1} \lambda_t L_{aux}^t (W)
\end{gather}
}}
where $L_{mas}(W)$ refers to the Euclidean loss function computed on the final predicted saliency map $\mathcal{S}_m^{T}$. We set $\lambda_t$ to 1 for all stages to balance the loss, which remains continuously differentiable. Each stage of prediction contains information related to two predictions, allowing our network to propagate supervised information from deep layers. This also begins with aligning the weights with the initial coarse representation,
leading to a coarse-to-fine learning process. The fused prediction generally appears much better than other stage-wise predictions since it contains the aggregated information from all the refinement stages. For saliency inference, we can simply feed an image of arbitrary size to the network and use the fused prediction as our final saliency map.
%%%%%%%%%%%%%%%%%%%%%%%%%
\section{COCO-SalRank Dataset}\label{sec:cocosalrank}
Given the stated objective of scaling up from the ranking benchmark derived from PASCAL-SR, we aim to construct a larger dataset suitable for saliency ranking analysis for use by the computer vision community to promote saliency ranking research. Designing a large-scale dataset requires a large number of decisions including data collection, processing, and the annotation protocol. Our choices were driven by the end goal of enabling immediate progress in the field of saliency ranking and allowing deeper exploration of relative saliency. The description in what follows may be viewed as having two contributions: i. A larger pool of data for training and benchmarks, which produces demonstrably better results for the PASCAL-SR derived ranking ii. A view into considerations relevant to producing additional data suitable to measuring the success of solutions to assigning relative rank according to saliency. With respect to this latter consideration, the approach and analysis reveals a number of nuances related to the problem space that are important to the content of this work, but also to future data-centric contributions for relative salience. \textcolor{violet}{This might be applied e.g. to the very recent SOC dataset \cite{SOCpaper} as an extension given its high quality annotations, size and variety of images. The SOC dataset in particular (released after submission of this manuscript in its initial form) may have more fine-grained pixel level labeling. In comparison, our dataset relies on the original MS-COCO labels that are somewhat coarse in comparison. With that said, at this time the data presented here is the largest scale and highest quality dataset for ranking, and we have provided a strong roadmap for extension to other datasets such as this one.}
\subsection{Description of the COCO-SalRank dataset}\label{des}
One rule of thumb for constructing a dataset perhaps, is to first analyze existing datasets for the same task. There exist a significant number of saliency detection datasets but the majority provide ground-truth in binary notation. Since saliency ranking requires ground-truth in the form of multiple observers' agreement, a primary objective is to arrive at a dataset that provides a faithful rank order of the salient objects. To do so, we at least require images with several distinct measures of what may be salient. Given the relatedness to human allocation of gaze, it is natural to consider what value may be derived from considering fixation maps across observers in conjunction with cases where instance-wise label maps are also present. Specifically, there is the opportunity to leverage \textcolor{black}{fixation maps} to assign rankings among category items present within instance-wise label maps. We therefore make the careful choice of MS COCO images that include multiple simulated moused-based fixation annotations from the training and validation data of the SALICON dataset~\cite{jiang2015salicon} to construct our proposed dataset. We obtain instance-wise mask annotations from the MS-COCO dataset as SALICON images are chosen from MS-COCO.

Directly combining these two data sources to achieve the desired objective is a more significant challenge than one might expect on the surface for reasons that are highlighted throughout this paper. \textcolor{violet}{Moreover, we have already discussed significant differences between gaze data and mouse-based assessments of salient object for ranking including low-level biases and differences in underlying cognitive processes}.  For this reason, we also include careful evaluation on data where both gaze and click based selections are present to validate that the end result comes very close in approximating mouse-based ranking data. Given the set of images with instance-wise labels and \textcolor{black}{mouse-based fixation maps}, we propose a novel approach to provide saliency ranking ground-truth. Initially we create a \emph{noisy} version (version \rom{1}) of the proposed dataset (COCO-SalRank) that includes 7047 training images and 3363 test images. Subsequently, we produce a refined \emph{clean} version (version \rom{2}) of annotations with 3052 training and 1381 test images. Note that in deriving the clean dataset, the cleaned data (version \rom{2}) is also manually checked  to remove obvious outliers, and to ensure labeling fidelity. The total number of images removed manually based on visual inspection is 605. The following presents details for deriving these datasets, including data refinement that have been considered in creating the COCO-SalRank dataset:
\begin{itemize}
	\setlength{\itemindent}{0em} 
	\item \textbf{Number of Instances:} We restrict the total number of instances in an image to maximum five.
	\item \textbf{Ranking Ties:} No instances are assigned tied values due to differences in relative saliency within labeled instance regions.
\end{itemize}

Considering the above-mentioned factors, we provide two different sets of ground-truth data that may each carry value in model training or evaluation. The justification for this latter statement is borne out in analysis of results.
\subsection{Ground Truth Annotation}
To obtain the saliency ranking ground-truth, as alluded to earlier, we propose a novel approach that assigns a rank order to salient objects in an image given instance-wise segmentations and associated stimulated \textcolor{black}{mouse-based fixation maps}. The instance-wise labeling in MS-COCO is not consistent with the fixation maps provided by SALICON, so we can not directly use these two sets of annotations in order to generate the saliency ranking annotation. There are many instances which are labeled in the instance map but are not a point of focus in the fixation map. Similarly, some instances have reasonable fixation density but are not labeled as an instance in the set of masks. Other challenges include over or under-segmented images, unlabeled non-class objects, and that explicit ranking from manual selections are unavailable for this data.
%%%%%%% Algorithm for ranking annotation
%%%%%%%%%%%%%%%%%%%%%%%%%%%%%%%%
\begin{figure*}[ht!]
	%\vspace{-0.7cm}	
	\begin{equation}
	%\mathcal{G}_{\vartheta} =
	\begin{bmatrix} \text{image \& mask}\\ \fcolorbox{red}{red} {\includegraphics[width=0.10\textwidth]{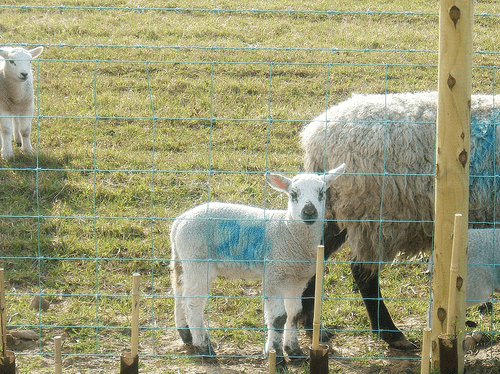}} \hspace{0.05cm}  \fcolorbox{red}{red} {\includegraphics[width=0.10\textwidth]{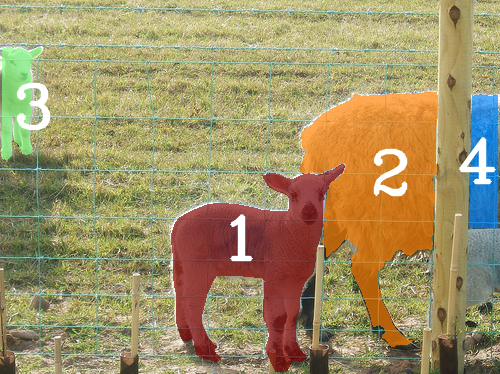}}
	\end{bmatrix}
	\begin{bmatrix} \text{fixation map} \\ \fcolorbox{red}{red} {\includegraphics[width=0.10\textwidth]{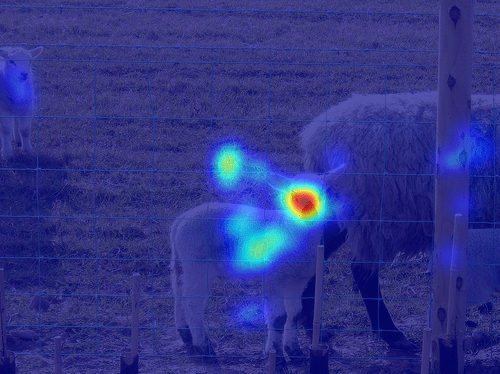}} \hspace{0.05cm}  \fcolorbox{red}{red} {\includegraphics[width=0.10\textwidth]{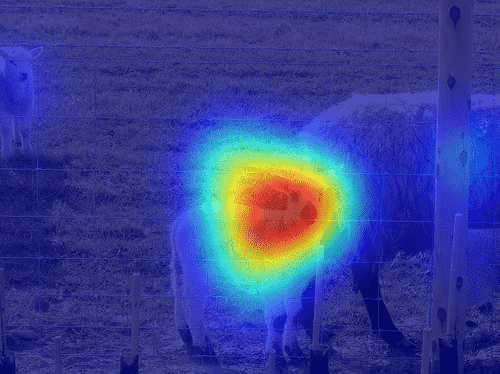}}
	\end{bmatrix}
	\begin{bmatrix} \text{blurring effect} \\ \fcolorbox{red}{red} {\includegraphics[width=0.10\textwidth, height=1.365cm]{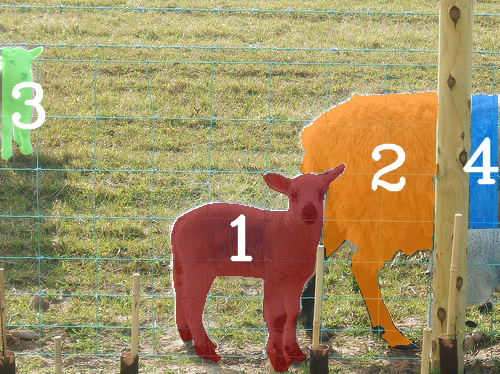}} \hspace{0.05cm}  \fcolorbox{red}{red} {\includegraphics[width=0.10\textwidth]{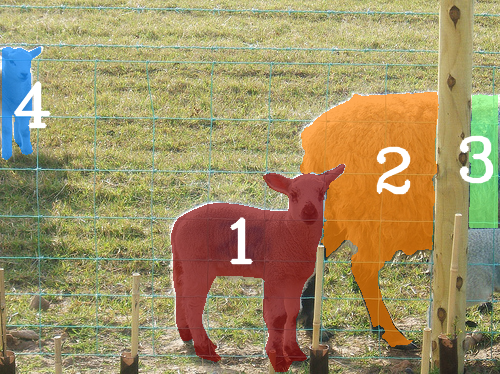}}
	\end{bmatrix}
	\begin{bmatrix} \text{power effect} \\ \fcolorbox{red}{red} {\includegraphics[width=0.10\textwidth]{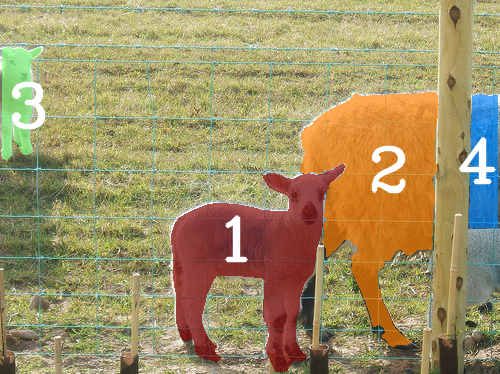}} \hspace{0.05cm}  \fcolorbox{red}{red} {\includegraphics[width=0.10\textwidth]{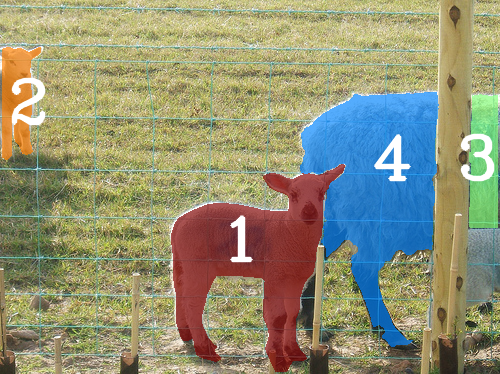}}
	\end{bmatrix} \nonumber
	\begin{matrix}  \\
	%{\includegraphics[width=0.02\textwidth, height = 1.6cm]{figures/jet-v.pdf}}
	\end{matrix}
	%\vspace{-0.5cm}		
	\end{equation}
	
	\caption{Sets from left to right: Input image and GT rank, \textcolor{black}{simulated mouse-based fixation maps} blurred with different Gaussian filters, predicted rank that corresponds to fixation maps in the previous set ($\alpha = 0.3$), predicted rank that corresponds to two different $\alpha$ ($\sigma = 10.5, \mu = 80$). \textcolor{violet}{Relative rank is indicated by the assigned color and number on each salient instance.} }
	\label{fig:blur}
	\vspace{-0.3cm}	
\end{figure*}
To overcome these limitations, we have arrived at a data refinement pipeline through experimentation that is shown to produce faithful rankings when measured against alternate smaller-scale data where gaze, instance labels, and explicit selection are all present. The methods that address this challenge are described in what follows. We first apply Gaussian blurring on the \textcolor{black}{fixation} locations to obtain the new fixation map $\mathcal{F}_i^\prime $. This is a crucial step in our approach since fixation locations provided by SALICON are generated by mouse tracking instead of a traditional eye tracker which implies a less diffuse distribution on the focus of attention such that fixation points may not overlap with the corresponding instances (even if they are proximal). In addition, as the labeling in MS-COCO does not capture the border of the instances accurately, blurring the fixation locations is especially important to allow density from border fixations to diffuse into the defined mask area. Another important step in our annotation protocol is pruning the original instance mask. As mentioned earlier, regions labeled in instance-wise label maps may not include all salient objects. Instance masks may therefore be pruned based on few carefully chosen criteria. Given the set of new fixation maps $\mathcal{F}^\prime $ and the provided instance-wise maps $\mathcal{I}$, we propose a \textit{Saliency Ranking} algorithm (Algorithm~\ref{alg:ranking}) that generates the ranking ground-truth. Algorithm~\ref{alg:ranking} describes the set of steps to obtain a rank order of salient instances in an image.
\begin{table}[H]
	\vspace{-0.4cm}
	\begin{algorithm}[H]
		%\scriptsize
		\caption{Saliency Ranking}\label{rank}
		\begin{algorithmic}[1]
			
			\Function{SalRank}{$\mathcal{I},\mathcal{F}, \sigma, \mu, \xi, \ell, \gamma$} \Comment{instance maps $\mathcal{I}$, fixation $\mathcal{F}$}
			
			\For{\texttt{each instance map $\mathcal{I}_i\in \mathcal{I}$}}
			
			%\State Rank list, $\mathbb{R}$
			
			\State $\mathcal{F}_i^\prime $  = \textit{Gaussian} ($\sigma, \mu, \mathcal{F}_i $) \hspace{0.3cm} overlap, $\vartheta = \mathcal{F}_i^\prime \times \mathcal{I}_i$
			
			%%% Calculate the ranking score
			%\State overlap, $\vartheta = \mathcal{F}_i^\prime \times \mathcal{I}_i$
			\For{\texttt{each instance $ \chi \in \mathcal{I}_i$}}
			\State score, $\mathbb{R}_\chi = \frac{\sum \vartheta(\chi)}{\sqrt[\epsilon]{\text{size}(\chi)}}$  \hspace{0.5cm} (Rank list, $\mathbb{R}$) %\Comment {$\epsilon = 0.3$}
			%\State $\mathbb{R}$.append($\mathbb{R}_\chi)$
			\EndFor
			
			\State $\mathcal{I}_i^\prime$ = \textbf{\color{red}Prune}($\mathcal{I}_i, \mathcal{F}_i^\prime, \mathbb{R}, \alpha_1, \alpha_2 )$
			
			\State overlap, $\vartheta^\prime$ = $\mathcal{I}_i^\prime \times \mathcal{F}_i$ \hspace{0.3cm}total instances, $\rho$ = \textit{unique}($\mathcal{I}_i^\prime$)
			%\State total instances, $\rho$ = \textit{unique}($\mathcal{I}_i^\prime$)
			\If{$\rho$ $> \xi$ \textbf{or}  $\frac{\sum(\vartheta^\prime)}{\sum(\mathcal{F}_i^\prime)}$ $<$ $\ell$ \textbf{or} $\sum(\mathcal{I}_i^\prime \neq 0)   > \gamma$  }
			\State ignore instance map $\mathcal{I}_i$
			\EndIf
			
			\EndFor

			\EndFunction
		\end{algorithmic}
		\label{alg:ranking}
		
	\end{algorithm}
	\vspace{-0.5cm}
\end{table}

First, we calculate overlap $\vartheta$ between the new fixation map $\mathcal{F}^\prime $ and the instance-wise map $\mathcal{I}_i$ to remove non-salient instances. Then, we generate a ranking score $\mathbb{R}_\chi$ for each salient instance $\chi$ by dividing the total saliency captured by $\chi$ in $\vartheta$ to the size of the instance raised to the power $\epsilon$. The \textit{prune} function (see Algorithm~\ref{alg:rankin2g} ) takes the newly generated fixation map $\mathcal{F}^\prime $, instance map $\mathcal{I}_i$, and two parameters ($\alpha_1 \& \alpha_2$) as input, resulting in a pruned instance mask, $\mathcal{I}_i^\prime$. The prune function focuses on removing instances with the following conditions: (1) If the size of a particular instance is greater than a certain threshold $\alpha_1$ (\textcolor{violet}{generally due to under-segmentation or close-up shot scenes}) or (2) The rank score of an instance $\mathbb{R}_\chi$ is less than $\alpha_2$ (generally due to receiving very little attention). Given the pruned instance map $\mathcal{I}_i^\prime$, we again calculate the overlap $\vartheta^\prime$ between $\mathcal{F}^\prime $ and $\mathcal{I}_i^\prime$ in order to disqualify the pruned instance maps $\mathcal{I}^\prime$. We apply the following conditions to filter instance maps: (a) The total number of instances is less than $\xi$ (b) The total saliency captured by the pruned instance mask is greater than $\ell$ compared to the fixation map (c) The ratio of background vs salient instances satisfies a certain threshold $\gamma$.
%%%%%%%%%%%%%%%%%%%%%%%%%%%%%%
\begin{table}[H]
	\vspace{-0.4cm}
	\begin{algorithm}[H]
		%	\scriptsize
		\caption{\textcolor{violet}{Prune} Instance Mask }\label{prune}
		\begin{algorithmic}[1]
			
			\Function{\color{red}\textbf{Prune}}{$\mathcal{I}_i, \mathcal{F}_i^\prime, \mathbb{R}, \alpha_1, \alpha_2$}
			\Comment{instance map $\mathcal{I}_i$, fixation $\mathcal{F}_i^\prime$}
			
			\For{\texttt{each instance $ \chi \in \mathcal{I}_i$}}
			
			\If{$ \text{size}(\chi) > \alpha_1 $ or $\mathbb{R}_\chi < \alpha_2$  }
			\State remove instance $\chi$
			\EndIf
			
			%				\If{$\mathbb{R}_\chi \in \mathbb{R} < \alpha_2$ }
			%				\State remove instance $\chi$
			%				\EndIf
			\EndFor
			
			\State return pruned instance mask $\mathcal{I}_i^\prime$
			\EndFunction
		\end{algorithmic}
		\label{alg:rankin2g}
	\end{algorithm}
	\vspace{-0.5cm}
\end{table}

Table \ref{table:thres} demonstrates the set of parameters used to generate both versions of the proposed ground-truth labels. If an instance map meets all of the three conditions, \textcolor{violet}{we assign relative salience to each instance based on their rank score}. Note that we propose two different strategies (relative and absolute) to assign a numeric rank in a range of [0, 255] and this is briefly described in Sec~\ref{sec:salience_assignment}.
\begin{table}[h]
	\vspace{-0.1cm}
	\caption{ The set of parameters used in our process. In version II, we specifically tighten $\ell$ and $\alpha_2$ to obtain more clean and reliable annotations.  }	
	\vspace{-0.3cm}	
	\centering
	\resizebox{0.48\textwidth}{!}{
		\begin{tabular}{ccccccccccccccc}
			\specialrule{1.1pt}{1pt}{1pt}
			\multicolumn{15}{c}{\color{black}COCO-SalRank } \\
			\specialrule{1.2pt}{1pt}{1pt}\	
			
			&\multicolumn{5}{c}{version I (noisy) } && \multicolumn{7}{c}{version II (clean) } \\
			
			$\sigma$ & $\mu$ & $\xi$ & $\ell$ & $\gamma$ & $\alpha_1$ & $\alpha_2$ && $\sigma$ & $\mu$ & $\xi$ & $\ell$ & $\gamma$ & $\alpha_1$ & $\alpha_2$ \\
			
			\cline{1-7} \cline{9-15}
			
			10.5	& 80& 	5& 	\color{blue}0.4& 	0.65& 	0.4	&\color{red} 0.7  &&  10.5	& 80& 	5& 	\color{blue}0.7& 	0.65& 	0.4	& \color{red}0.9\\
			
			\specialrule{1.2pt}{1pt}{1pt}
		\end{tabular}}
		
		\label{table:thres}
		\vspace{-0.4cm}		
\end{table}	
%%%%%%%%%%%%%%%%%%%
%%%%%%%%%%%%%%%%%%%%%%%
\subsection{Dataset Analysis}
We mention \textcolor{violet}{earlier (Sec.~\ref{des})} that our dataset is more suitable for saliency ranking than existing saliency detection datasets. It is evident that it is mandatory to have multi-user agreement to assign a rank score for individual salient instances. In what follows, we discuss a few aspects of the proposed dataset and ranking algorithm, and their justification. 

\noindent \textbf{Effect of Blurring and instance size in ranking? }\\
As mentioned prior \textcolor{violet}{(Sec.~\ref{des})}, we use instance maps from MS-COCO and the \textcolor{black}{mouse-based fixation} locations from SALICON in order to generate ranking labels. For each instance in an instance map, we calculate a score that represents the degree of saliency for that instance. A natural approach to calculate the rank score is to compute the amount of saliency that an instance captures. However, larger objects tend to capture more fixations than smaller objects which implies that larger objects may be implicitly biased towards having a higher rank order compared to smaller ones. Therefore, the size of the object is a key deciding factor in the process of calculating the ranking score for each object. Furthermore, the degree of Gaussian burring of the fixation locations also influences the amount of saliency that an instance may capture and as noted, has a proximity effect in relation to object instances and is also influenced by the undefined boundary region.
%%%%%%%%%%
\begin{figure}[H]
	\vspace{-0.2cm}
	\centering
	\setlength\tabcolsep{0.9pt}
	\def\arraystretch{0.8}
	\resizebox{0.48\textwidth}{!}{
		\begin{tabular}{c}
			
			\begin{tabular}{ c c c c c}
				
				\fcolorbox{red}{red} {\includegraphics[width=0.12\textwidth]{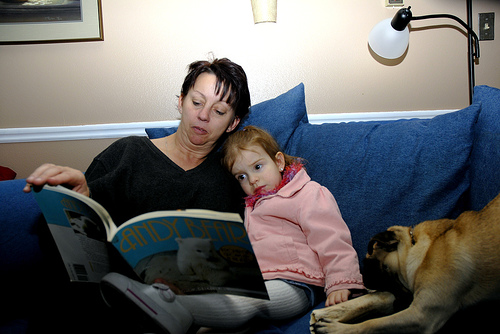}}&
				\fcolorbox{red}{red}{\includegraphics[width=0.12\textwidth]{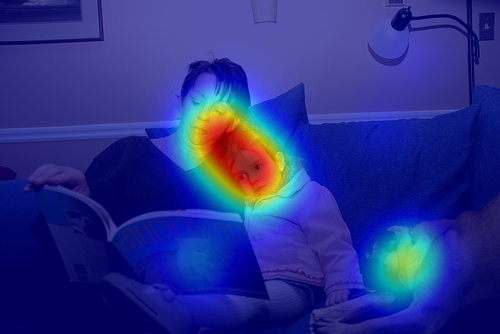}}&
				\fcolorbox{red}{red}{\includegraphics[width=0.12\textwidth]{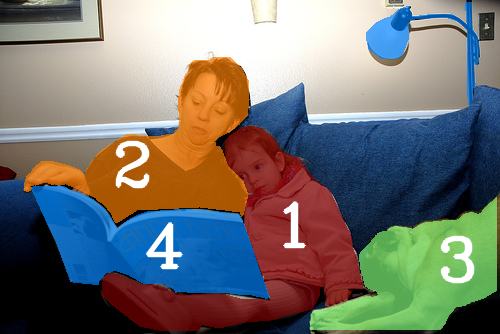}}&
				\fcolorbox{red}{red}{\includegraphics[width=0.12\textwidth]{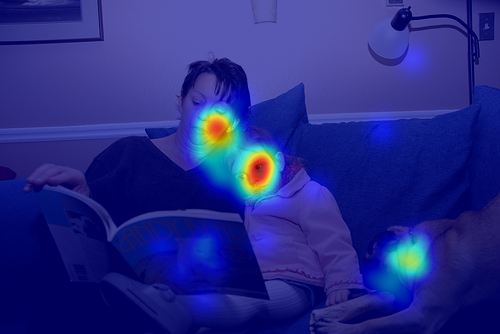}}&
				\fcolorbox{red}{red}{\includegraphics[width=0.12\textwidth]{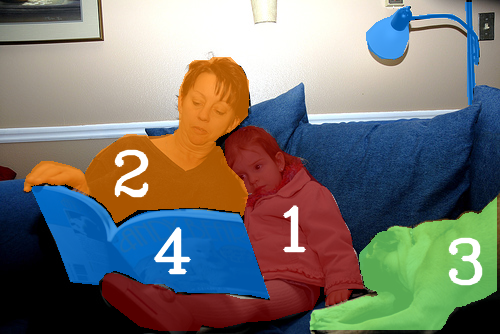}} \\
				
				\fcolorbox{red}{red}{\includegraphics[width=0.12\textwidth]{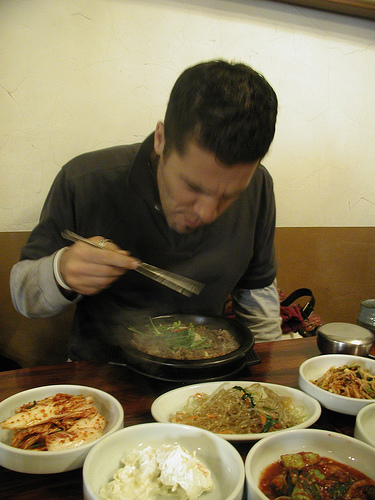}}&
				\fcolorbox{red}{red}{\includegraphics[width=0.12\textwidth]{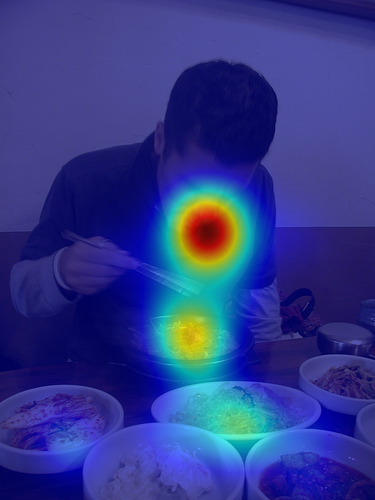}}&
				\fcolorbox{red}{red}{\includegraphics[width=0.12\textwidth]{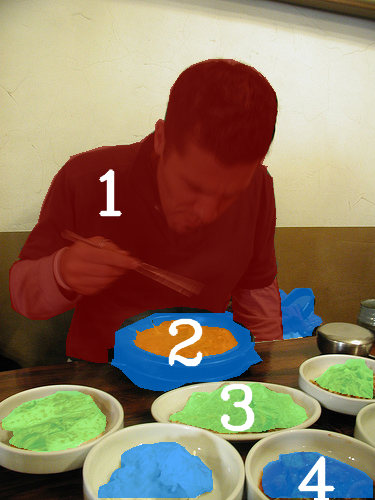}}&
				\fcolorbox{red}{red}{\includegraphics[width=0.12\textwidth]{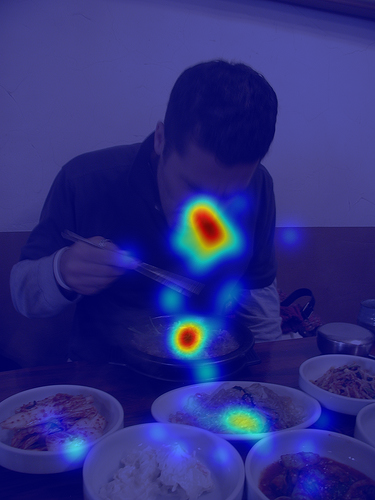}}&
				\fcolorbox{red}{red}{\includegraphics[width=0.12\textwidth]{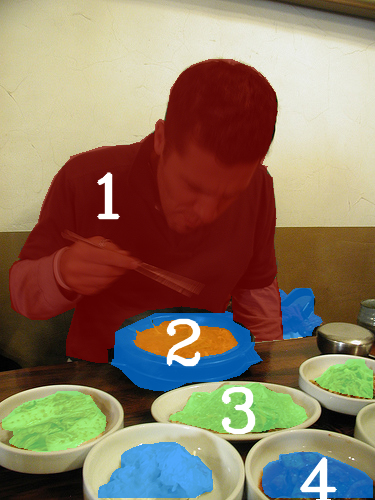}} \\

				image & Fixation GT&Rank GT& Fixation (ours)&  ours\\
				
			\end{tabular}
			
		\end{tabular}}
		\caption{Qualitative comparison for ground-truth annotation using our algorithm for the PASCAL-S dataset. \textcolor{violet}{ Relative rank is indicated by the assigned color and number on each salient instance.}}
		\label{fig:pascals_annot}
		\vspace{-0.2cm}
\end{figure}

In order to examine how the size of each instance and the blurring of fixation locations might sway the rank for each instance, and to investigate the possible ways to integrate these variables in the calculation of the ranking score, we measure the impact of the ranking parameters against the PASCAL-S dataset since it provides both fixation locations and the manually chosen rank of salient objects annotated by multiple observers. In other words, we use the simulated mouse-based fixation locations and the segmentation masks provided by the PASCAL-S dataset to generate our ranking label which is compared against the ranking labels provided by the PASCAL-S dataset (see Fig.~\ref{fig:pascals_annot}). We compute each instance ranking score according to equation \ref{eq:eq1}.
\noindent We first examine the effect of changing $\alpha$ by fixing the Gaussian blurring filter size to $\mu=80$ and the standard deviation $\sigma = 10.5$. We change the value of $\alpha$ and calculate the Salient Object Ranking (SOR)~\cite{cvpr18_rank} score between the predicted and the ground truth ranking label provided by PASCAL-S as shown in Table~\ref{table:blur}.
\begin{table}[H]
	\vspace{-0.2cm}	
	\caption{ The impact of applying different power $\alpha$, filter size $\mu$, standard deviation $\sigma$ on ranking performance. Note that in the right set, we fix $\alpha =0.3$ and use the filter size equivalent to $\mu = \sigma \times 7$ whereas in left set we fix both $\mu = 80,  \sigma= 10.5$ }
	\vspace{-0.2cm}	
	\centering
	\resizebox{0.49\textwidth}{!}{
		\begin{tabular}{c|c|c|c|c|c|c}
			\specialrule{1.1pt}{1pt}{1pt}
			
			\multirow{1}{*}{  \textbf{$\alpha$}}&  1 &		0.6&	0.4	&0.3&	0.2&	0.1   \\
			\hline
			\multirow{1}{*}{ \textbf{SOR}}&  0.76&		0.86&		0.89&	0.90&	0.88&	0.87   \\
			\specialrule{1.1pt}{1pt}{1pt}
			
		\end{tabular}
		
		\begin{tabular}{c|c|c|c|c|c}
			\specialrule{1.1pt}{1pt}{1pt}\
			
			\multirow{1}{*}{ $\sigma$}&  5&		13&	21&	29		&37  \\
			\hline
			\multirow{1}{*}{ \textbf{SOR}}&  0.89&		0.89&		0.88&		0.87&		0.86  \\
			\specialrule{1.1pt}{1pt}{1pt}
			
		\end{tabular}
	}
	
	\label{table:blur}
	\vspace{-0.2cm}	
\end{table}

\noindent Similarly, we examine the Gaussian burring effect by fixing the power $\alpha = 0.3$ and changing the standard deviation of the Gaussian blurring $\sigma$. Note that the filter size corresponding to a specific $\sigma$ in Table \ref{table:blur} is $\mu = \sigma *7$. We found empirically that a Gaussian burring filter size $\mu=80$, the standard deviation $\sigma = 10.5$, and $\alpha = 0.3$ tends to correspond to the set of conditions for which fixations best determine manually chosen rankings (based on SOR score for PASCAL-S). Thus, we apply these values as our threshold while generating ground truth for our proposed COCO-SalRank dataset. Fig. \ref{fig:blur} further demonstrates both of those effects.

\begin{figure} [t]
	%\vspace{-0.1cm}
	\centering
	%\vspace{-0.4cm}
	\setlength\tabcolsep{1.0pt}
	\resizebox{0.49\textwidth}{!}{
		\begin{tabular}{*{2}{c }}
			
			\includegraphics[width=0.24\textwidth]{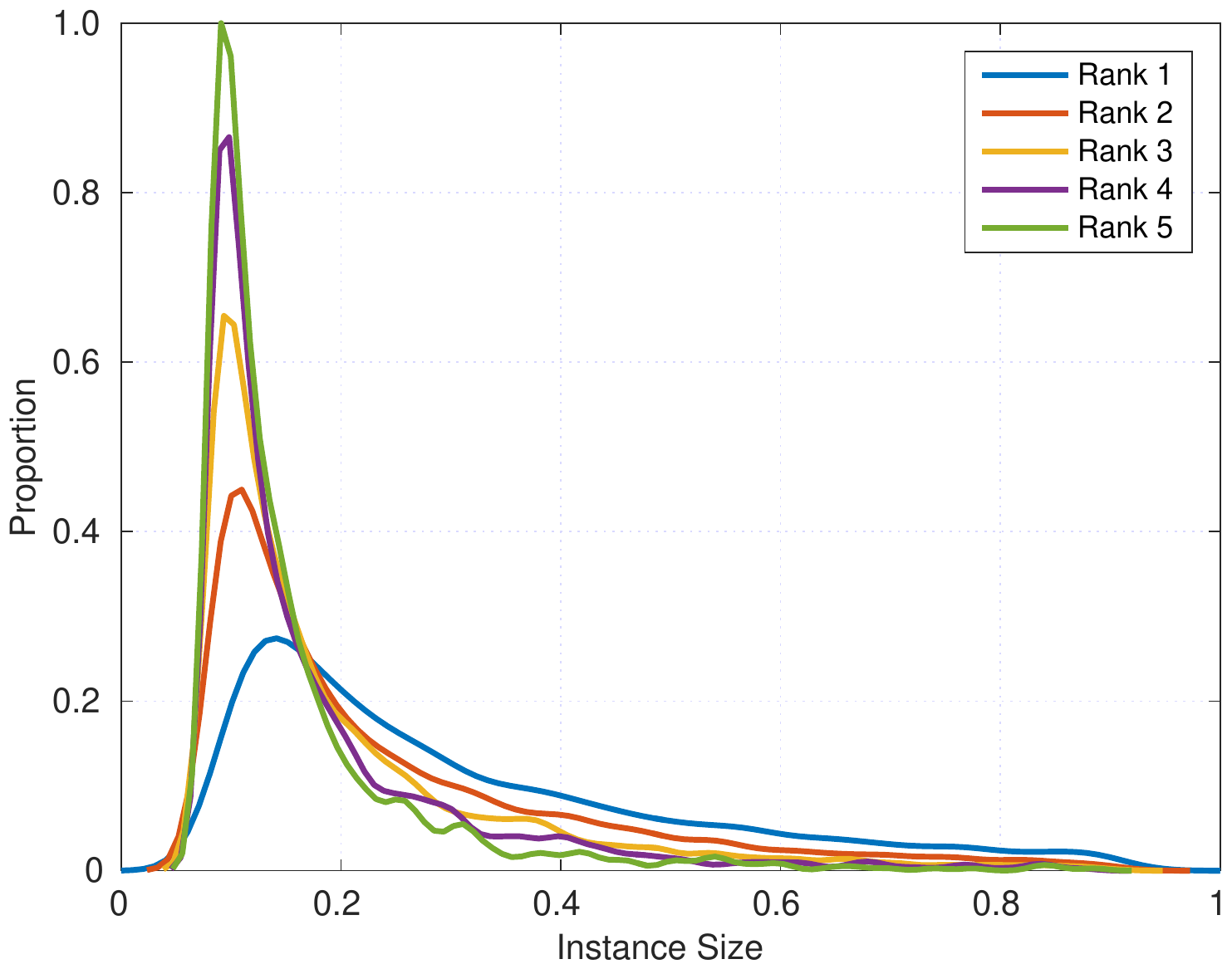}&	
			\includegraphics[width=0.24\textwidth]{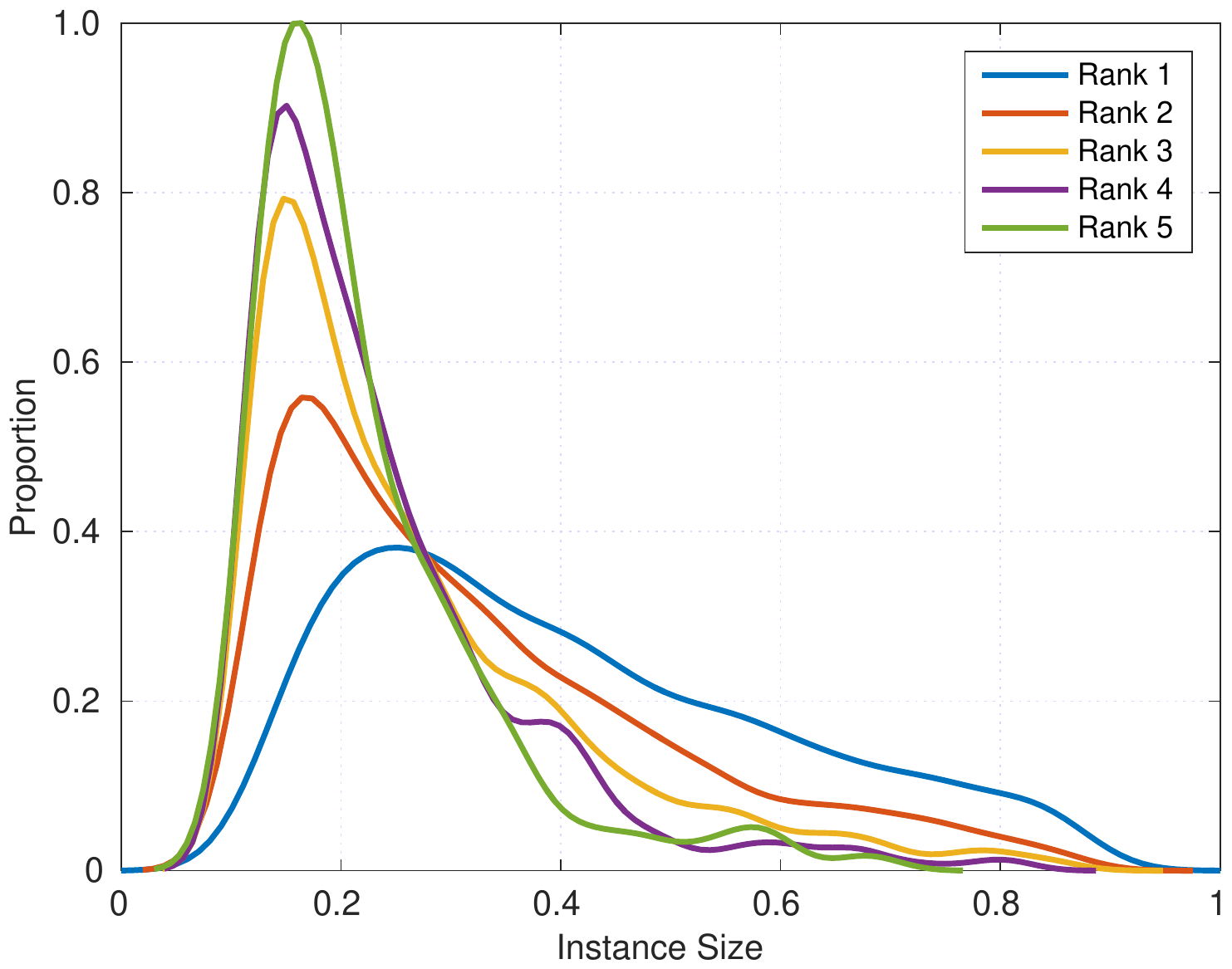}\\	
		\end{tabular}}
		\caption{\textcolor{violet}{The distribution of instance size according to their associated ranks on the COCO-SalRank dataset. Left : version I, Right: version II.}}
		\label{fig:sizevsrank}
		\vspace{-0.4cm}
\end{figure}

\noindent \textcolor{violet}{\textbf{Distribution of Ranking among Images:}}
\noindent \textcolor{violet}{The distribution of the images in COCO-SalRank dataset with respect to differing salient object rank order is shown in Table~\ref{table:dis_rank}. It is an evident from the table that, there is a considerable number of images with more than one salient object which is ideal for a saliency ranking problem.}
\begin{table}[H]
	\vspace{-0.3cm}
	\caption{ \textcolor{violet}{Distribution of images corresponding to different rank order of salient objects on the COCO-SalRank dataset.}  }	
	\vspace{-0.3cm}	
	\centering
	\resizebox{0.48\textwidth}{!}{
		\begin{tabular}{ccccccccccc}
			%\specialrule{1.1pt}{1pt}{1pt}
			%\multicolumn{11}{c}{\color{black}COCO-SalRank } \\
			\specialrule{1.2pt}{1pt}{1pt}
			
			\multicolumn{5}{c}{version I - \textbf{10,410}} && \multicolumn{5}{c}{version II - \textbf{4,433}} \\

			1 & 2 &3 & 4 & 5 && 1 & 2 &3 & 4 & 5 \\
			
			\cline{1-5} \cline{7-11}
			
			1080     &   3058      &  3162    &    2097   &     1013 && 708    &    1732    &    1304     &    556    &     133 \\
			
			%10.5	& 80& 	5& 	\color{blue}0.4& 	0.65& 	0.4	&\color{red} 0.7  &&  10.5	& 80& 	5& 	\color{blue}0.7& 	0.65& 	0.4	& \color{red}0.9\\
			
			\specialrule{1.2pt}{1pt}{1pt}
		\end{tabular}}
		
		\label{table:dis_rank}
		\vspace{-0.2cm}		
\end{table}	
\noindent\textcolor{violet}{ \textbf{Size vs Rank Distribution:} According to~\cite{li2014secrets}, the size of an instance-level salient object can be defined as the proportion of pixels in the image. As shown in Fig.~\ref{fig:sizevsrank}, the rank of salient instances in our COCO-SalRank varies significantly. The large-scale instances are likely to have higher rank compared to the medium-sized or smaller-sized instances.} \\
\noindent \textbf{Bias in Data Driven Saliency Annotation:} \textcolor{violet}{The proposed saliency ranking dataset is data-driven and primarily based on the fixation annotations derived from the MS-COCO and SALICON. We know that the data-driven saliency annotation results will be influenced by the annotators whose choices may depend on a variety of factors including at least (culture, gender, background, prior experience). It is a challenge to control for these effects, and this implies that any dataset will always carry some inherent biases that are not image dependent. Related to this, the selection images, and associated rankings for the COCO-SalRank is based on parameters obtained from the PASCAL-SR dataset. An implication of this is that biases in the source data may transfer to the generated dataset. Moreover, one needs to exercise care in considering whether base characteristics of the datasets are similar in terms of masks, source data and fixations derived behaviour. To this end, we examine characteristics of the two datasets, and assigned salient regions to satisfy ourselves that they share similar characteristics. While any base biases may be transferred, this is intrinsic to the data collection process and unavoidable. With that said, we can take some measures to consider what impact any such bias may have insofar as it interacts with the target dataset. To this end, we include a variety of analysis and statistics in Table~\ref{table:bias} that reveals characteristics of selected salient regions, and also sensitivity to number of annotators in deriving the target data from the source data. From Table~\ref{table:bias}, we observe that there is a high correlation in ranking between the original ground truth and one produced when we randomly remove some annotators. This speaks to questions such as the extent to which size of objects matters (which we also examine in Fig. \ref{fig:sizevsrank}), correlation between gaze and object ranking across datasets, and the validity of normalization parameters which seem to be relatively stable even as the number of annotators is artificially suppressed.}
\begin{table}[h]
	\vspace{-0.3cm}
	\caption{Average SOR for five trials on PASCAL-SR dataset where in each trial we randomly remove annotations provided by $\mathcal{Y}$ annotators.}	
	\vspace{-0.3cm}	
	\centering
	\resizebox{0.48\textwidth}{!}{
		\begin{tabular}{c|cccccccccc}
			\specialrule{1.1pt}{1pt}{1pt}
			%\multicolumn{16}{c}{\color{red}PASCAL-SR } \\

			$\mathcal{Y}$&1&2&3&4&5&6&7&8&9&10 \\
			\midrule
			
			SOR$_\text{avg}$ & 0.967 &0.968&0.955&0.966&0.958&0.957&0.963&0.953&0.942& 0.931\\
			
			\specialrule{1.2pt}{1pt}{1pt}
		\end{tabular}}
		
		\label{table:bias}
		\vspace{-0.5cm}		
\end{table}	
%\vspace{-0.3cm}
\subsection{Relative and Absolute Salience Assignment} \label{sec:salience_assignment}
\textcolor{violet}{The existing dataset (Pascal-S~\cite{li2014secrets}) contains} information that allows for implicit assignment of relative salience based on agreement among multiple observers while in our case we propose to assign rank value under two different settings (\textit{Relative} and \textit{Absolute}). In the relative setting, we assign rank value based on total number of instances in the mask and their rank score, $\mathbb{R}_\chi$ where $\chi$ is a particular instance. For example, if we have total $\tau$ number of instances in the mask, we divide the range [0, 255] by $\tau$ to obtain the numeric rank value. While in the absolute case, rank values are assigned based on the percentile of the rank score set and then re-scaled to the range [50 255] which corresponds to [20\% - 100\%] of the gray-scale levels. Similar to~\cite{cvpr18_rank}, we generate a stack representation of the ground truth where each stack consists of five slices since a cap of five objects is set in filtering to avoid over-segmentation. For the relative case, the first slice of the stack contains the most salient object, the second slice contains the top two salient objects and so on. However, for the absolute case, the first slice accounts for less than 20\% of the fixations, the second one accounts for less than 40\% of the fixations, and so on.
%%%%%% Stack represenation of ground-truth
\begin{equation}\label{eq:stack_gt1}
	%\vspace{-0.4cm}
	\mathcal{R}_{\vartheta} =
	\begin{bmatrix} R_i \\ \includegraphics[width=0.05\textwidth]{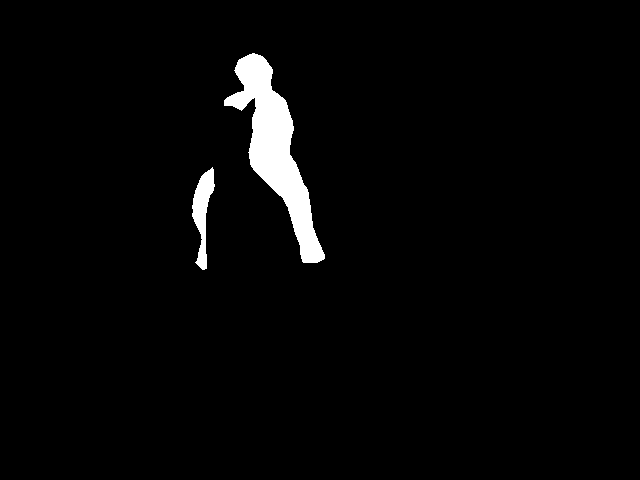} \end{bmatrix}
	\begin{bmatrix} R_{i+1} \\ \includegraphics[width=0.05\textwidth]{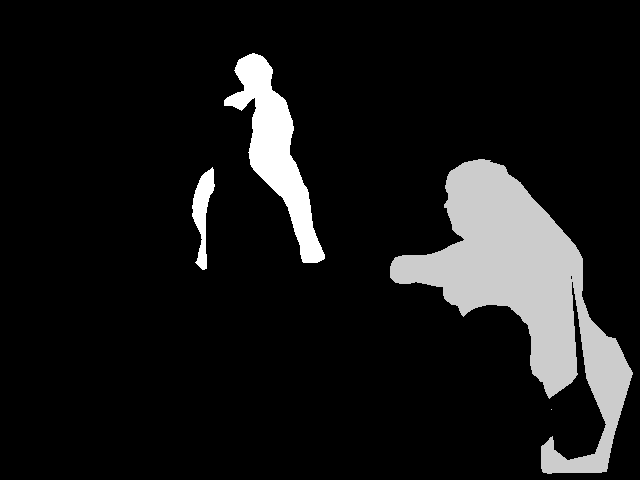} \end{bmatrix}
	\begin{bmatrix} R_{i+2} \\ \includegraphics[width=0.05\textwidth]{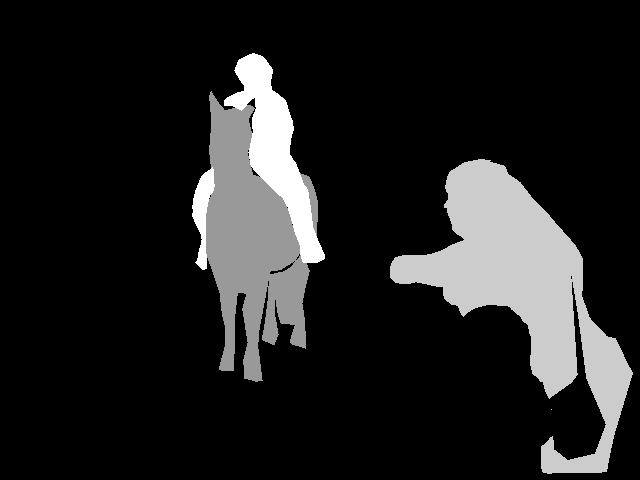} \end{bmatrix}
	\begin{bmatrix} R_{i+3} \\ \includegraphics[width=0.05\textwidth]{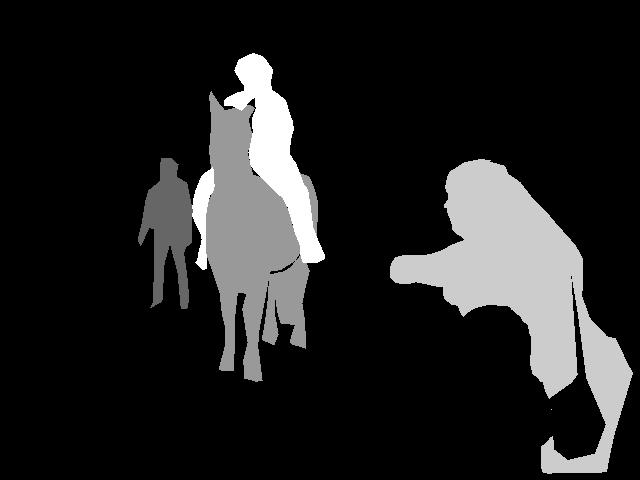} \end{bmatrix}
	\begin{bmatrix} R	_{i+4} \\ \includegraphics[width=0.05\textwidth]{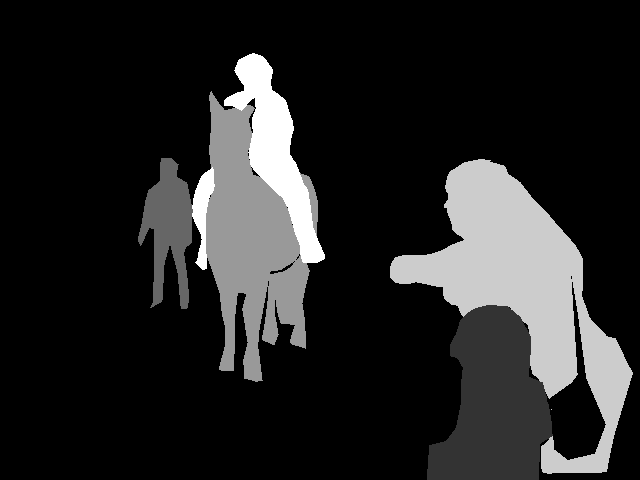} \end{bmatrix}
\end{equation}
\begin{equation}\label{eq:stack_absolute}
	\mathcal{A}_{\vartheta} =
	\begin{bmatrix} A_i \\ \includegraphics[width=0.05\textwidth]{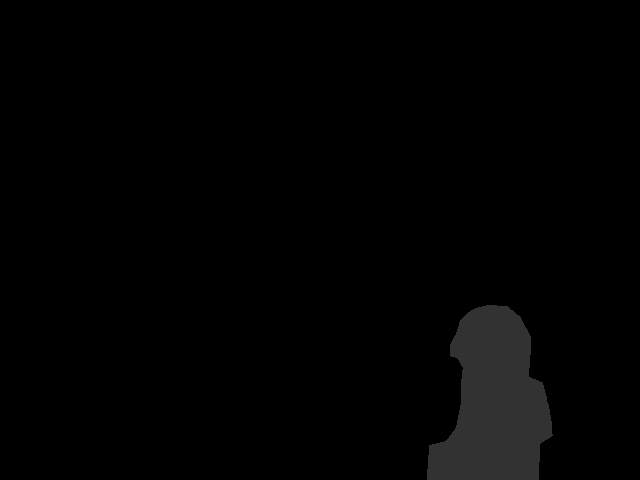} \end{bmatrix}
	\begin{bmatrix} A_{i+1} \\ \includegraphics[width=0.05\textwidth]{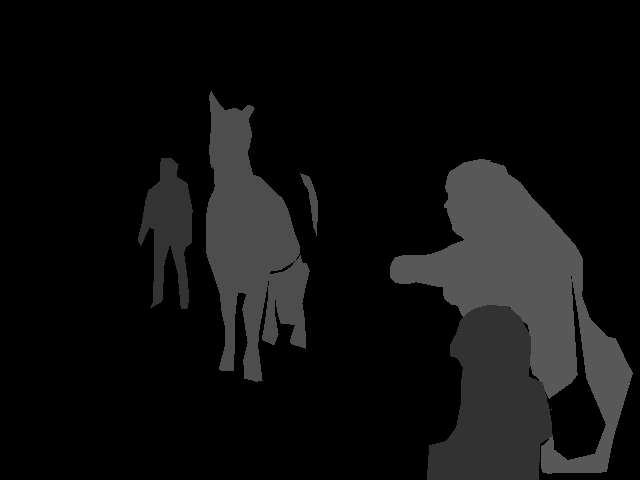} \end{bmatrix}
	\begin{bmatrix} A_{i+2} \\ \includegraphics[width=0.05\textwidth]{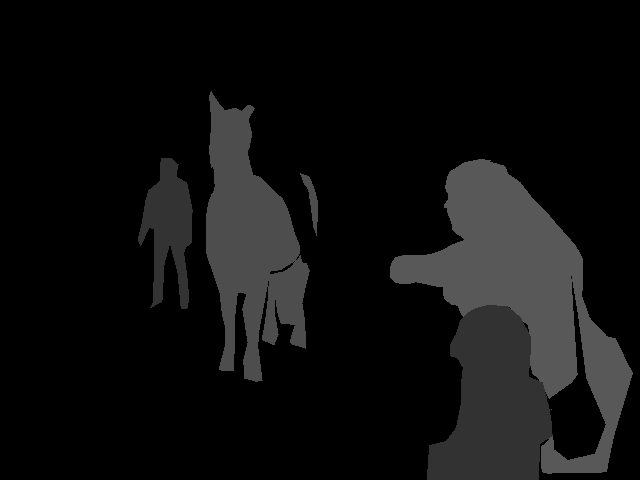} \end{bmatrix}
	\begin{bmatrix} A_{i+3} \\ \includegraphics[width=0.05\textwidth]{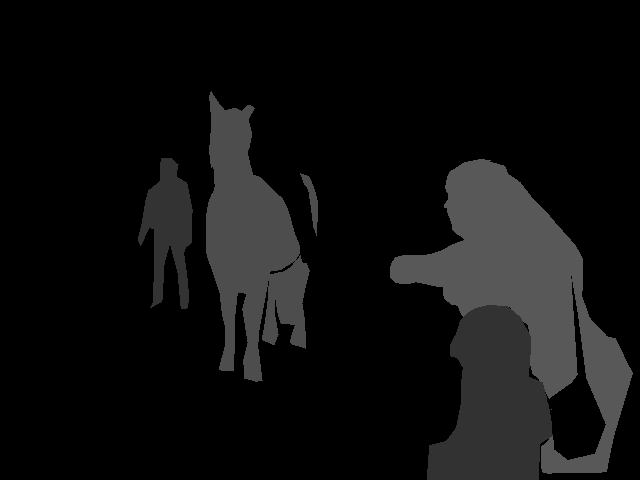} \end{bmatrix}
	\begin{bmatrix} A_{i+4} \\ \includegraphics[width=0.05\textwidth]{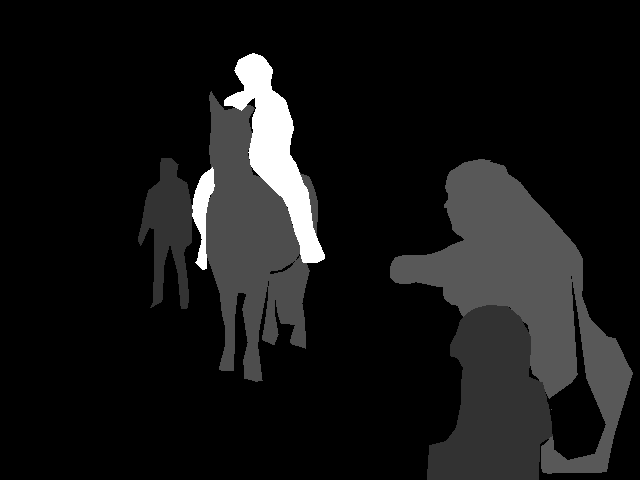} \end{bmatrix}
\end{equation}
The stacked representation of the ground-truth in relative and absolute settings are shown in Eq.~\ref{eq:stack_gt1} and Eq.~\ref{eq:stack_absolute} respectively. As shown in this example (in the case of five ranked instances), a new instance is added to each slice in the relative case, whereas, multiple instances may be added at once in the absolute case if they reside within the same percentile rank.	
\subsubsection{Ranking Mechanism}
As saliency ranking is a newly proposed problem, there is no universally agreed upon metric to obtain the rank order of salient instances. We start with the metric proposed in our earlier work~\cite{cvpr18_rank} and explore a few alternatives ranking mechanisms. Firstly, rank order of a salient instance is obtained by averaging the degree of saliency within that instance mask. Then we propose to assign the rank order from an output saliency map by dividing the total degree of saliency within an instance to its size raised to a certain power. Finally, we obtain the rank order by taking the max value of the saliency within the instance region. These considerations relate to the earlier observations concerning the size-saliency tradeoff and its relation to ranking. We can write these metrics as follows:
	\begin{equation}  \text{Rank} =  \left\{
	\begin{array}{ll}
	\text{SOR}_\text{avg} (\mathcal{S} (\delta)) = &\frac{\sum_{i=1}^{\rho_\delta} \delta(x_i, y_i)}{\rho_\delta}\\
	
	\text{SOR}_\text{pow} (\mathcal{S} (\delta); \alpha) = &\frac{\sum_{i=1}^{\rho_\delta} \delta(x_i, y_i)}{\rho_\delta^\alpha}  \\
	
	\text{SOR}_\text{max} (\mathcal{S} (\delta)) =  &\text{max} (  \delta(x_i, y_i)) \\
	\label{eq:eq1}
	\end{array}
	\right.
	\end{equation}
where $\delta$ represents a particular instance of the predicted saliency map (${S}$),  $\alpha$ is set to 0.3, $\rho_\delta$ denotes total numbers of pixels $\delta$ contains, and $\delta(x_i, y_i)$ refers to saliency score for the pixel $(x_i, y_i)$.
\begin{figure}[t]
	%\vspace{-0.4cm}
	\centering
	\setlength\tabcolsep{1.0pt}
	\def\arraystretch{0.9}
	\resizebox{0.49\textwidth}{!}{
		\begin{tabular}{c}
			
			\begin{tabular}{ c c c }
				
				\fcolorbox{black}{black} {\includegraphics[width=0.12\textwidth]{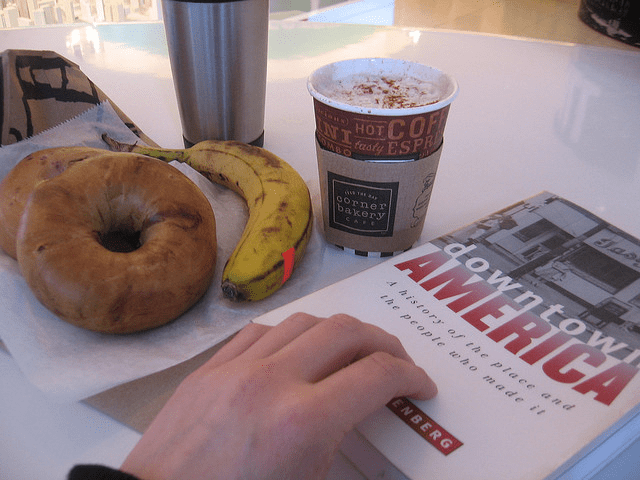}}&
				\fcolorbox{black}{black} {\includegraphics[width=0.12\textwidth]{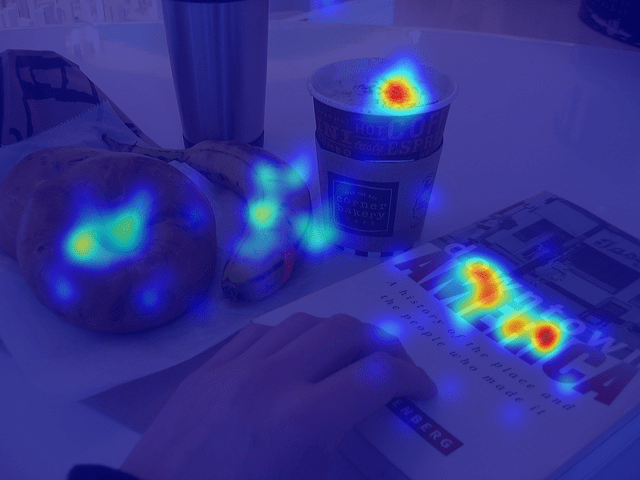}}&
				\fcolorbox{black}{black} {\includegraphics[width=0.12\textwidth]{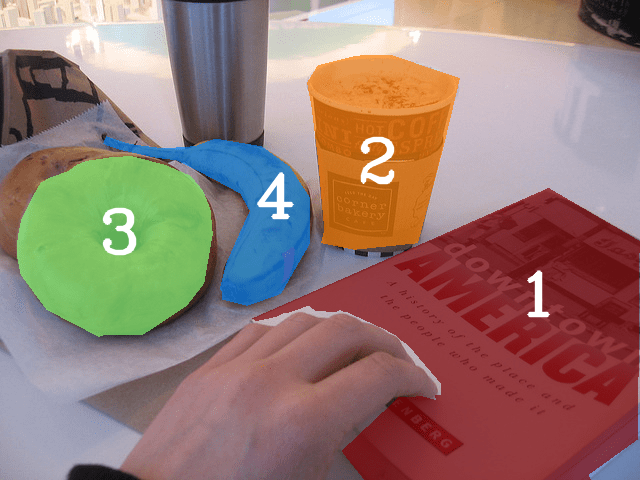}}\\
				
				\fcolorbox{black}{black} {\includegraphics[width=0.12\textwidth]{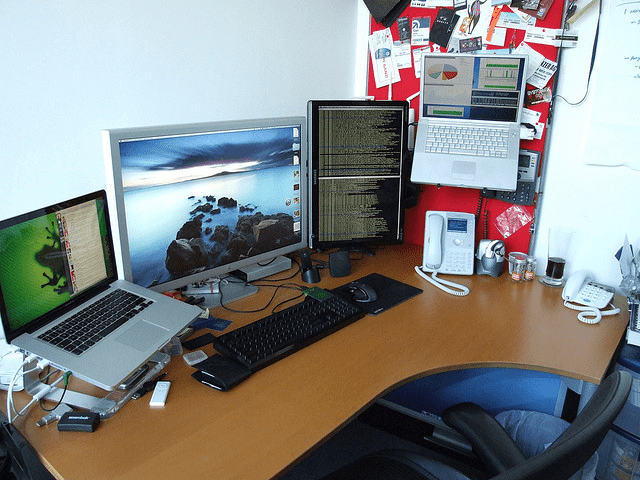}}&
				\fcolorbox{black}{black} {\includegraphics[width=0.12\textwidth]{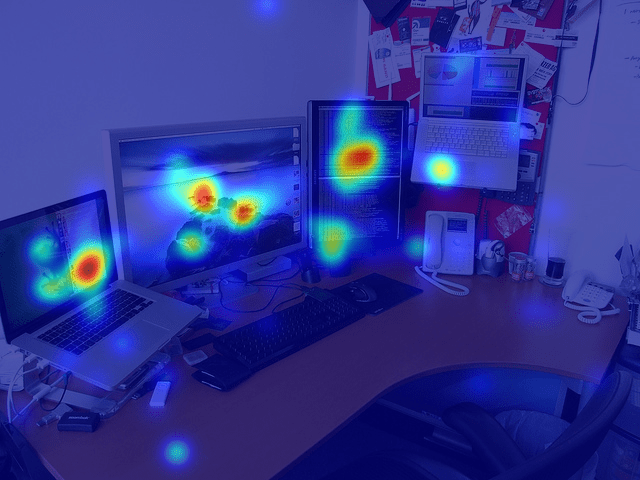}}&
				\fcolorbox{black}{black} {\includegraphics[width=0.12\textwidth]{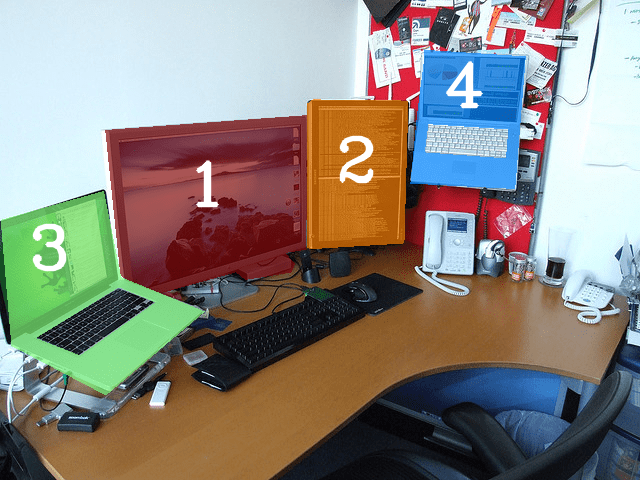}}\\
				
				\fcolorbox{black}{black} {\includegraphics[width=0.12\textwidth]{images/ranking_ours_2/im_1905_im.png}}&
				\fcolorbox{black}{black} {\includegraphics[width=0.12\textwidth]{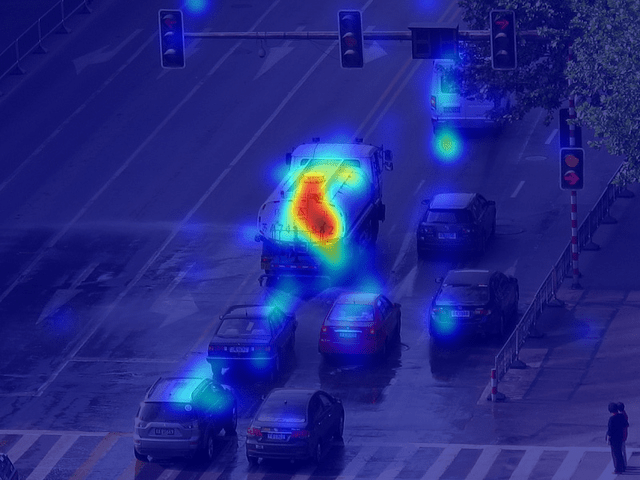}}&
				\fcolorbox{black}{black} {\includegraphics[width=0.12\textwidth]{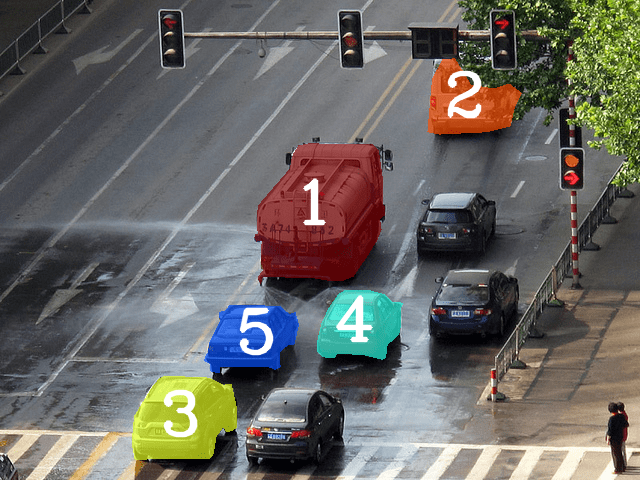}}\\
				
				image & fixation & gt \vspace{0.1cm}\\

			\end{tabular}
			\hspace{0.001cm}
			\begin{tabular}{ c c c }
				
				\fcolorbox{black}{black} {\includegraphics[width=0.12\textwidth]{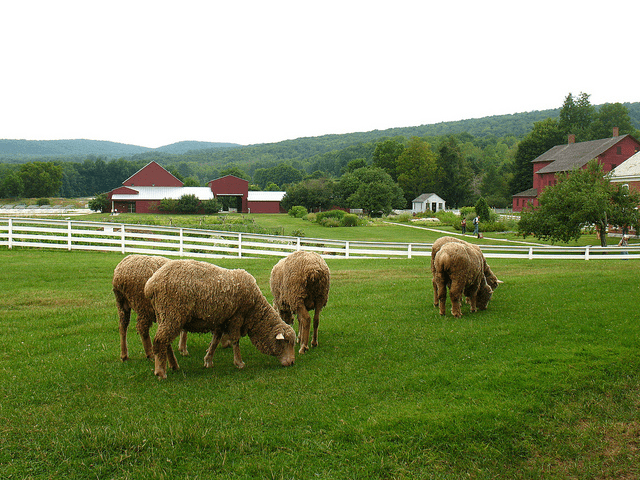}}&
				\fcolorbox{black}{black} {\includegraphics[width=0.12\textwidth]{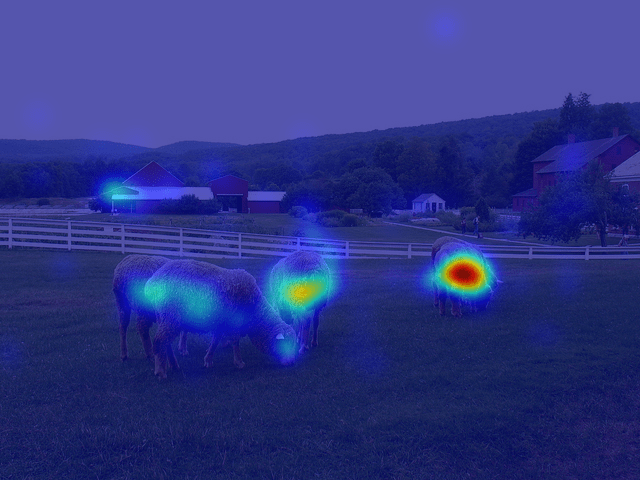}}&
				\fcolorbox{black}{black} {\includegraphics[width=0.12\textwidth]{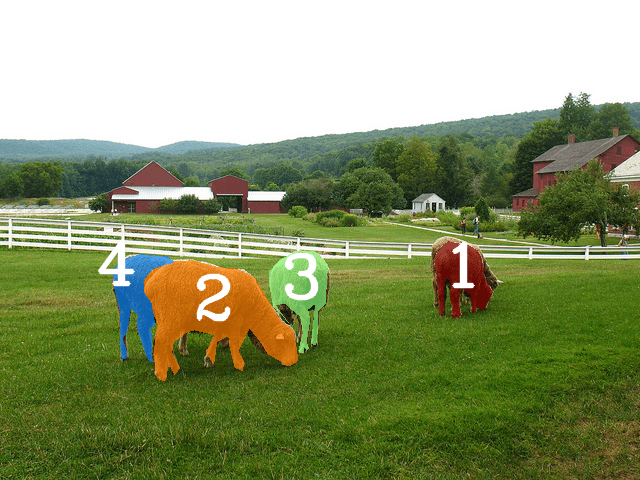}}\\
				
				\fcolorbox{black}{black} {\includegraphics[width=0.12\textwidth]{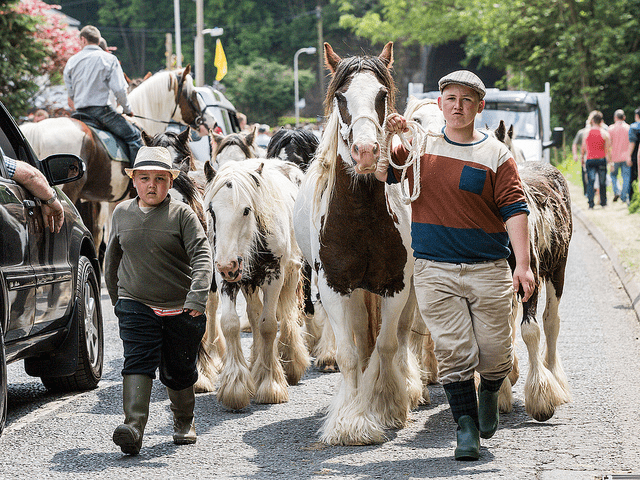}}&
				\fcolorbox{black}{black} {\includegraphics[width=0.12\textwidth]{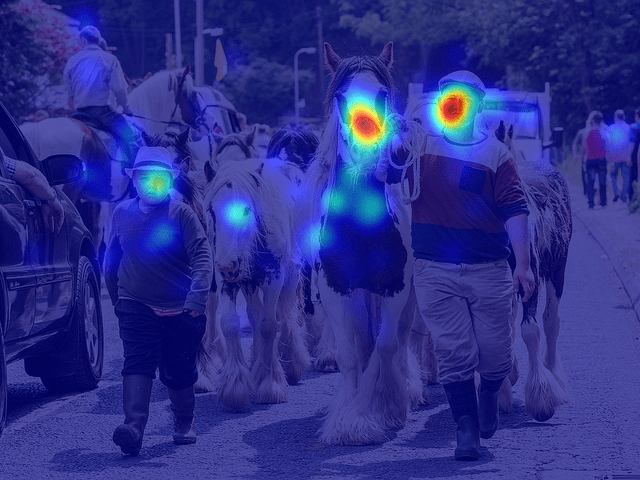}}&
				\fcolorbox{black}{black} {\includegraphics[width=0.12\textwidth]{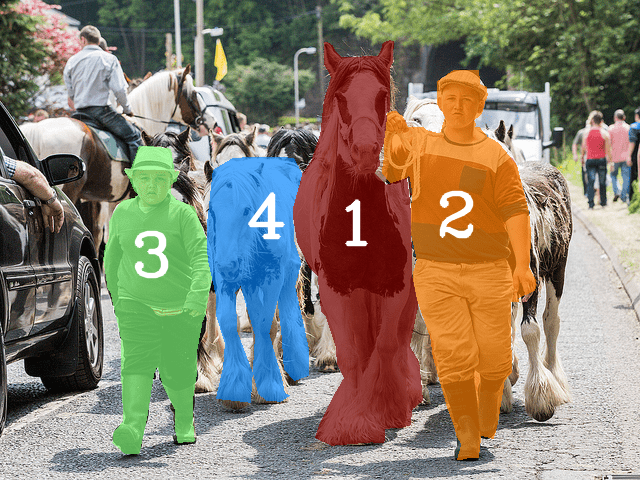}}\\
				
				\fcolorbox{black}{black} {\includegraphics[width=0.12\textwidth]{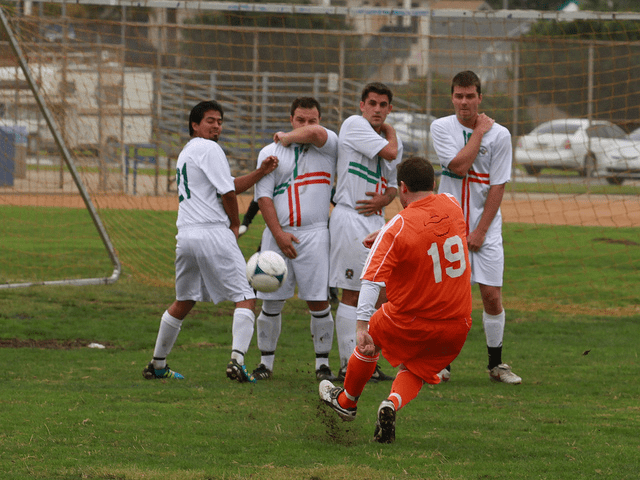}}&
				\fcolorbox{black}{black} {\includegraphics[width=0.12\textwidth]{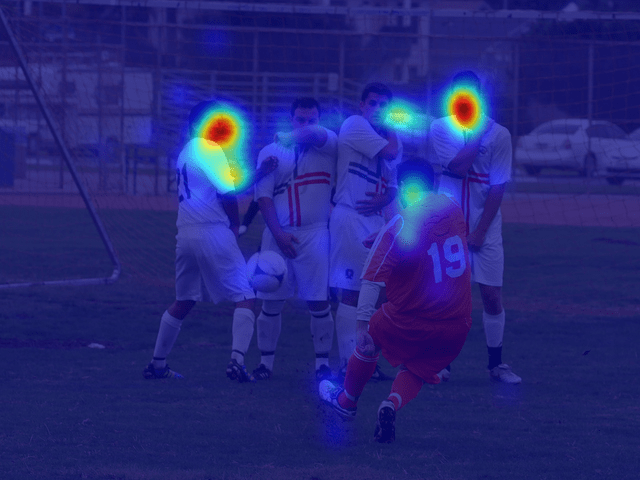}}&
				\fcolorbox{black}{black} {\includegraphics[width=0.12\textwidth]{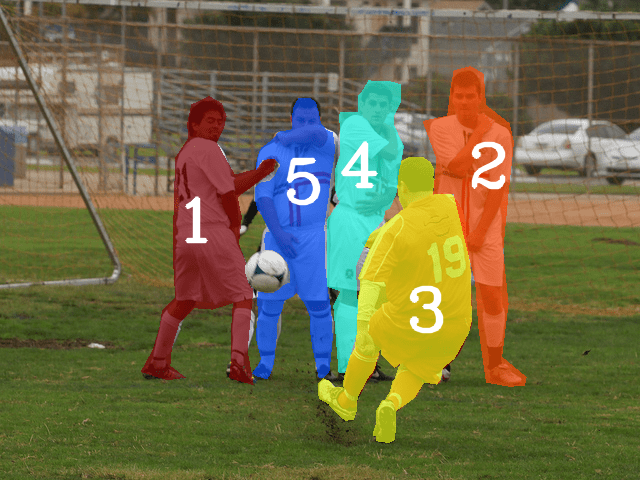}}\\
					
				image & fixation & gt \vspace{0.1cm}\\ %\vspace{0.1cm}\\
				
			\end{tabular}

		\end{tabular}}
		\caption{Qualitative illustration of obtained ground-truth samples on COCO-SalRank dataset (noisy version). \textcolor{violet}{Relative rank is indicated by the assigned color and number on each salient instance}. The consistency among \textcolor{black}{simulated mouse-based fixation maps} and ground-truth ranking shows good agreement and an intuitive ranking for our proposed dataset.}
		\label{fig:cocosalrank_gt}
		\vspace{-0.2cm}
\end{figure}
\subsection{Qualitative Examples of COCO-SalRank Dataset}
Fig.~\ref{fig:cocosalrank_gt} depicts visual examples of generated saliency ranking ground-truth on the COCO-SalRank dataset with respect to fixation maps. It is an evident that our algorithm produces ground-truth maps that have an intuitive correspondence with rank and that are consistent with fixation maps in various challenging cases.
\subsection{Challenging Saliency Ranking Cases}
\begin{figure}[h]
			\vspace{-0.1cm}
			\centering
			\setlength\tabcolsep{1.0pt}
			\def\arraystretch{0.8}
			\resizebox{0.49\textwidth}{!}{
				\begin{tabular}{c}
					
					\begin{tabular}{ c c c }
						
						\fcolorbox{black}{black} {\includegraphics[width=0.12\textwidth]{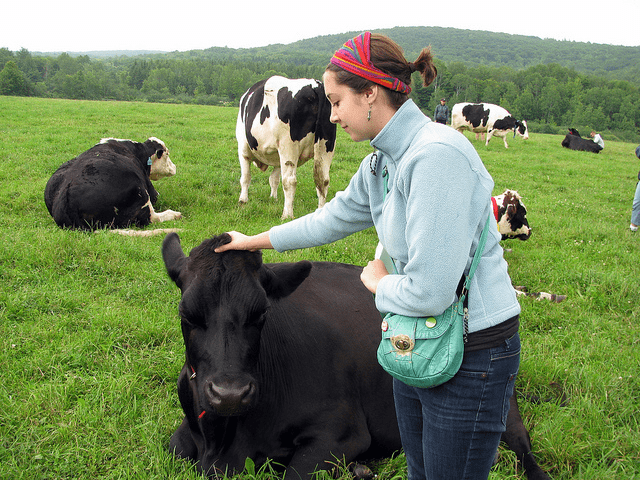}}&
						\fcolorbox{black}{black} {\includegraphics[width=0.12\textwidth]{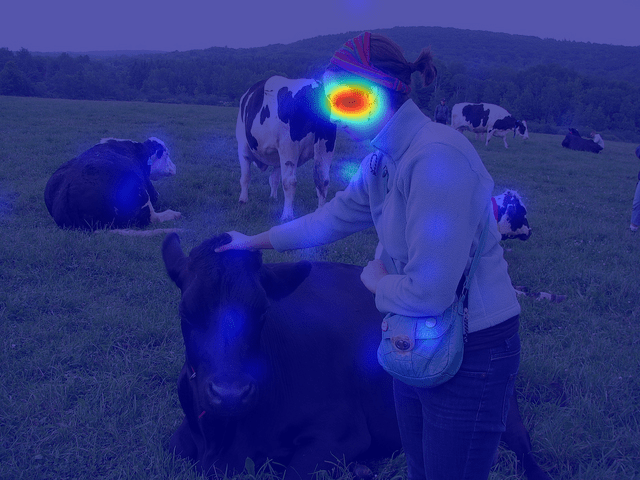}}&
						\fcolorbox{black}{black} {\includegraphics[width=0.12\textwidth]{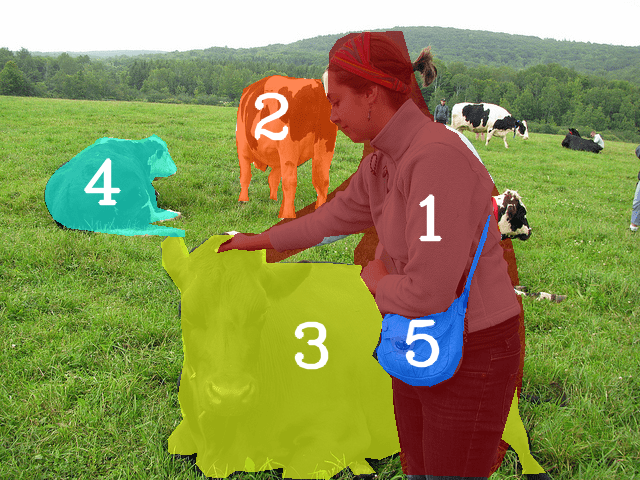}}\\
						
						\fcolorbox{black}{black} {\includegraphics[width=0.12\textwidth]{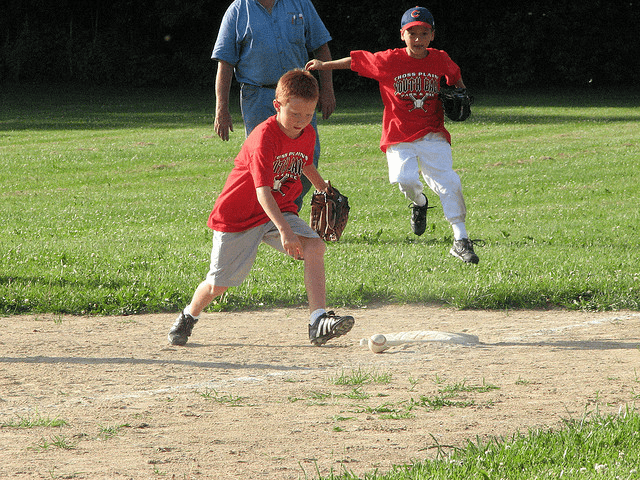}}&
						\fcolorbox{black}{black} {\includegraphics[width=0.12\textwidth]{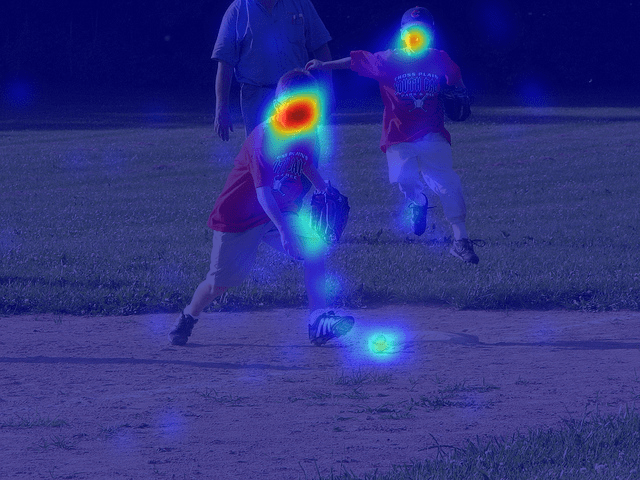}}&
						\fcolorbox{black}{black} {\includegraphics[width=0.12\textwidth]{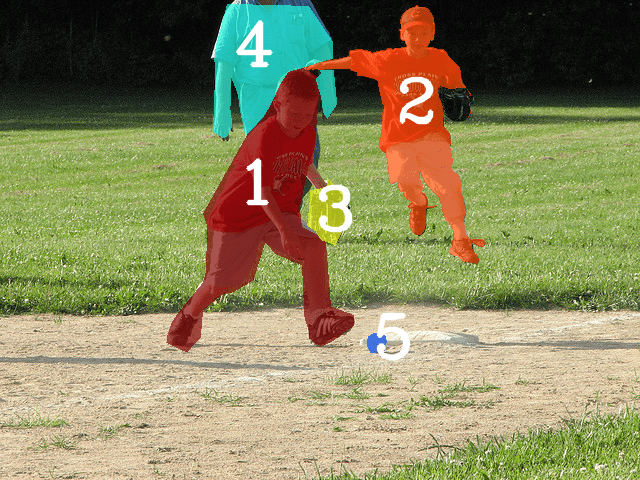}}\\
									
						image & fixation & gt \vspace{0.1cm}\\
						
					\end{tabular}
					
					\hspace{0.04cm}
					\begin{tabular}{ c c c ccc}
								
						\fcolorbox{black}{black} {\includegraphics[width=0.12\textwidth]{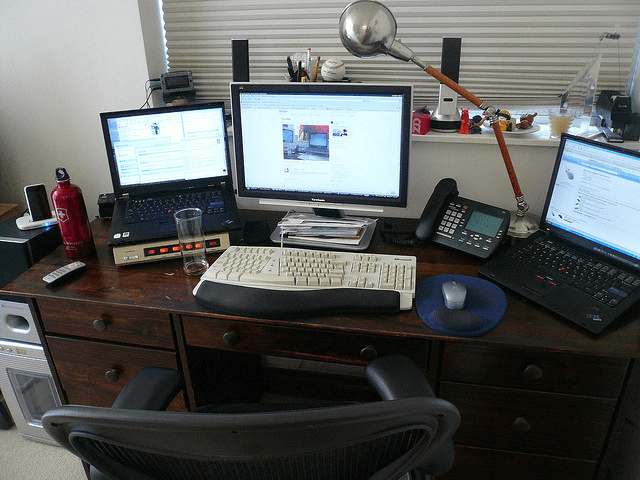}}&
						\fcolorbox{black}{black} {\includegraphics[width=0.12\textwidth]{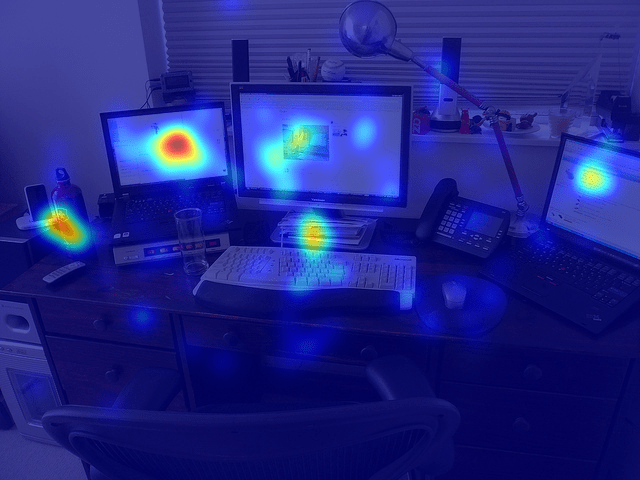}}&
						\fcolorbox{black}{black} {\includegraphics[width=0.12\textwidth]{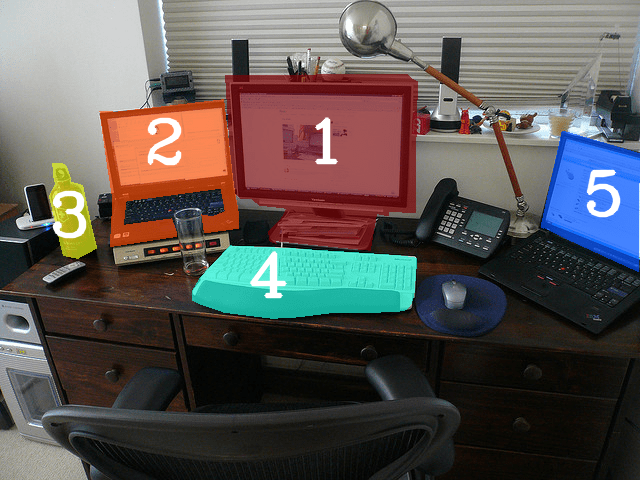}}&&&
						\multirow{2}{*}[1.42cm]{\includegraphics[width=0.031\textwidth]{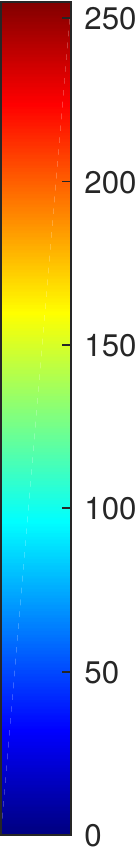}}\\
						
						\fcolorbox{black}{black} {\includegraphics[width=0.12\textwidth]{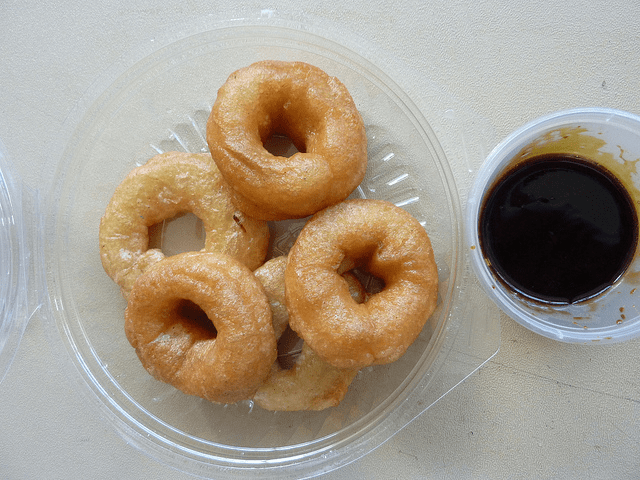}}&
						\fcolorbox{black}{black} {\includegraphics[width=0.12\textwidth]{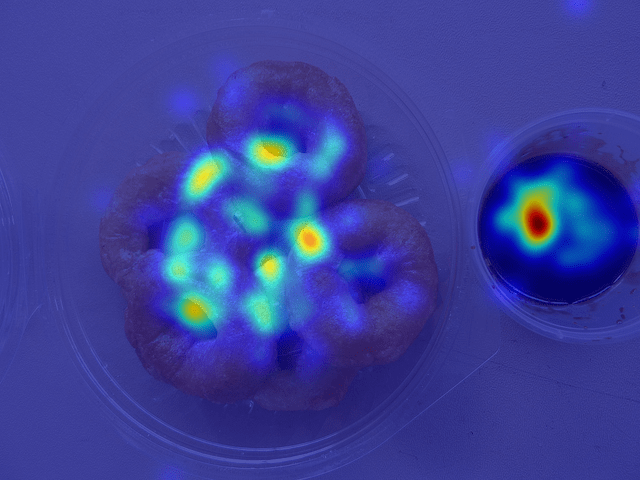}}&
						\fcolorbox{black}{black} {\includegraphics[width=0.12\textwidth]{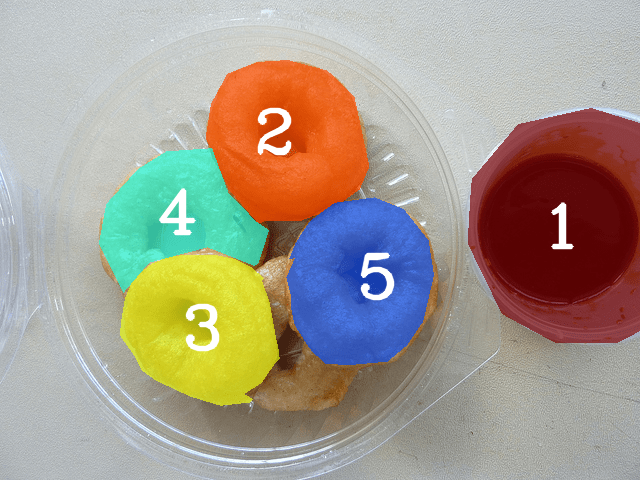}}&&&\\

						image & fixation & gt \vspace{0.1cm}\\

					\end{tabular}

				\end{tabular}}
				\caption{Shown are some illustrative examples of inconsistency among simulated mouse-based fixation maps and the generated ground-truth ranking on COCO-SalRank dataset. These cases are most common for overlapping instances, and for scenes with fixations spread over multiple salient objects. \textcolor{violet}{Assigned colors and numbers indicate the relative rank of each salient instance.}}
				\label{fig:cocosalrank_failure}
				\vspace{-0.5cm}
\end{figure}

Despite the consistent ground-truth quality for the majority of cases; there are samples that are especially challenging to assign rank order by algorithmic means (see Fig.~\ref{fig:cocosalrank_failure}). Sometimes, fixations are distributed among multiple objects which makes the ranking task challenging. Other failures of ranking happen when a higher ranked object overlaps with or exists close to a less salient object. In that case, the lower ranked object receives some attention attributed to fixations from the higher ranked instance. An additional challenge is the lack of consistency in instance-wise labeling in MS-COCO.
\section{Experimental Results}\label{sec:exp}
The core of our model follows a structure based on ResNet-101~\cite{He2015} with pre-trained weights to initialize the encoder portion. %and random initialization of any newly added layers that follow the encoder drawn from a standard gaussian distribution.
 \textcolor{violet}{ The network is trained using stochastic gradient descent for 20k iterations with momentum of 0.9, weight decay of 0.0005 and the ``poly'' learning rate policy. Testing uses the full resolution image while training relies on random cropping for memory savings.}
A few variants of the basic architecture are proposed, and we report numbers for the following  variants that are described in what follows:

\noindent \textbf{RSDNet:} This network includes dilated ResNet-101~\cite{chen2016deeplab} + NRSS + SCM.
\textbf{RSDNet-A:} This network is the same as RSDNet except the ground-truth is scaled by a factor of 1000, encouraging the network to explicitly learn deeper contrast. \textbf{RSDNet-B:} The structure follows RSDNet except that an \textcolor{violet}{atrous pyramid pooling module~\cite{chen2016deeplab} is applied to get NRSS prediction}. \textbf{RSDNet-C:} RSDNet-B + the ground-truth scaling. \textbf{RSDNet-R:} RSDNet with stage-wise rank-aware refinement units + multi-stage saliency map fusion. %to generate the final prediction map.

\textcolor{violet}{Additionally, we experiment with DeepLabv2-VGG~\cite{chen2016deeplab} and PSPNet~\cite{zhao2017pyramid} networks paired with NRSS to establish initial baselines for our proposed COCO-SalRank dataset. It is worth noting that the inclusion of NRSS at the end of the network architectures allows any model to be trained for the saliency ranking task.}
\subsection{Datasets and Evaluation Metrics}\label{sec:dataset}
\noindent \textbf{Datasets:} We report experimental results on \textcolor{violet}{PASCAL-SR (extended version of publicly available PASCAL-S)} and our proposed COCO-SalRank datasets. The PASCAL-SR dataset includes 850 natural images with multiple complex objects derived from the PASCAL VOC 2012 validation set~\cite{everingham2010pascal}. We randomly split the PASCAL-SR dataset into two subsets (425 for training and 425 for testing). In this dataset, salient object labels are based on an experiment using 12 participants to label salient objects. Virtually all existing approaches for salient object segmentation or detection threshold the ground-truth saliency map to obtain a binary saliency map. This operation seems somewhat arbitrary since the threshold can require consensus among $k$ observers, and the value of $k$ varies from one study to another. This is one of the most highly used salient object segmentation datasets, but is unique in having multiple explicitly tagged salient regions provided by a reasonable sample size of observers. Since a key objective of this work is to rank salient objects in an image, we use the original ground-truth maps (each pixel having a value corresponding to the number of observers that deemed it to be a salient object) rather than trying to predict a binary output based on an arguably contentious thresholding process.
\\
\noindent \textbf{Evaluation Metrics:}
For the multiple salient object detection task, we use five different metrics to measure performance including precision-recall (PR) curves, F-measure (maximal along the curve), S-measure~\cite{fan2017structure}, Area under ROC curve (AUC), and mean absolute error (MAE). Since some of these rely on binary decisions, we threshold the ground-truth saliency map based on the number of participants that deem an object salient, resulting in 12 binary ground truth maps. For each binary ground truth map, multiple thresholds of a predicted saliency map allow for calculation of the true positive rate (TPR), false positive rate (FPR), precision and recall, and corresponding ROC and PR curves. Given that methods that predate this work are trained based on varying thresholds and consider a binary ground truth map, scores are reported based on the binary ground truth map that produces the best AUC or F-measure score (and the corresponding curves are shown). Max F-measure and average F-measure  are also reported to provide a sense of how performance varies as a function of the threshold chosen. We also report the MAE score i.e. the average pixel-wise difference between the predicted saliency map and the binary ground-truth map that produces the minimum score. \textcolor{violet}{S-measure computes the structural similarity between the predicted and ground-truth saliency map}.  

In ordered to evaluate the rank order of salient objects, we introduce the \textit{Salient Object Ranking} (SOR) metric which is defined as the Spearman's Rank-Order Correlation between the ground truth rank order and the predicted rank order of salient objects. SOR score is normalized to [0 1] for ease of interpretation. Scores are reported based on the average SOR score for each method considering the whole dataset.
\subsection{Performance Comparison with State-of-the-art}
\begin{figure} [H]
	\centering
	\vspace{-0.2cm}
	\setlength\tabcolsep{1.0pt}
	\resizebox{0.45\textwidth}{!}{
		\begin{tabular}{*{2}{c }}
			
			\includegraphics[width=0.24\textwidth,height=3.64cm]{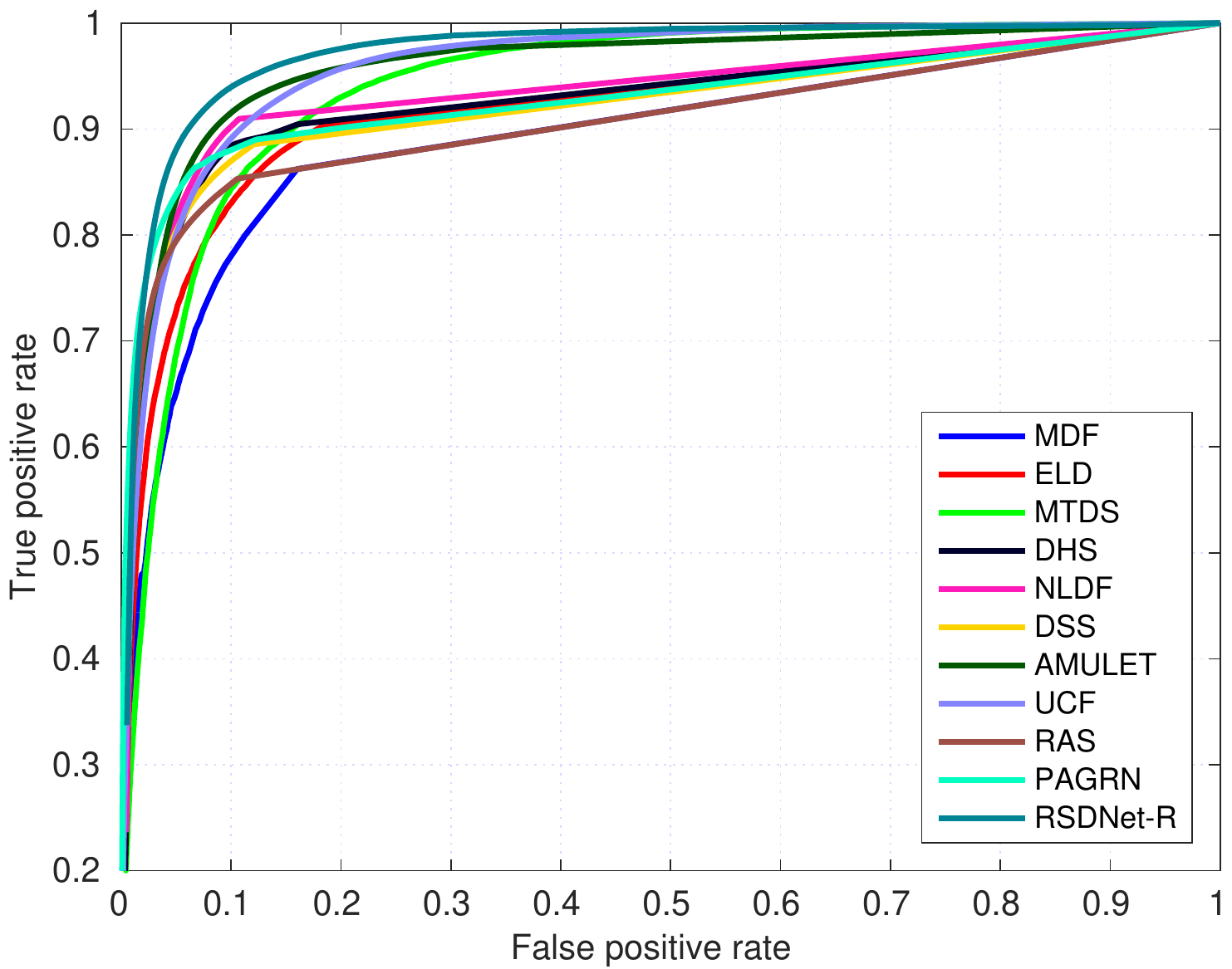}&	\includegraphics[width=0.24\textwidth,height=3.64cm]{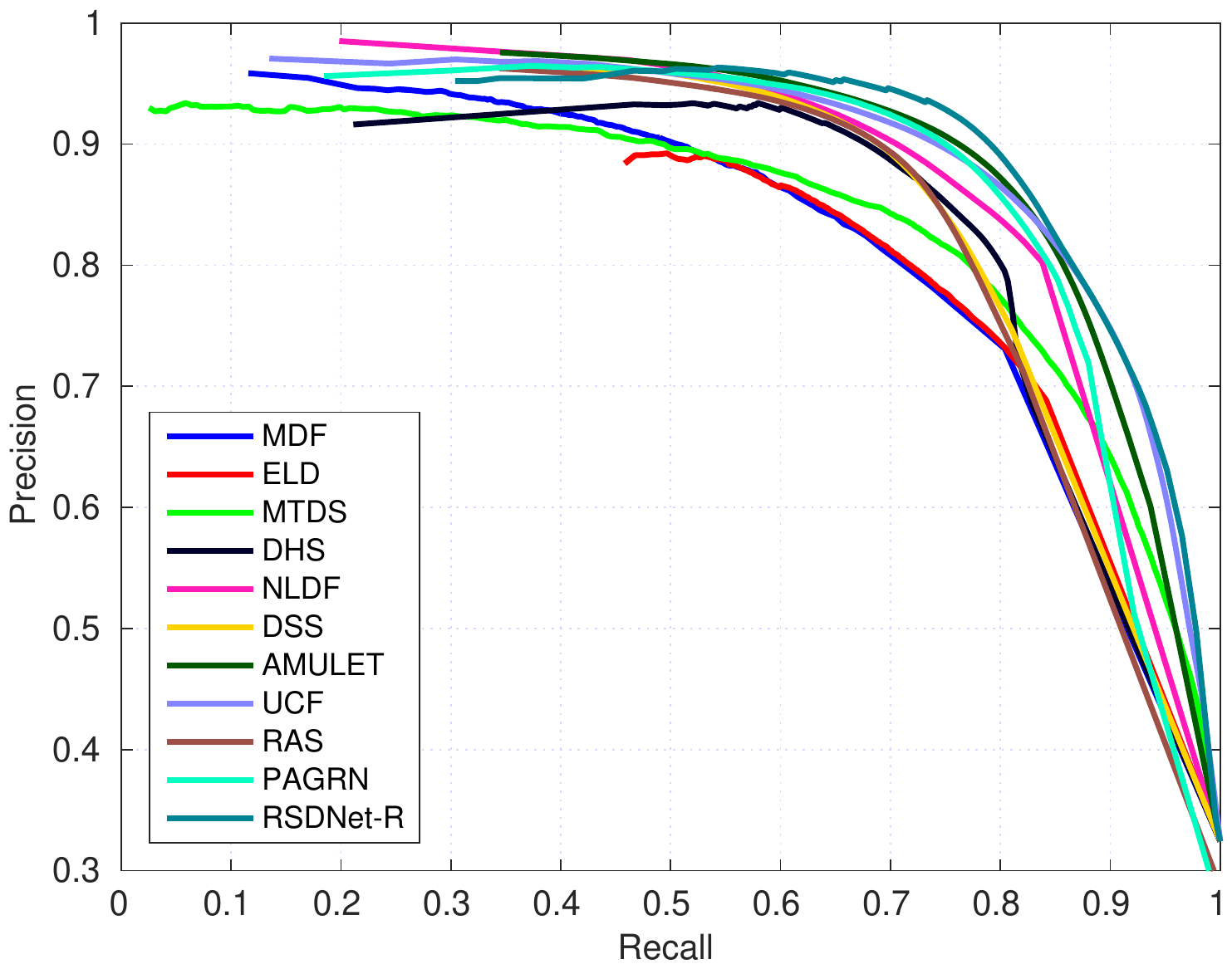}\\
			
		\end{tabular}}
		\caption{ROC (Left) and  Precision-Recall (Right) curves corresponding to different state-of-the-art methods \textcolor{violet}{on PASCAL-SR dataset}.}
		\label{fig:pr_curves_pascals}
		\vspace{-0.1cm}
\end{figure}

The problem of evaluating salient detection models is challenging in itself which has contributed to differences among benchmarks that are used. In light of these considerations, the specific evaluation we have applied to all the methods aims to remove any advantages of one algorithm over another. We compare our proposed method with recent state-of-the-art approaches, including \textcolor{violet}{PAGRN~\cite{zhang2018progressive}, RAS~\cite{chen2018reverse}, Amulet~\cite{amulet_2017}, UCF~\cite{ucf_2017}, DSS~\cite{hou2016deeply}, NLDF \cite{luo2017non}, DHSNet~\cite{liu2016dhsnet}, MDF~\cite{li2015visual}, ELD \cite{lee2016deep}, and MTDS~\cite{li2016deepsaliency}}. For fair comparison, we build the evaluation code based on the publicly available code provided in \cite{li2013contextual} and we use saliency maps provided by authors of models compared against, or by running their pre-trained models with recommended parameter settings.
\begin{table}[t]
	\caption{Quantitative comparison of methods using metrics including AUC, F-measure (max, average), MAE, average S-measure, and SOR \textcolor{violet}{on PASCAL-SR dataset}. The best three results are shown in red, brown and blue respectively.}
	\vspace{-0.3cm}
	\centering
	\def\arraystretch{1.2}
	\resizebox{0.49\textwidth}{!}{
		\begin{tabular}{l|c|c|c|c|c|c}
			\specialrule{1.2pt}{1pt}{1pt}\
			%\hline
			%\hline
			\multirow{1}{*}{$\ast$}& AUC& max-$F_m$ &avg-$F_m$ & MAE& avg-$S_m$ & SOR \\
			\specialrule{1.2pt}{1pt}{1pt}
				
			MDF~\cite{li2015visual}&0.892&0.787 &	0.730& 0.138&0.675&	0.768 \\	
			ELD~\cite{lee2016deep}	&0.916 &0.789 &	0.774 &0.123 &0.752&	0.792 \\
			MTDS~\cite{li2016deepsaliency}&	0.941&0.805 &	0.664 &0.176 &	0.737& 0.782\\
			
			DHSNet~\cite{liu2016dhsnet} &0.927&0.837 & 0.833& 0.092&0.797&0.781	\\
			
			NLDF~\cite{luo2017non}&0.933&	0.846 &0.836 & 0.099&0.793& 0.783\\
			
			DSS~\cite{hou2016deeply}& 0.918&	0.841 & 0.830 & 0.099	&0.792& 0.770\\
			
			AMULET~\cite{amulet_2017}& 0.957&	0.865& \color{red}	\textbf{0.841} & 0.097	&0.817& 0.788\\
			
			UCF~\cite{ucf_2017}&0.959& 	0.858 &	0.813 & 0.123	&0.802& 0.792\\
			RAS~\cite{chen2018reverse}&0.907& 	0.841 &	0.828 & 0.101	&0.787& 0.774\\
			PAGRN~\cite{zhang2018progressive} & 0.928 & 0.862 & \color{brown}\textbf{0.840} & \color{red}\textbf{0.089}& 0.811&0.828\\
			\hline
			\textbf{RSDNet} &\color{brown}\textbf{0.972}&0.873& 0.834 & \color{blue} \textbf{0.091} &\color{brown}\textbf{0.832}&	0.825 \\
			RSDNet-\textbf{A} &\color{red}\textbf{0.973}&\color{blue} \textbf{0.874} &0.796& 0.103&0.823& 0.838 \\

			RSDNet-\textbf{B} &0.969&\color{brown} \textbf{0.877} &0.831&0.100 & \color{blue} \textbf{0.828}&\color{blue} \textbf{0.840} \\
			
			RSDNet-\textbf{C} &\color{brown} \textbf{0.972}&\color{blue} \textbf{0.874}&0.795&0.110&0.823&\color{brown} \textbf{0.848} \\
			
			RSDNet-\textbf{R} &\color{blue}\textbf{0.971}&\color{red} \textbf{0.880} & \color{blue}\textbf{0.837} & \color{brown} \textbf{0.090} &\color{red}\textbf{0.836}& \color{red} \textbf{0.852} \\
			
			\specialrule{1.2pt}{1pt}{1pt}
		\end{tabular}}
		
		\label{table:quant_pascals}
		\vspace{-0.3cm}
\end{table}
\noindent \textbf{Quantitative Evaluation:}
Table~\ref{table:quant_pascals} shows the performance score of all the variants of our model, and other recent methods on salient object detection. It is evident that, RSDNet-R outperforms other recent approaches for \textcolor{violet}{most of the evaluation metrics including AUC, max F-measure and average S-measure}, which establishes the effectiveness of our proposed hierarchical nested relative salience stack. From the results we have few fundamental observations: (1) Our network improves the max F-measure by a considerable margin on the PASCAL-SR dataset which indicates that our model is general enough that it achieves higher precision with higher recall (see Fig.~\ref{fig:pr_curves_pascals}).
\noindent (2) \textcolor{violet}{Our model increases the average S-measure on the PASCAL-SR dataset and achieves higher area under the ROC curve (AUC) score compared to the baselines shown in Fig.~\ref{fig:pr_curves_pascals}.} (3) Although our model is only trained on a subset of PASCAL-SR, it significantly outperforms other algorithms that also leverage large-scale saliency datasets. Overall, this analysis hints at strengths of the proposed hierarchical stacked refinement strategy to provide a more accurate saliency map. In addition, it is worth mentioning that RDSNet-R outperforms all the recent deep learning based methods intended for salient object detection/segmentation without any post-processing techniques that are typically used to boost scores.

\begin{figure}[H]
	\begin{center}
		\vspace{-0.1cm}
		\setlength\tabcolsep{0.7pt}
		\def\arraystretch{0.2}
		\resizebox{0.49\textwidth}{!}{
			\begin{tabular}{*{11}{c c  @{\hskip 0.04in}c @{\hskip 0.04in} c c c  }}
				
				\includegraphics[width=0.09\textwidth]{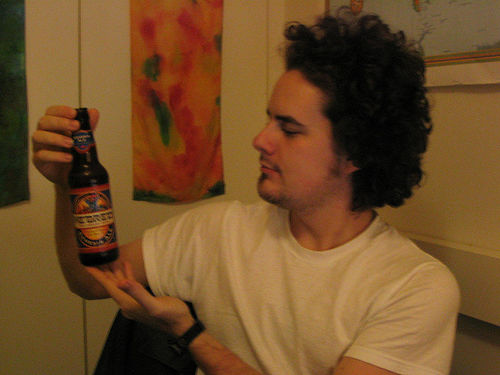}
				&
				\includegraphics[width=0.09\textwidth]{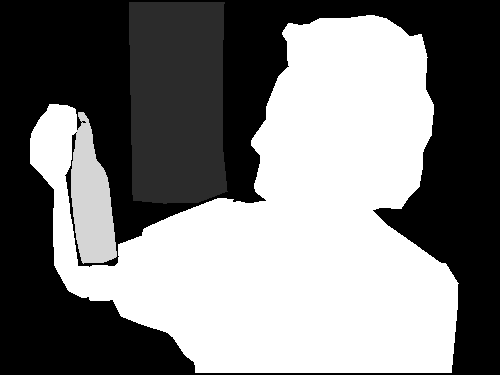}
				&
				\includegraphics[width=0.09\textwidth]{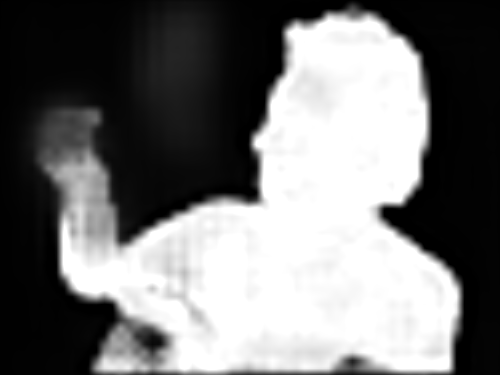}
				&
				\includegraphics[width=0.09\textwidth]{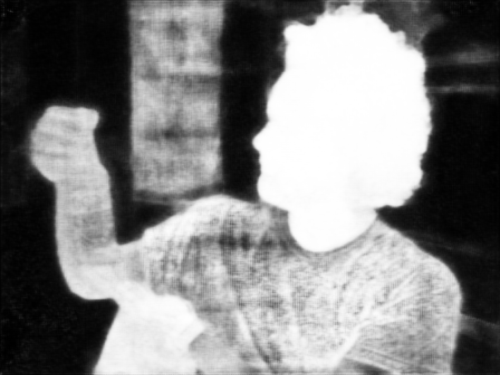}
				&
				\includegraphics[width=0.09\textwidth]{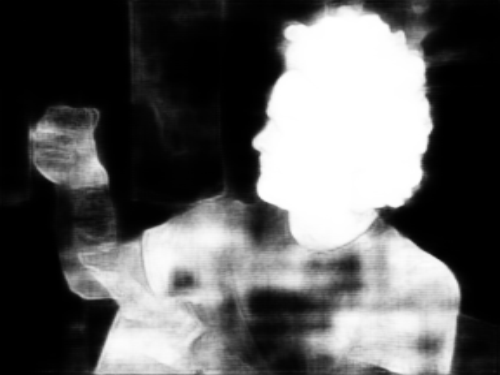}
				&
				\includegraphics[width=0.09\textwidth]{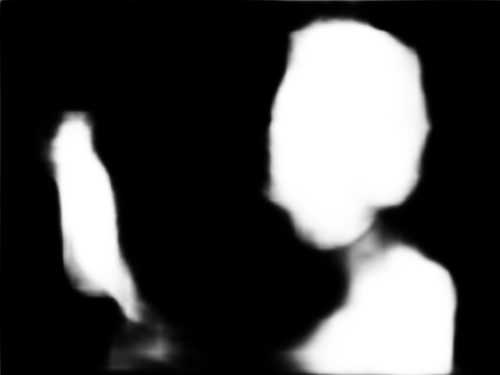}\\

				\includegraphics[width=0.09\textwidth]{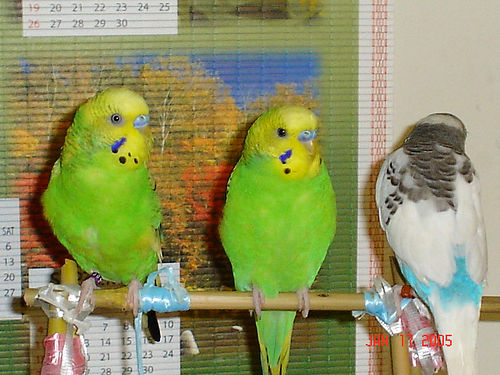}
				&
				\includegraphics[width=0.09\textwidth]{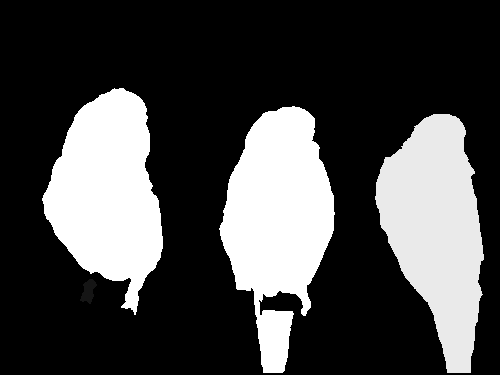}
				&
				\includegraphics[width=0.09\textwidth]{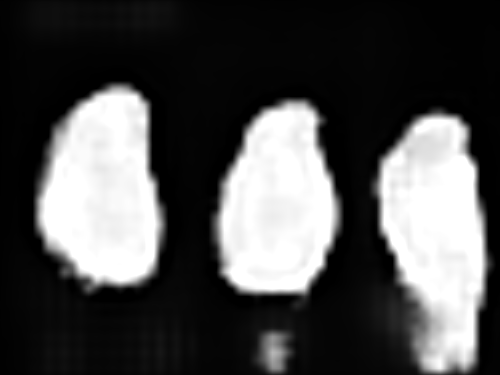}
				&
				\includegraphics[width=0.09\textwidth]{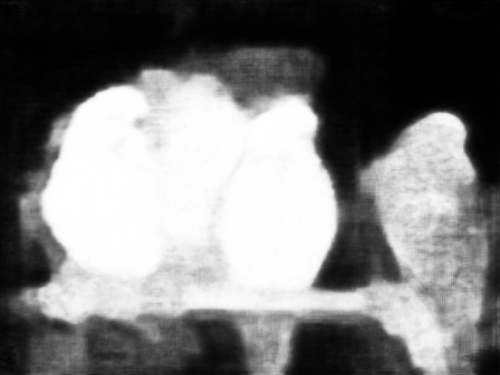}
				&
				\includegraphics[width=0.09\textwidth]{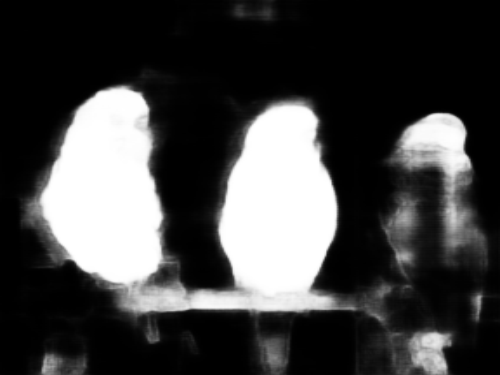}
				&
				\includegraphics[width=0.09\textwidth]{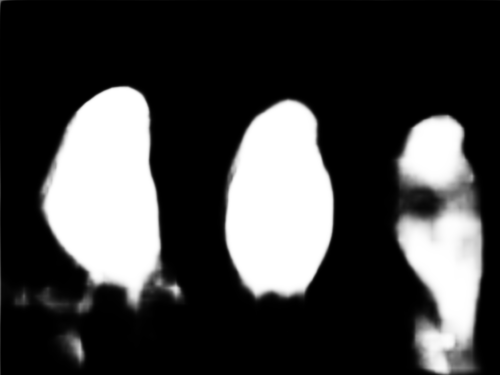} \\
				
				\specialrule{0pt}{0pt}{1pt}\

				\rowfont{\scriptsize} Image &\rowfont{\scriptsize} 	\rowfont{\scriptsize}GT &\rowfont{\scriptsize} RSDNet-R &\rowfont{\scriptsize} UCF~\cite{ucf_2017} &\rowfont{\scriptsize} Amulet~\cite{amulet_2017} &\rowfont{\scriptsize} DSS~\cite{hou2016deeply} 
			\end{tabular}}
			%-------------------------------------
			
			\caption{Predicted salient object regions for the PASCAL-SR dataset. Each row shows outputs corresponding to different algorithms designed for the salient object detection/segmentation task.}
			\label{fig:qual_pascals}
		\end{center}
		\vspace{-0.2cm}
\end{figure}

\noindent \textbf{Qualitative Evaluation:}
Fig.~\ref{fig:qual_pascals} depicts a visual comparison of RSDNet-R with respect to other state-of-the-art methods on the PASCAL-SR dataset. We can see that our method can predict salient regions accurately and produces output closer to ground-truth maps in various challenging cases e.g., instances touching the image boundary ($1^\text{st}$ row), multiple instances of same object ($2^\text{nd}$ row). The nested relative salience stack at each stage provides distinct representations to differentiate between multiple salient objects and allows for reasoning about their relative salience to take place.
%-------------------Salient region prediction------------------
\subsection{Application: Ranking by Detection}
As salient instance ranking is a completely new problem, there is not existing benchmark. In order to promote this direction of studying this problem, we are interested in finding the ranking of salient objects from the predicted saliency map. Rank order of a salient instance is obtained by averaging the degree of saliency within that instance mask. %We can write the operation as follows:
\begin{gather}
\text{Rank} (\mathcal{S}_m^{T} (\delta)) = \frac{\sum_{i=1}^{\rho_\delta} \delta(x_i, y_i)}{\rho_\delta}
\end{gather}
where $\delta$ represents a particular instance of the predicted saliency map (${S}_m^{T}$), $\rho_\delta$ denotes total numbers of pixels $\delta$ contains, and $\delta(x_i, y_i)$ refers to saliency score for the pixel $(x_i, y_i)$.
While there may exist alternatives for defining rank order, this is an intuitive way of assigning this score. With that said, we expect that this is another interesting nuance of the problem to explore further; specifically salience vs. scale, and part-whole relationships.
We use the provided instance-wise segmentation and predicted saliency map to calculate the ranking for each image.
%%%%%%%%%%%%
To demonstrate the effectiveness of our approach, we compare the overall ranking score with recent approaches. Note that no prior methods report results for salient instance ranking.
We apply the proposed SOR evaluation metric to report how different models gauge relative salience. The last column in Table~\ref{table:quant_pascals} shows the SOR score of our approach and comparisons with other state-of-the-art methods. We achieve 85.2\% correlation score for the best variant of our model. The proposed method significantly outperforms other approaches in ranking multiple salient objects and our analysis shows that learning salient object detection implicitly learns rank to some extent, but explicitly learning rank can also improve salient object detection irrespective of how the ground truth is defined. 
% Ranking Visualization
\begin{figure} [t]
	%\vspace{-0.2cm}
	\centering
	\setlength\tabcolsep{0.8pt}
	\def\arraystretch{0.8}
	\resizebox{0.48\textwidth}{!}{
		\begin{tabular}{c}
			
			\begin{tabular}{ c c c c c cccc}
				
				\fcolorbox{black}{black} {\includegraphics[width=0.12\textwidth]{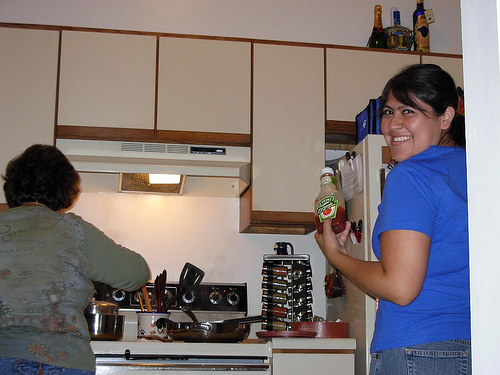}} &
				\fcolorbox{black}{black} {\includegraphics[width=0.12\textwidth]{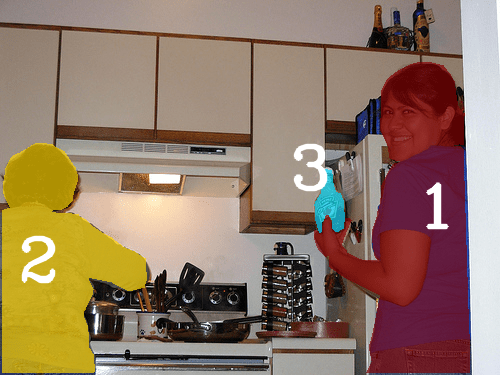}} &
				\fcolorbox{blue}{blue} {\includegraphics[width=0.12\textwidth]{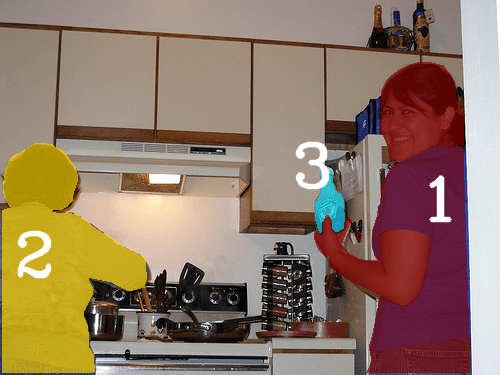}}&
				\fcolorbox{red}{red} {\includegraphics[width=0.12\textwidth]{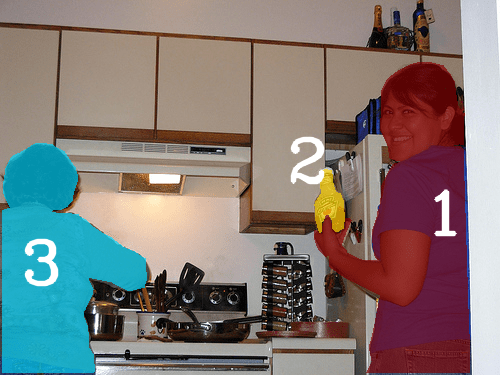}}&
				\fcolorbox{red}{red} {\includegraphics[width=0.12\textwidth]{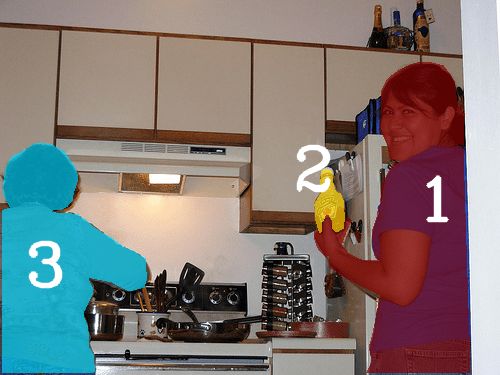}} &&&
				\multirow{3}{*}[1.43cm]{\includegraphics[width=0.031\textwidth, height = 5.6cm]{figures/jet-v.pdf}}\\

				\fcolorbox{black}{black} {\includegraphics[width=0.12\textwidth]{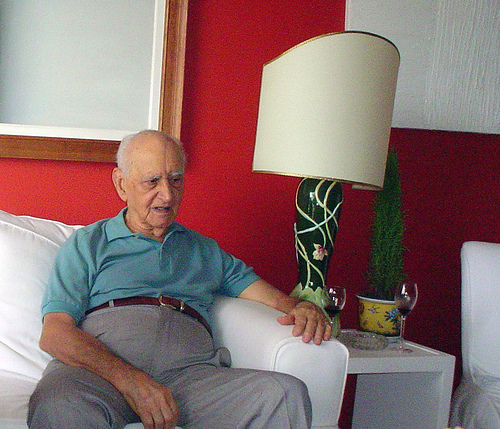}} &
				\fcolorbox{black}{black} {\includegraphics[width=0.12\textwidth]{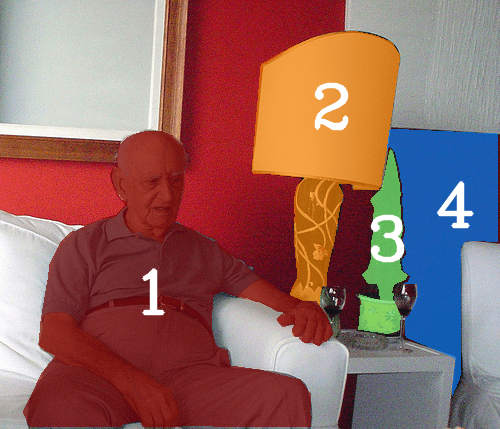}} &
				\fcolorbox{blue}{blue} {\includegraphics[width=0.12\textwidth]{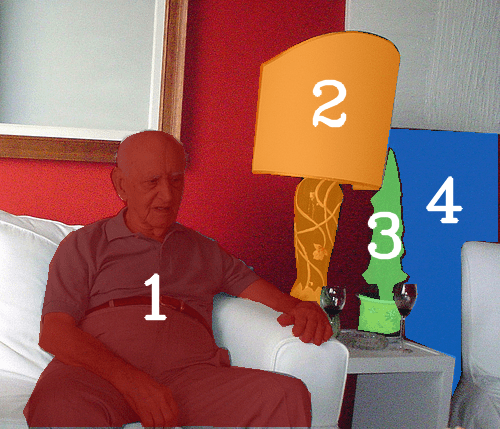}}&
				\fcolorbox{red}{red} {\includegraphics[width=0.12\textwidth]{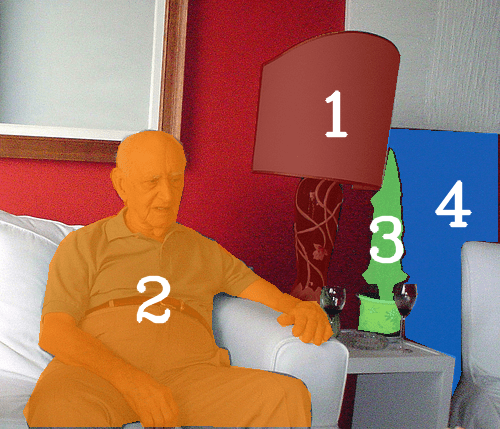}}&
				\fcolorbox{blue}{blue} {\includegraphics[width=0.12\textwidth]{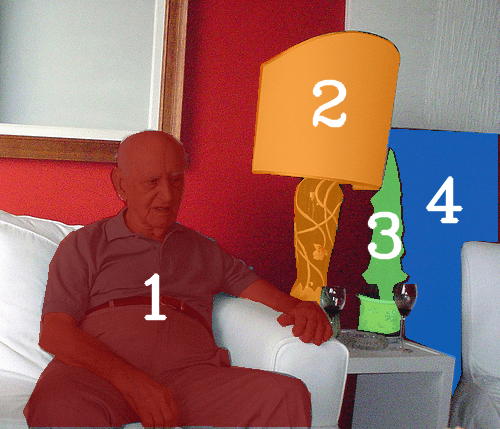}} &&&\\

				\fcolorbox{black}{black} {\includegraphics[width=0.12\textwidth]{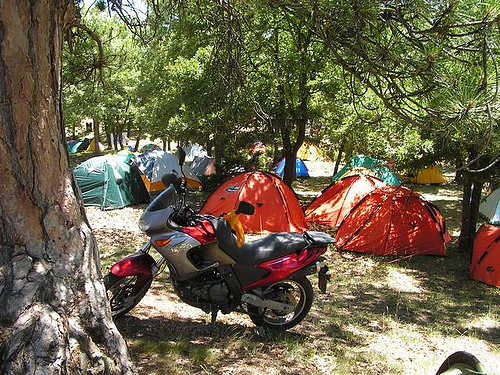}} &
				\fcolorbox{black}{black} {\includegraphics[width=0.12\textwidth]{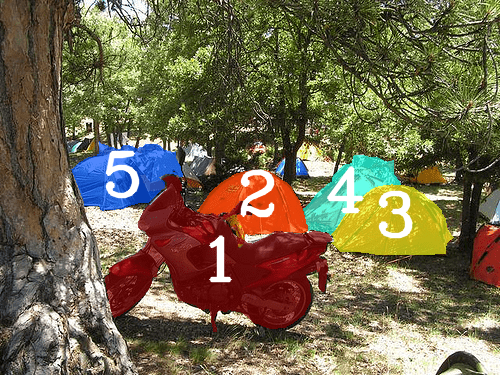}} &
				\fcolorbox{blue}{blue} {\includegraphics[width=0.12\textwidth]{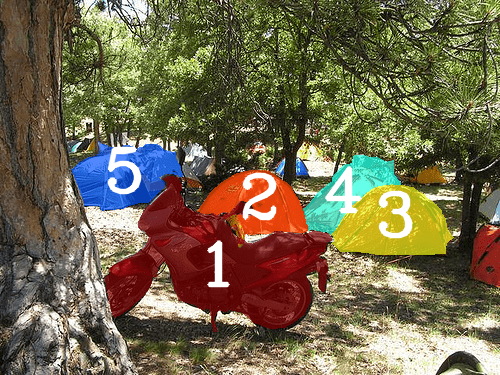}}&
				\fcolorbox{red}{red} {\includegraphics[width=0.12\textwidth]{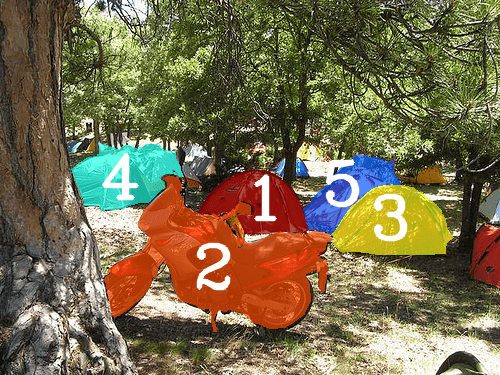}}&
				\fcolorbox{red}{red} {\includegraphics[width=0.12\textwidth]{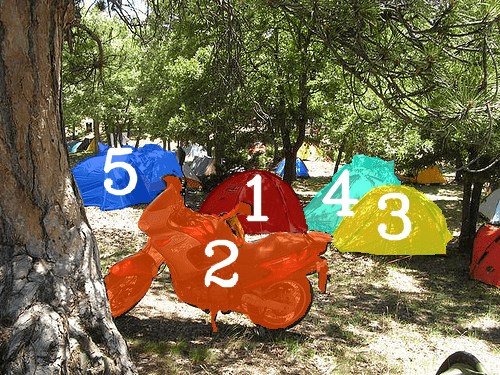}} &&&\\
				
				Image& GT& RSDNet-R&  AMULET~\cite{amulet_2017}& UCF~\cite{ucf_2017}\\
				
			\end{tabular}
			
		\end{tabular}}
		\caption{Qualitative depiction of rank order of salient objects \textcolor{violet}{on PASCAL-SR dataset}. \textcolor{violet}{ Relative rank is indicated by the assigned color and number on each salient instance}. Blue and red image borders indicate correct and incorrect ranking respectively.}
		\label{fig:pascals_ranking}
		\vspace{-0.3cm}
\end{figure}
Fig.~\ref{fig:pascals_ranking} shows a qualitative comparison of the state-of-the-art approaches designed for salient object detection on PASCAL-SR dataset. Note that the role of ranking for more than three objects is particularly pronounced.
%%%%%%%%%%%%%%%%%%%%%%
\subsection{Extended Ranking Evaluation on PASCAL-SR}
We perform several experiments on the PASCAL-SR dataset under different settings shown in Table~\ref{table:quant_pascals_extend} to justify the correctness of our proposed dataset . 
\begin{table}[H]
	\vspace{-0.2cm}
	\caption{Comparison of Saliency ranking score of several networks on the PASCAL-SR dataset. All the baseline numbers are reported from~\cite{cvpr18_rank}.}
	\vspace{-0.2cm}
	\centering
	\setlength\tabcolsep{2.9pt}
	\def\arraystretch{1.1}
	\resizebox{0.49\textwidth}{!}{
		\begin{tabular}{c|cccccccc}
			\specialrule{1.2pt}{1pt}{1pt}\
			\multirow{1}{*}{$\ast$}&RSDNet$^\star$& RSDNet$^\ddagger$ & RSDNet$^\dagger$ & RSDNet~\cite{cvpr18_rank} & UCF~\cite{luo2017non} & Amulet~\cite{luo2017non}& DSS~\cite{luo2017non}& NLDF~\cite{luo2017non}  \\
			\specialrule{1.2pt}{1pt}{1pt}\
			SOR$_\text{avg}$  &0.848 &\color{violet}\textbf{0.862}&0.832&0.825&0.792&0.788&0.770&0.783\\
			SOR$_\text{pow}$  &0.848 &0.843&\color{violet}\textbf{0.867}&0.839& 0.820& 0.823&  0.834&    0.850 \\
			SOR$_\text{max}$  &0.831&0.855&\color{violet}\textbf{0.857}&0.824& 0.837& 0.840&  0.810&    0.851 \\
			
			\specialrule{1.2pt}{1pt}{1pt}
			
		\end{tabular}}
		
		\label{table:quant_pascals_extend}
		\vspace{-0.2cm}
\end{table}
First, we train several baseline models using our proposed dataset and evaluate on the test subset of the PASCAL-SR dataset. It is a evident from Table~\ref{table:quant_pascals_extend} that, RSDNet$^\dagger$ (\textit{train} : COCO-SalRank, \textit{test}: test set of PASCAL-SR) outperforms RSDNet in term of saliency ranking. When we fine-tune the trained model with the train set of PASCAL-SR, RSDNet$^\ddagger$ further improves the ranking performance by a considerable margin. Note that we perform the same experiments for both versions of our dataset under relative ranking scenario. When we train RSDNet with COCO-SalRank (version \rom{2}) and test on PASCAL-SR we achieve a significant boost (+2\%) with RSDNet$^\star$  compared to RSDNet. One could argue that, the reason behind the improvement of ranking performance is the large training set compared to the small one of PASCAL-SR, which further underscores the fidelity and effectiveness of our dataset.
%%%%%%%%%%%%%%%%%%%%%%
%--------Salient region prediction for Pascal S ------
\begin{figure}[h]
	\vspace{-0.2cm}
	\centering
	\setlength\tabcolsep{1.0pt}
	\resizebox{0.49\textwidth}{!}{
		\begin{tabular}{c}
			
			\begin{tabular}{ c c  c c c c ccc}
				
				\fcolorbox{black}{black} {\includegraphics[width=0.12\textwidth]{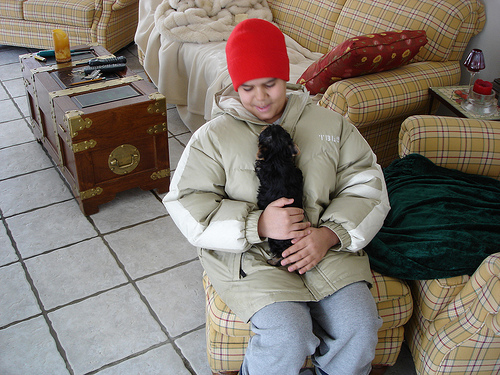}}&
				\fcolorbox{black}{black} {\includegraphics[width=0.12\textwidth]{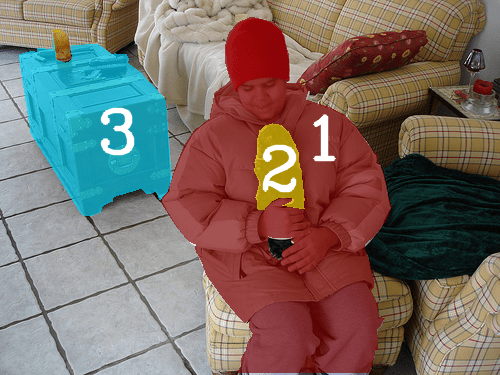}}&
				\fcolorbox{black}{black} {\includegraphics[width=0.12\textwidth]{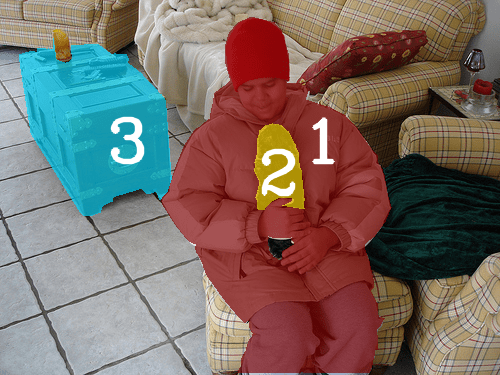}}&
				\fcolorbox{blue}{blue} {\includegraphics[width=0.12\textwidth]{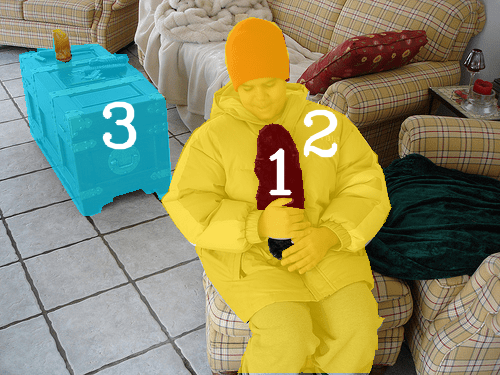}}&
				\fcolorbox{blue}{blue} {\includegraphics[width=0.12\textwidth]{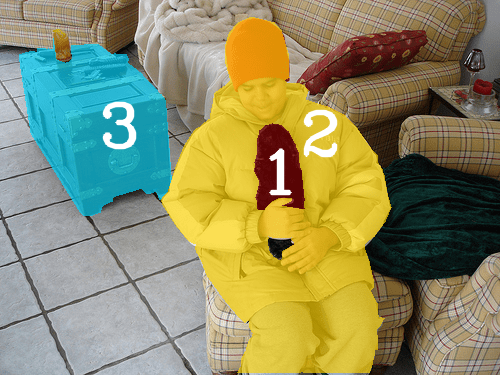}}&
				\fcolorbox{red}{red} {\includegraphics[width=0.12\textwidth]{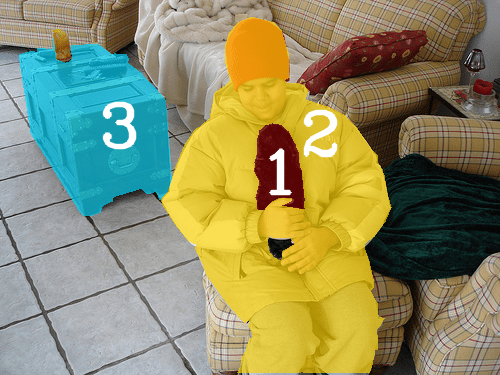}} &&&
				\multirow{2}{*}[1.4cm]{\includegraphics[width=0.031\textwidth]{figures/jet-v.pdf}}\\
				
				\fcolorbox{black}{black} {\includegraphics[width=0.12\textwidth]{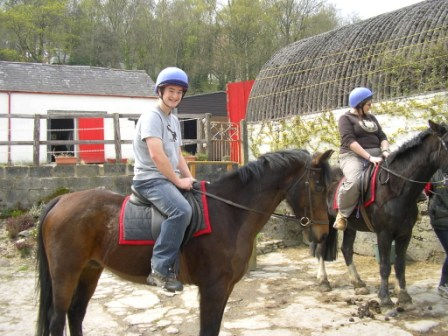}}&
				\fcolorbox{black}{black} {\includegraphics[width=0.12\textwidth]{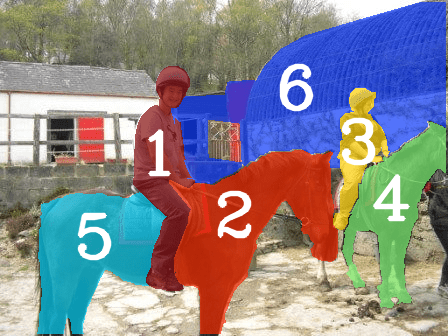}}&
				\fcolorbox{black}{black} {\includegraphics[width=0.12\textwidth]{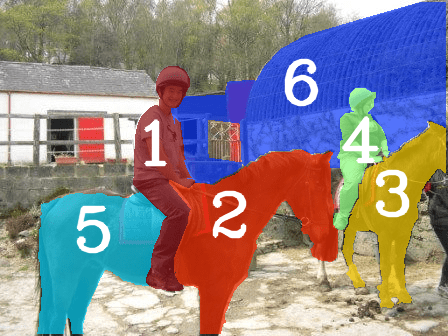}}&
				\fcolorbox{blue}{blue} {\includegraphics[width=0.12\textwidth]{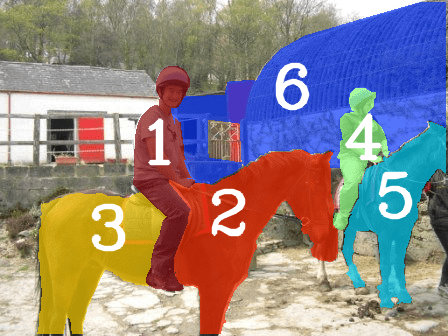}}&
				\fcolorbox{blue}{blue} {\includegraphics[width=0.12\textwidth]{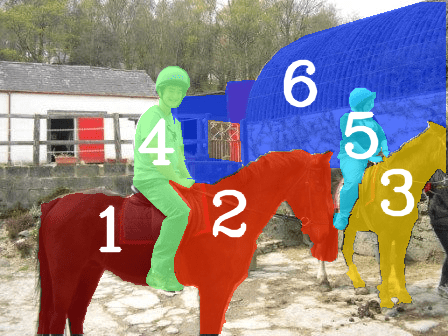}}&
				\fcolorbox{red}{red} {\includegraphics[width=0.12\textwidth]{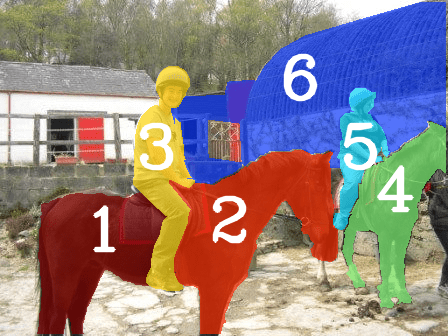}} &&&\\
				
				image & gt& RSDNet$^\ast$&RSDNet~\cite{cvpr18_rank}&  AMULET~\cite{amulet_2017}& UCF~\cite{ucf_2017}\\
				
			\end{tabular}
			
		\end{tabular}}
		\caption{ Qualitative illustration of rank order of salient instances \textcolor{violet}{on PASCAL-SR dataset}. \textcolor{violet}{Relative rank is indicated by the assigned color and number on each salient instance.}} %Note that RSDNet$^\ast$ is trained with COCO-SalRank dataset and fine-tuned on Pascal-S training set.}
		\label{fig:pascals_ranking_ex}
		\vspace{-0.1cm}
\end{figure}

Fig.~\ref{fig:pascals_ranking_ex} presents a qualitative comparison of the state-of-the-art approaches designed for salient instance ranking or detection. Note that the role of ranking for more than three objects (2$^{\text{nd}}$ row) is especially pronounced in revealing the significance of our dataset.
%%%%%%%%%%%%%%%%%%%%%%%%%%%%
\subsection{Ranking Evaluation on COCO-SalRank Under Relative and Absolute Rank Setting}\label{sec:pascals_rank}
The saliency ranking performance of all baseline models under relative and absolute settings are listed in Table~\ref{table:rank-noisy} (version \rom{1} \& version \rom{2}). The performance in terms of all metrics reveals the generality and reliability of the proposed dataset. Given the more limited number of training samples, baseline networks achieve a slightly lower ranking score from version \rom{2} compared to version \rom{1}, which implies and interesting trade-off between signal-to-noise in the data (version \rom{1}) and sheer volume of data. Also the two different settings (relative vs absolute) maintain a comparable performance across models and metrics. Further, as shown in Table~\ref{table:rank-noisy}, the $\text{SOR}_\text{max}$ ranking strategy shows higher scores for all baselines under different settings, demonstrating value in considering different vantage points in how assignments of rank are derived from an underlying saliency map.

\begin{table}[H]
	\vspace{-0.2cm}
	\caption{Saliency ranking performance comparison for different methods subject to relative and absolute ranking settings on our COCO-SalRank dataset.}
	\vspace{-0.2cm}
	\centering
	\def\arraystretch{1.2}
	\resizebox{0.49\textwidth}{!}{
		\begin{tabular}{c|c|ccc|ccc}
			\specialrule{1.2pt}{1pt}{1pt}\	
			
			\multirow{2}{*}{$\ast$} &\multirow{2}{*}{Methods}& \multicolumn{3}{c}{Relative }&\multicolumn{3}{c}{Absolute } \\
			\cline{3-8}
			&& SOR$_\text{avg}$ & SOR$_\text{pow}$ & SOR$_\text{max}$ &SOR$_\text{avg}$ &SOR$_\text{pow}$ & SOR$_\text{max}$ \\
			
			\cline{1-1} \cline{2-2} \cline{3-5} \cline{6-8}
			
			\multirow{3}{*}{ver I}&RSDNet~\cite{cvpr18_rank} & \color{violet}\textbf{0.736} &\color{violet}\textbf{0.727} &	\color{violet}\textbf{0.767}&  \color{violet}\textbf{0.732}	&\color{violet}\textbf{0.726}	&0.761  \\
			&PSPNet + NRSS~\cite{zhao2017pyramid} & 0.720&	0.709&	0.764 & 0.730	&0.725	&\color{violet}\textbf{0.770} \\
			&DeepLabv2 + NRSS~\cite{chen2016deeplab} & 0.708&	0.710	&0.753 & 0.716&	0.713	&0.750\\

			\specialrule{1.2pt}{1pt}{1pt}\
			
			\multirow{3}{*}{ver II}&RSDNet~\cite{cvpr18_rank} & \color{violet}\textbf{0.715}	&\color{violet}\textbf{0.693}	&\color{violet}\textbf{0.758}   &    \color{violet}\textbf{0.709}&	\color{violet}\textbf{0.696}&	0.745  \\
			&PSPNet + NRSS~\cite{zhao2017pyramid} & 0.689&	0.682&	0.754    &   0.695&	0.692&	\color{violet}\textbf{0.771} \\
			&DeepLabv2 + NRSS~\cite{chen2016deeplab} & 0.680	&0.682	&0.742   &   0.676&	0.688	& 0.727\\
			
			\specialrule{1.2pt}{1pt}{1pt}
		\end{tabular}}
		
		\label{table:rank-noisy}
		\vspace{-0.1cm}
\end{table}
%%%%%%%%%%%%%%%%%%%%%55
 Fig.~\ref{fig:cocosalrank_ranking} shows a visual comparison of saliency ranking on COCO-SalRank dataset with respect to different baselines. We can see that the baselines can predict rank order of salient objects quite accurately and this produces output closer to ground-truth maps in various challenging cases. Recall that each model is paired with the nested relative salience stack (NRSS)~\cite{cvpr18_rank} at each slice and provides distinct representations to differentiate between multiple salient objects and allows for reasoning about their
relative salience to take place.

%%%%%%%%%%%%% COCO-SalRank Ranking Visualization
\begin{figure}[h]
	\vspace{-0.2cm}
	\centering
	\setlength\tabcolsep{0.7pt}
	\def\arraystretch{0.9}
	\resizebox{0.49\textwidth}{!}{
		\begin{tabular}{c}
			
			\begin{tabular}{ c c   c c cccc }
				
				 \fcolorbox{red}{red}{\includegraphics[width=0.12\textwidth]{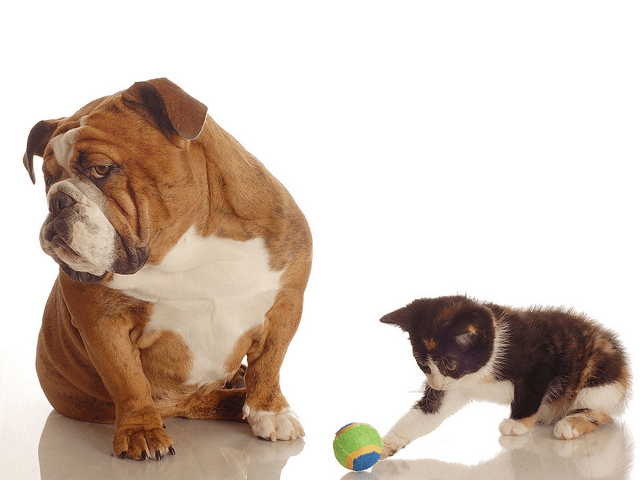}}&
				\fcolorbox{red}{red}{\includegraphics[width=0.12\textwidth]{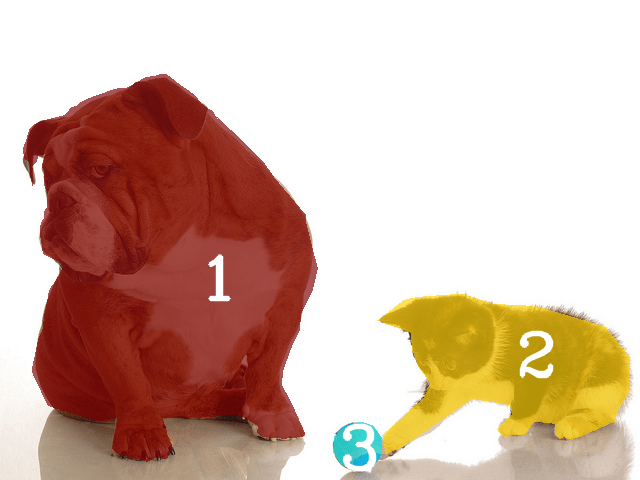}}&
				 %{\includegraphics[width=0.12\textwidth]{images/ranking_ours/im_86_saliency.png}}&
				 \fcolorbox{red}{red}{\includegraphics[width=0.12\textwidth]{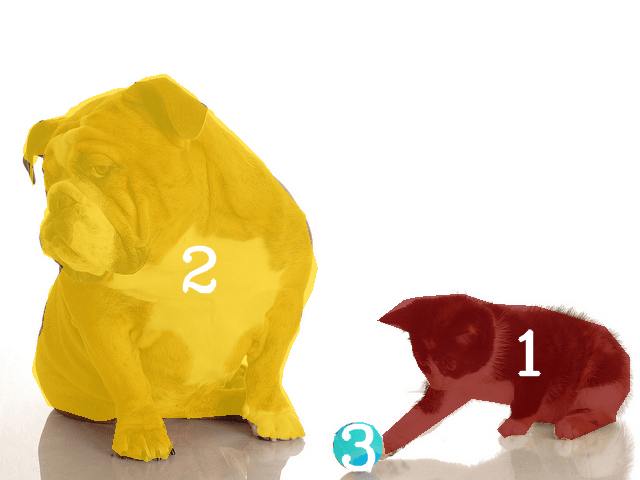}}&
				 \fcolorbox{red}{red}{\includegraphics[width=0.12\textwidth]{images/ranking_ours/im_86_pspnet.png}}&
				 \fcolorbox{red}{red}{\includegraphics[width=0.12\textwidth]{images/ranking_ours/im_86_vgg16.png}} &&&
				 \multirow{4}{*}[1.4cm]{\includegraphics[width=0.03\textwidth, height=5.3cm]{figures/jet-v.pdf}}\\
				
				 \fcolorbox{red}{red}{\includegraphics[width=0.12\textwidth]{images/ranking_ours/im_92_im.png}}&
				 \fcolorbox{red}{red}{\includegraphics[width=0.12\textwidth]{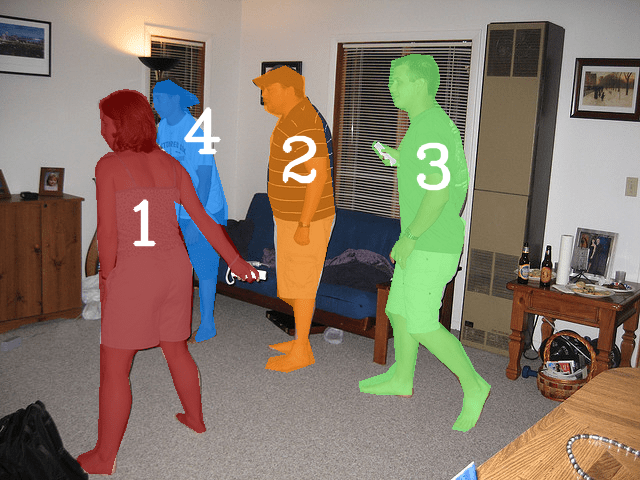}}&
				 %{\includegraphics[width=0.12\textwidth]{images/ranking_ours/im_92_saliency.png}}&
				 \fcolorbox{red}{red}{\includegraphics[width=0.12\textwidth]{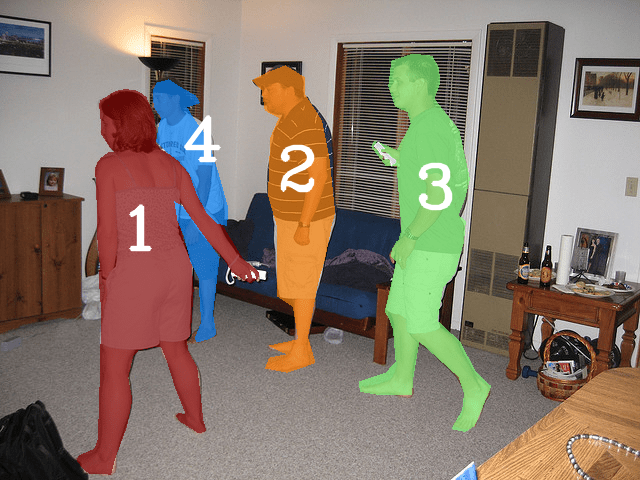}}&
				 \fcolorbox{red}{red}{\includegraphics[width=0.12\textwidth]{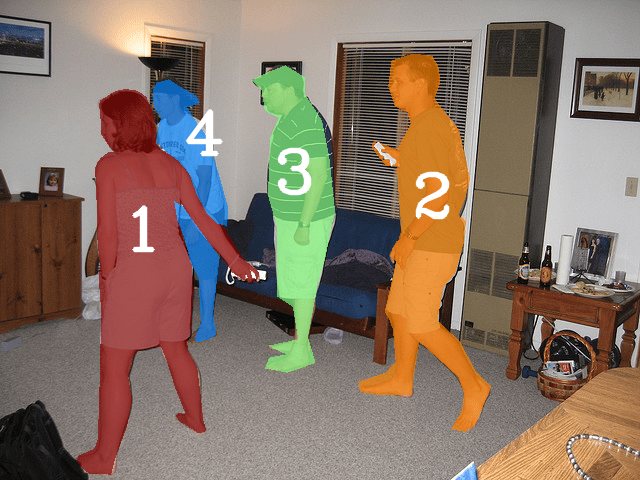}}&
				 \fcolorbox{red}{red}{\includegraphics[width=0.12\textwidth]{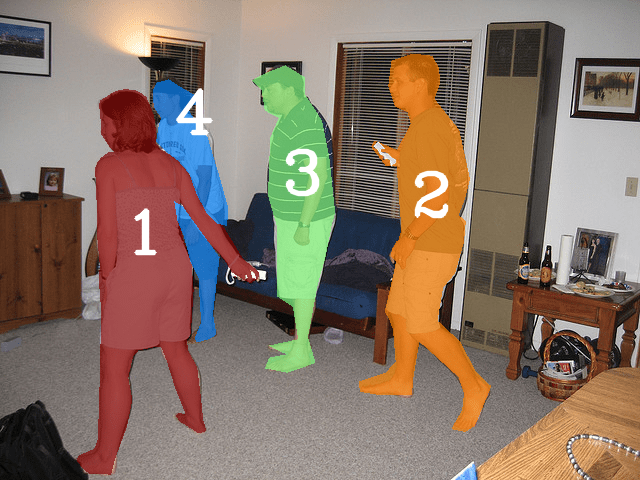}}&&& \\	

				 \fcolorbox{red}{red}{\includegraphics[width=0.12\textwidth]{images/ranking_ours/im_2762_im.png}}&
				 \fcolorbox{red}{red}{\includegraphics[width=0.12\textwidth]{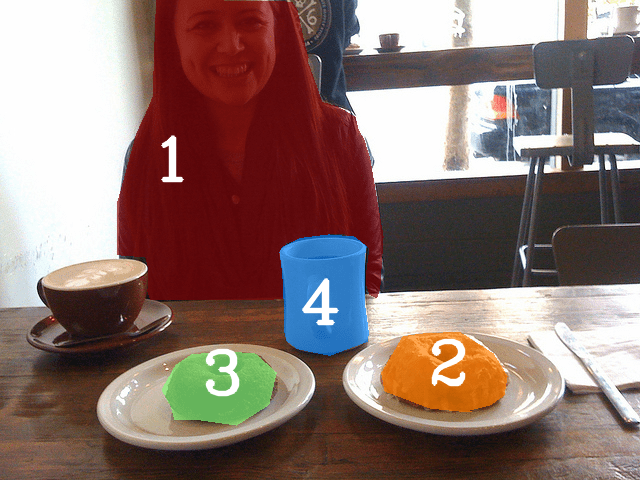}}&
				% {\includegraphics[width=0.12\textwidth]{images/ranking_ours/im_2762_saliency.png}}&
				\fcolorbox{red}{red} {\includegraphics[width=0.12\textwidth]{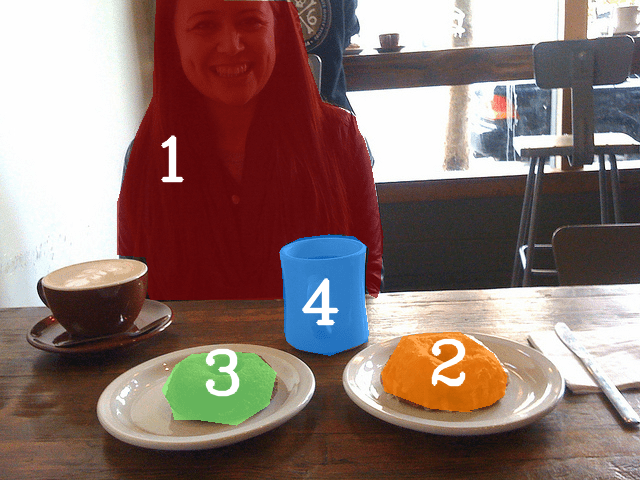}}&
				 \fcolorbox{red}{red}{\includegraphics[width=0.12\textwidth]{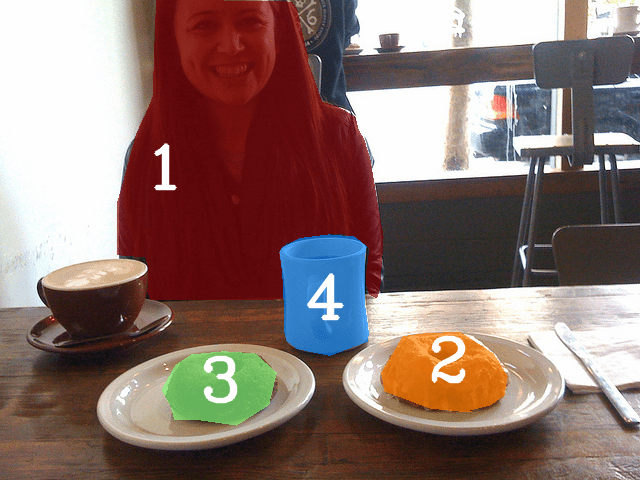}}&
				 \fcolorbox{red}{red}{\includegraphics[width=0.12\textwidth]{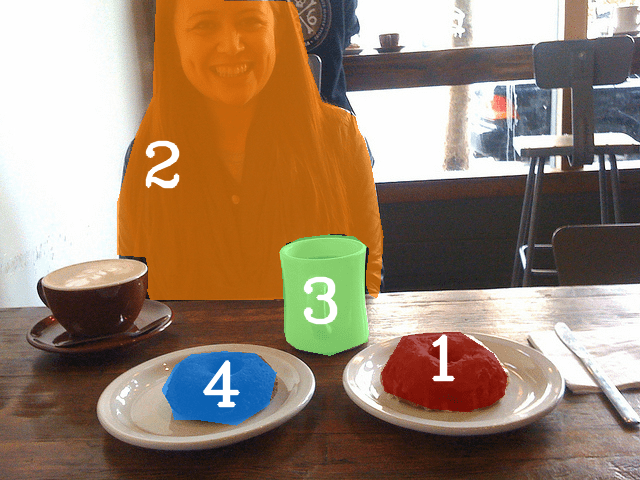}} &&&\\
				
				image & gt&RSDNet& PSPNet&  DeepLabv2\vspace{0.1cm}&&&\\

			\end{tabular}
			
		\end{tabular}}
		\caption{ Qualitative illustration of rank order of salient objects on COCO-SalRank dataset. \textcolor{violet}{Relative rank is indicated by the assigned color and number on each salient instance.}}
		\label{fig:cocosalrank_ranking}
		\vspace{-0.3cm}
\end{figure}
%%%%%%%%%%%%%%%%%%%%%%%%%%%5
\subsection{Extended Detection Evaluation on PASCAL-SR Under Relative and Absolute Rank}\label{sec:detection}
To further evaluate the generalization capability and reliability of our proposed dataset, we use the predicted saliency maps for the traditional problem of salient object detection. Note that we do not explicitly train the networks for detection; instead we use the same predicted saliency map produced by networks trained for the ranking problem. We start with performing comprehensive analysis on a subset of the PASCAL-SR dataset to demonstrate the value of our dataset on multiple salient object detection with performance values shown in Table~\ref{table:quant_pascals_detect}.
%%%%%%% Quantitative results for F-measure and AUC on our dataset
%%%%%%%% Ranking on pascal-S

\begin{table}[t]
	\caption{Quantitative comparison of baselines including max F-measure, avg F-measure, AUC, MAE, and avg S-measure on PASCAL-SR dataset.}
	\vspace{-0.3cm}
	\centering
	\def\arraystretch{1.1}
	\resizebox{0.49\textwidth}{!}{
		\begin{tabular}{c|ccccccc}
			\specialrule{1.2pt}{1pt}{1pt}\
			\multirow{1}{*}{$\ast$}& RSDNet$^\star$& RSDNet$^\ddagger$ & RSDNet$^\dagger$ & RSDNet & UCF~\cite{luo2017non} & Amulet~\cite{luo2017non}& DSS~\cite{luo2017non}  \\
			\specialrule{1.2pt}{1pt}{1pt}\
			max-$F_m$& 0.849  &\color{violet}\textbf{0.889} &0.855   &0.873 & 0.858  & 0.865 & 0.841 \\
			avg-$F_m$&0.787 &\color{violet}\textbf{0.834} &   0.789 &  \color{violet}\textbf{0.834} & 0.813 & 0.841 & 0.830   \\
			AUC&   0.963& \color{violet}\textbf{0.977} & 0.962  & 0.972&0.959 & 0.957&  0.918  \\
			
			MAE&0.092 & \color{violet}\textbf{0.082}&   0.095  & 0.091&0.123 & 0.097& 0.099 \\
			avg-$S_m$ &0.829& \color{violet}\textbf{0.847} &0.830 &  0.832&0.802&0.817&0.792\\
			
			\specialrule{1.2pt}{1pt}{1pt}
			
		\end{tabular}}
	
		\label{table:quant_pascals_detect}
		\vspace{-0.2cm}
		
\end{table}
%%%%%%%%%%%%%%%%%%%%%%%
%%%%%
Similar to (Sec.~\ref{sec:pascals_rank}), we train a baseline network (RSDNet) with our dataset (version \rom{1}) under the relative rank setting and test on PASCAL-SR. Further, in RSDNet$^\ddagger$ we fine-tune the trained model with the training set of PASCAL-SR and evaluate on the test set. From Table~\ref{table:quant_pascals_detect}, we can see that RSDNet$^\ddagger$ outperforms RSDNet in terms of all metrics with a reasonable margin. Fig.~\ref{fig:qual_pascals_ext} depicts a visual comparison of baseline methods along with RSDNet$^\ddagger$. Considering the different strategy for assigning relative salience on the PASCAL-SR dataset, the baseline models achieve comparable numbers only when trained on our dataset.
%--------Salient region prediction for Pascal S ------
\begin{figure}[h]
	\vspace{-0.2cm}
	\begin{center}
		\setlength\tabcolsep{0.7pt}
		\def\arraystretch{0.2}
		\resizebox{0.49\textwidth}{!}{
			\begin{tabular}{*{11}{c c  @{\hskip 0.04in}c @{\hskip 0.04in} c c c c     }}
				
				\includegraphics[width=0.09\textwidth]{images/pred/201_image.png}
				&
				\includegraphics[width=0.09\textwidth]{images/pred/201_GT.png}
				&
				\includegraphics[width=0.09\textwidth]{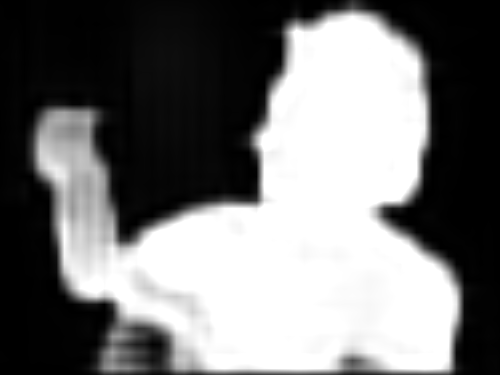}
				&
				\includegraphics[width=0.09\textwidth]{images/pred/201_OURS_V2.png}
				&
				\includegraphics[width=0.09\textwidth]{images/pred/201_UCF.png}
				&
				\includegraphics[width=0.09\textwidth]{images/pred/201_AMULET.png}
				&
				\includegraphics[width=0.09\textwidth]{images/pred/201_DSS.png}
				  \\

								\includegraphics[width=0.09\textwidth]{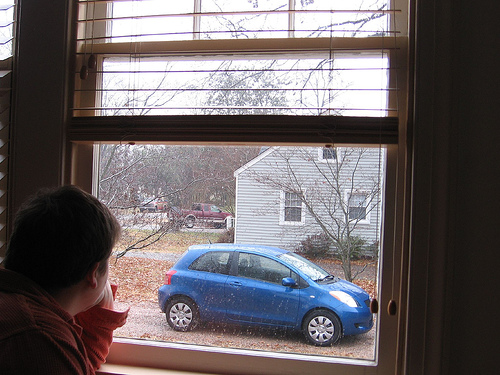}
								&
								\includegraphics[width=0.09\textwidth]{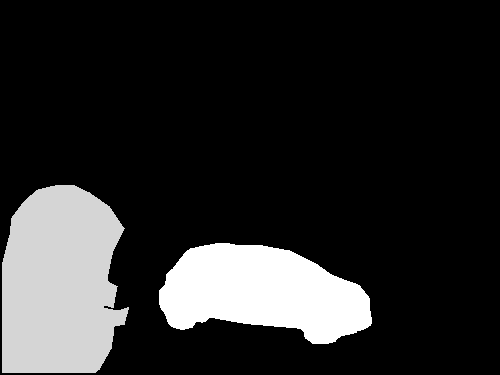}
								&
								\includegraphics[width=0.09\textwidth]{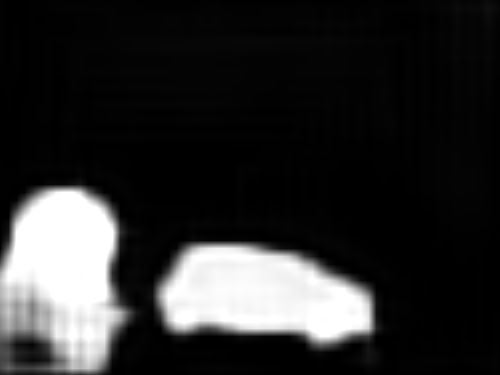}
								&
								\includegraphics[width=0.09\textwidth]{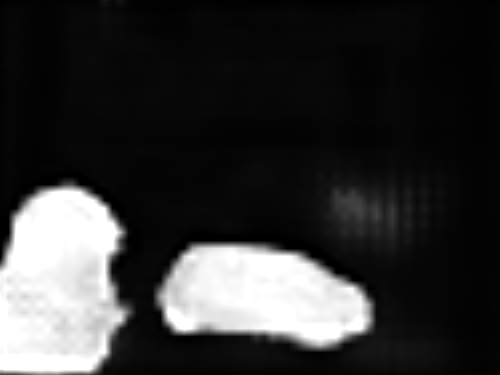}
								&
								\includegraphics[width=0.09\textwidth]{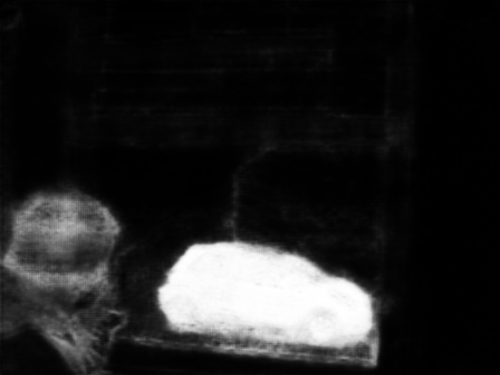}
								&
								\includegraphics[width=0.09\textwidth]{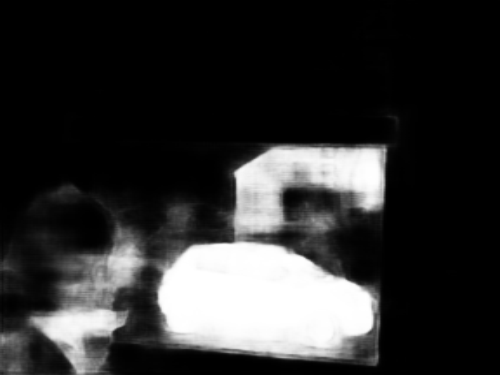}
								&
								\includegraphics[width=0.09\textwidth]{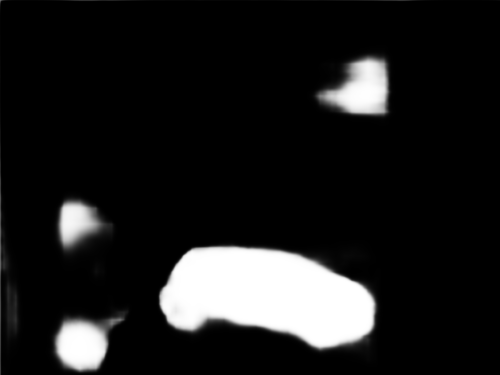}
								
								\\

				\includegraphics[width=0.09\textwidth]{images/pred/796_image.png}
				&
				\includegraphics[width=0.09\textwidth]{images/pred/796_GT.png}
				&
				\includegraphics[width=0.09\textwidth]{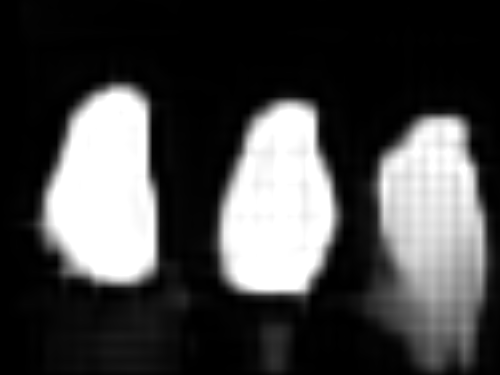}
				&
				\includegraphics[width=0.09\textwidth]{images/pred/796_OURS_V2.png}
				&
				\includegraphics[width=0.09\textwidth]{images/pred/796_UCF.png}
				&
				\includegraphics[width=0.09\textwidth]{images/pred/796_AMULET.png}
				&
				\includegraphics[width=0.09\textwidth]{images/pred/796_DSS.png}
				  \\
				
				\specialrule{0pt}{0pt}{1pt}\

				\rowfont{\scriptsize} Image &\rowfont{\scriptsize} 	\rowfont{\scriptsize}GT &\rowfont{\scriptsize} RSDNet$^\star$ & \rowfont{\scriptsize} RSDNet &\rowfont{\scriptsize} UCF &\rowfont{\scriptsize} Amulet &\rowfont{\scriptsize} DSS  \\
			\end{tabular}}
			%-------------------------------------
			
			\caption{Predicted salient object regions for the Pascal-SR dataset. Each row shows outputs corresponding to different algorithms designed for the salient object detection/segmentation task.}
			\label{fig:qual_pascals_ext}
			\vspace{-0.2cm}
		\end{center}
\end{figure}

For further analysis, we report detection performance for baseline models on our dataset (version \rom{1} \& version \rom{2}). Table~\ref{table:noisy-analysis} shows the comparison of detection score for baseline methods in terms of F-measure, AUC, and MAE. Similar to saliency ranking performance, we see consistent performance across different versions, metrics, and ranking settings.
\begin{table}[H]
	\vspace{-0.3cm}
	\caption{\textcolor{violet}{Quantitative comparison of baseline methods including max and average F-measure, AUC, and MAE on COCO-SalRank dataset (version \rom{1} \& version \rom{2}) under relative and absolute ranking settings.}}
	\vspace{-0.3cm}
	\centering
	\def\arraystretch{1.1}
	\resizebox{0.49\textwidth}{!}{
		\begin{tabular}{c|c|cccc|cccccc}
			\specialrule{1.2pt}{1pt}{1pt}\	
			
			\multirow{2}{*}{$\ast$}&\multirow{2}{*}{Methods}& \multicolumn{4}{c}{Relative}&\multicolumn{4}{c}{Absolute} \\
			\cline{3-10}
			&& max-$F_m$&  avg-$F_m$ & AUC & MAE &max-$F_m$ &avg-$F_m$ & AUC&  MAE \\
			
			\cline{1-1}  \cline{2-2} \cline{3-6} \cline{7-10}
			%\hline
			%\hline
			
			\multirow{3}{*}{I}&RSDNet~\cite{cvpr18_rank} & 0.780	&	0.705&	0.936&	\color{violet}\textbf{0.135}   & 0.783&		0.663&	\color{violet}\textbf{0.944}&	0.158  \\
			&PSPNet+NRSS~\cite{zhao2017pyramid} & \color{violet}\textbf{0.796}&		\color{violet}\textbf{0.728}&	\color{violet}\textbf{0.944}&	0.136   & \color{violet}\textbf{0.794}&	\color{violet}\textbf{	0.737}&	0.944&	\color{violet}\textbf{0.132}  \\
			&DeepLab+NRSS~\cite{chen2016deeplab} & 0.743& 	 	0.649& 	0.917& 	0.162   & 0.735&		0.631&	0.911&	0.166  \\
			
			\specialrule{1.2pt}{1pt}{1pt}\
			
			\multirow{3}{*}{II}&RSDNet~\cite{cvpr18_rank} & 	0.803&	0.777&	0.954&	0.138   &   0.832& 	 	0.762& 	0.953& 	0.149\\
			&PSPNet+NRSS~\cite{zhao2017pyramid} & \color{violet}\textbf{0.850}	&	\color{violet}\textbf{0.792}&	\color{violet}\textbf{0.957}&	\color{violet}\textbf{0.13}   &  \color{violet}\textbf{0.845}& 	 	\color{violet}\textbf{0.79}& 	\color{violet}\textbf{0.956} &	\color{violet}\textbf{0.133} \\
			&DeepLab+NRSS~\cite{chen2016deeplab} & 0.805&		0.729&	0.934&	0.165   & 0.800&	0.692&	0.932&0.175  \\
			
			\specialrule{1.2pt}{1pt}{1pt}
			
		\end{tabular}}
		
		\label{table:noisy-analysis}
		\vspace{-0.1cm}
\end{table}
%%%%%%%%%%%%%%%%%%%%%%%%%%%%
\subsection{Cross-Dataset Evaluation}\label{sec:experiments}
To further investigate the role and value of the proposed saliency ranking dataset in detail, we conduct cross-dataset evaluation on two different respective versions of the dataset (\textit{noisy} \& \textit{clean}) under relative ranking setting. First, we report saliency ranking performance of baselines for all the alternative metrics in Table~\ref{table:rank-cross}. As shown in Table~\ref{table:rank-cross}, the set of baseline methods achieves better performance in the scenario (train:(version \rom{1}), test: (version \rom{2})) for relative ranking settings when compared with the numbers reported in Table~\ref{table:rank-noisy}. However, when we use the clean version (version \rom{2}) for training, and evaluate the model on the noisy version (version \rom{1}), all the baselines achieve a lower saliency ranking performance compared to the reported number in Table~\ref{table:rank-noisy}. This analysis hints at strengths of the proposed saliency ranking dataset in its value for boosting performance in training algorithms for saliency ranking, and also for validation and testing.
%%%%%%% Quantitative results for ranking for noisy and clean version
\begin{table}[h]
	\caption{Saliency ranking performance comparison for different methods with respect to cross-dataset evaluation under the relative ranking setting.}
	\vspace{-0.3cm}
	\centering
	\def\arraystretch{1.1}
	\resizebox{0.48\textwidth}{!}{
		\begin{tabular}{c|ccccccc}
			\specialrule{1.2pt}{1pt}{1pt}\	
			
			\multirow{2}{*}{Methods}& \multicolumn{3}{c}{train:(v\rom{1}), test: (v\rom{2})}&&\multicolumn{3}{c}{train:(v\rom{2}), test: (v\rom{1})} \\
			\cline{2-8}
			%	&&&\multicolumn{3}{c}{Relative}&&&\multicolumn{3}{c}{Relative } \\
			& SOR$_\text{avg}$ & SOR$_\text{pow}$ & SOR$_\text{max}$ &&SOR$_\text{avg}$ &SOR$_\text{pow}$ & SOR$_\text{max}$ \\
			
			\cline{1-1} \cline{2-4} \cline{6-8}
			
			RSDNet~\cite{cvpr18_rank} & \color{violet}\textbf{0.721}&	\color{violet}\textbf{0.696}&	0.780 &&\color{violet} \textbf{0.727}&	\color{violet}\textbf{0.724}&	\color{violet} \textbf{0.753} \\
			PSPNet + NRSS~\cite{zhao2017pyramid} & 0.697&	0.684&\color{violet}	\textbf{0.781} && 0.700&0.706&	0.729\\
			DeepLabv2 + NRSS~\cite{chen2016deeplab} &0.680&	0.680&	0.752  && 0.709&	0.715&	0.748\\

			\specialrule{1.2pt}{1pt}{1pt}
		\end{tabular}}
		
		\label{table:rank-cross}
		%\vspace{-0.5cm}
\end{table}
%%%%%%%%%%%%%%%%%%%%%%%%%%%%%%%%%%%%%%%
\section{Conclusion}\label{sec:conclude}
In this paper, we have presented a neural framework for detecting and ranking multiple salient objects that includes a stack refinement mechanism to achieve better performance. Central to the success of this approach, is how to represent relative saliency both in terms of ground truth, and \textit{in network} in a manner that produces stable performance. We highlight the fact that to date, salient object detection has assumed a relatively limited, and sometimes inconsistent problem definition. Comprehensive experiments demonstrate that the proposed architecture outperforms state-of-the-art approaches across a broad gamut of metrics.

In addition to this, we have considered the problem of saliency ranking from a data driven standpoint and in doing so have presented a large scale benchmark dataset for saliency ranking. We have also introduced a novel set of methods that make use of existing gaze or related data paired with object annotations to generate large-scale rank-ordered data for salient objects. We validate the proposed benchmark by applying state-of-the-art saliency models, and demonstrate the value of a nested representation in defining a relative ranking. We also show that \textcolor{violet}{models trained on our proposed dataset achieve higher ranking performance compared to PASCAL-SR}.

Many interesting research questions arise from the approach and results presented in this paper. One avenue for further investigation is to inspect different representations of the ground truth stack and their impact on the overall performance of both saliency ranking and detection tasks. Another interesting avenue to examine is how to assign rank order based on the predicted saliency maps accounting for characteristics such as instance size, object scale and other characteristics. Additionally, \textcolor{violet}{performance on traditional saliency object detection datasets has largely saturated. Further exploration of different methods to assign rank order will be a fruitful path for further research, and this paper along with the work it builds on provides a solid foundation for future efforts targeting the problem of salient object detection.}

\bibliographystyle{IEEEtran}
\bibliography{paper}

%\begin{IEEEbiography}[{\includegraphics[width=1in,height=1.25in,clip,keepaspectratio]{./bioimages/bruce.jpg}}]{Michael Shell}
\begin{IEEEbiography}[{\includegraphics[width=1in,height=1.25in,clip,keepaspectratio]{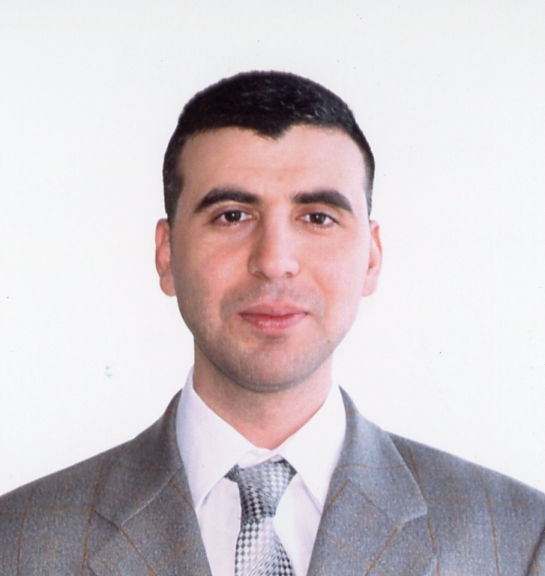}}]{Mahmoud Kalash}
received his B.Eng. degree in Automation and Computer Engineering from Damascus University, Syria in 2014. He received the M.Sc. degree in Computer Science from University of Manitoba, Canada in 2018. Currently, he is a computer vision research engineer at MicroTraffic Inc. His research interests lie in computer vision and deep learning.
\end{IEEEbiography}
\vspace{-0.9cm}
\begin{IEEEbiography}[{\includegraphics[width=1in,height=1.25in,clip,keepaspectratio]{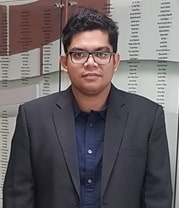}}]{Md Amirul Islam}
	 received his B.Sc. degree in Computer Science and Engineering from North South University, Bangladesh in 2014, and the M.Sc. degree in Computer Science from University of Manitoba, Canada in 2017. He is currently perusing his Ph.D. degree in Computer Science from Ryerson University, Canada. He is also a Postgraduate Affiliate at the Vector Institute, Toronto. His research interests  include  computer  vision  and  deep learning.
\end{IEEEbiography}
\vspace{-16.1cm}
\begin{IEEEbiography}[{\includegraphics[width=1in,height=1.25in,clip,keepaspectratio]{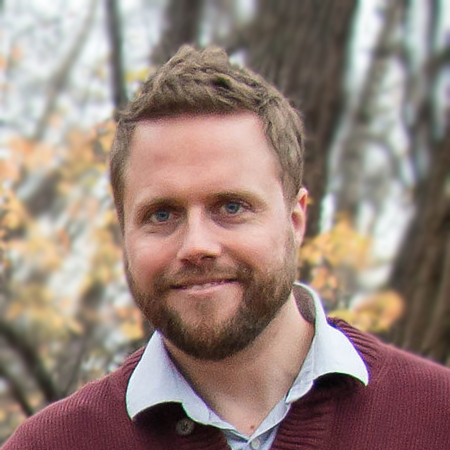}}]{Neil D. B. Bruce}
	received his B.Sc. degree in both Computer Science and Pure Mathematics from the University of Guelph, Canada in 2001. He received an M.A.Sc. degree in System Design Engineering from the University of Waterloo in 2003. He completed his Ph.D. degree at York University, Canada in 2008 and has been a faculty member at the University of Manitoba and in his current position at Ryerson University. He is a Faculty Affiliate at the Vector Institute, a member of the Institute for Biomedical Engineering, Science and Technology and a Faculty Affiliate at St. Michael's Hospital in Toronto. His research interests include computer vision, machine learning, neural networks and computational neuroscience.
\end{IEEEbiography}
%% insert where needed to balance the two columns on the last page with
%%\newpage

\end{document}